\documentclass[pdflatex,sn-mathphys]{sn-jnl}

\jyear{2024}%

\theoremstyle{thmstyleone}%
\theoremstyle{thmstyletwo}%

\theoremstyle{thmstylethree}%

\usepackage{derivative}
\usepackage{bbding}
\usepackage{xr}
\makeatletter

\newcommand*{\addFileDependency}[1]{
\typeout{(#1)}
\@addtofilelist{#1}
\IfFileExists{#1}{}{\typeout{No file #1.}}
}\makeatother

\usepackage{subcaption}
\usepackage[section]{placeins}
\usepackage{tabularx}
\usepackage{multirow}
\usepackage{CJKutf8}
\usepackage{graphicx} 
\graphicspath{{figures/}} 

\usepackage{float}
\usepackage{etoolbox}
\makeatletter
\patchcmd{\ps@headings}
{\hbox to \hsize{\hfill Springer Nature 2021 \LaTeX\ template\hfill}}
{\hbox to \hsize{}}
{}
{}
\patchcmd{\ps@titlepage}
{\hbox to \hsize{\hfill Springer Nature 2021 \LaTeX\ template\hfill}}
{\hbox to \hsize{}}
{}
{}
\makeatother
\makeatletter
\patchcmd{\ps@headings}
{\hbox to \hsize{\hfill Springer Nature 2021 \LaTeX\ template\hfill}}
{\hbox to \hsize{}}
{}
{}
\patchcmd{\ps@headings}
{\hbox to \hsize{\hfill Springer Nature 2021 \LaTeX\ template\hfill}}
{\hbox to \hsize{}}
{}
{}
\patchcmd{\ps@titlepage}
{\hbox to \hsize{\hfill Springer Nature 2021 \LaTeX\ template\hfill}}
{\hbox to \hsize{}}
{}
{}
\makeatother

\begin{document}

\begin{CJK*}{UTF8}{gbsn}

\title[FuXi Weather]{FuXi Weather: A data-to-forecast machine learning system for global weather}

\author[1]{\fnm{Xiuyu} \sur{Sun}}\email{xiuyu.sxy@gmail.com}
\equalcont{These authors contributed equally to this work.}

\author[2]{\fnm{Xiaohui} \sur{Zhong}}\email{x7zhong@gmail.com}
\equalcont{These authors contributed equally to this work.}

\author[3,1,4]{\fnm{Xiaoze} \sur{Xu}}\email{xuxiaoze1998@163.com}

\author[1]{\fnm{Yuanqing} \sur{Huang}}\email{yuanqinghuang828@gmail.com}

\author*[2,1]{\fnm{Hao} \sur{Li}}\email{lihao$\_$lh@fudan.edu.cn}

\author*[5]{\fnm{J. David} \sur{Neelin}}\email{neelin@atmos.ucla.edu}

\author[6,7]{\fnm{Deliang} \sur{Chen}}\email{deliang@gvc.gu.se}

\author[8,1]{\fnm{Jie} \sur{Feng}}\email{fengjiefj@fudan.edu.cn}

\author*[4]{\fnm{Wei} \sur{Han}}\email{hanwei@cma.gov.cn}

\author[9,10,11,1]{\fnm{Libo} \sur{Wu}}\email{wulibo@fudan.edu.cn}

\author*[2,1]{\fnm{Yuan} \sur{Qi}}\email{qiyuan@fudan.edu.cn}

\affil[1]{\orgname{Shanghai Academy of Artificial Intelligence for Science}, \orgaddress{\city{Shanghai}, \postcode{200232}, \country{China}}}

\affil[2]{\orgdiv{Artificial Intelligence Innovation and Incubation Institute}, \orgname{Fudan University}, \orgaddress{\city{Shanghai}, \postcode{200433}, \country{China}}}

\affil[3]{\orgdiv{School of Atmospheric Physics}, \orgname{Nanjing University of Information Science and Technology}, \orgaddress{\city{Nanjing}, \postcode{210044}, \country{China}}}

\affil[4]{\orgdiv{Earth System Modeling and Prediction Centre}, \orgname{China Meteorological Administration}, \orgaddress{\city{Beijing}, \postcode{100081}, \country{China}}}

\affil[5]{\orgdiv{Department of Atmospheric and Oceanic Sciences}, \orgname{University of California}, \orgaddress{\city{Los Angeles}, \postcode{90095}, \country{USA}}}

\affil[6]{\orgdiv{Ministry of Education Key Laboratory for Earth System Modeling, Department of Earth System Science}, \orgname{Tsinghua University}, \orgaddress{\city{Beijing}, \postcode{100084}, \country{China}}}

\affil[7]{\orgname{University of Gothenburg}, \orgaddress{\city{Gothenburg}, \country{Sweden}}}

\affil[8]{\orgdiv{Department of Atmospheric and Oceanic Sciences and Institute of Atmospheric Sciences}, \orgname{Fudan University}, \orgaddress{\city{Shanghai}, \postcode{200433}, \country{China}}}

\affil[9]{\orgdiv{School of Data Science}, \orgname{Fudan University}, \orgaddress{\city{Shanghai}, \postcode{200433}, \country{China}}}

\affil[10]{\orgdiv{Institute for Big Data}, \orgname{Fudan University}, \orgaddress{\city{Shanghai}, \postcode{200433}, \country{China}}}

\affil[11]{\orgdiv{MOE Laboratory for National Development and Intelligent Governance}, \orgname{Fudan University}, \orgaddress{\city{Shanghai}, \postcode{200433}, \country{China}}}

\abstract{
Weather forecasting traditionally relies on numerical weather prediction (NWP) systems that integrates global observational systems, data assimilation (DA), and forecasting models.
Despite steady improvements in forecast accuracy over recent decades, further advances are increasingly constrained by high computational costs, the underutilization of vast observational datasets, and the challenges of obtaining finer resolution.
These limitations, alongside the uneven distribution of observational networks, result in global disparities in forecast accuracy, leaving some regions vulnerable to extreme weather.
Recent advances in machine learning present a promising alternative, providing more efficient and accurate forecasts using the same initial conditions as NWP.
However, current machine learning models still depend on the initial conditions generated by NWP systems, which require extensive computational resources and expertise.
Here we introduce FuXi Weather, a machine learning weather forecasting system that assimilates data from multiple satellites.
Operating on a 6-hourly DA and forecast cycle, FuXi Weather generates reliable and accurate 10-day global weather forecasts at a spatial resolution of $0.25^\circ$.
FuXi Weather is the first system to achieve all-grid, all-surface, all-channel, and all-sky DA and forecasting, extending skillful forecast lead times beyond those of the European Centre for Medium-range Weather Forecasts (ECMWF) high-resolution forecasts (HRES) while using significantly fewer observations.
FuXi Weather consistently outperforms ECMWF HRES in observation-sparse regions, such as central Africa, demonstrating its potential to improve forecasts where observational infrastructure is limited.
}


\maketitle

\section{Introduction}

Accurate weather forecasting is essential for informed decision-making and serves as the foundation of early warning systems \cite{grasso2011early,rogers2011costs} that help to mitigate the impacts of extreme weather events and save lives.
Since the first successful numerical weather prediction (NWP) \cite{pu2019numerical} using the ENIAC computer in 1950 \cite{charney1950numerical}, forecast accuracy has steadily improved \cite{bauer2015quiet}, driven by advances in data assimilation (DA), spatial resolution, computational power, observational infrastructure, and physical parameterizations.
However, substantial global disparities remain, with wealthier nations benefiting from better resources and more accurate forecasting \cite{Linsenmeier2023}, while many low-income countries, particularly in Africa, continue to struggle with forecasts only marginally better than climatology \cite{Vogel2020}.
These disparities are especially concerning as many low-income countries are particularly vulnerable to the impacts of climate change and extreme weather \cite{Carleton2022}.

Expanding observational infrastructure could help to alleviate this issue, but the financial investment required is prohibitive for many poorer nations.
Additionally, the further enhancement of traditional NWP systems is increasingly challenging owing to high computational costs and the complexities of parallelizing models on modern supercomputers \cite{bauer2024}.
Meanwhile, recent advances in machine learning present a promising alternative, offering more efficient and accurate forecasts using the same initial conditions as traditional NWP \cite{de2023machine,bauer2024}.
State-of-the-art machine learning models, such as Pangu-Weather, GraphCast, FuXi, and AIFS \cite{pathak2022fourcastnet,bi2023accurate,lam2022graphcast,chen2023fuxi,bouallegue2024aifs}, have demonstrated forecasting skills that rival or even surpass traditional high-resolution forecasts (HRES) from the European Centre for Medium-range Weather Forecasts (ECMWF) \citep{ECMWF2021}.
Nevertheless, NWP models and DA systems remain indispensable, because they provide the initial conditions necessary for both traditional and machine learning forecasting models \cite{Bonavita2024}: this raises the question of whether machine learning based DA could further improve forecast accuracy.

DA is a complex, nonlinear process that incorporates vast, multi-source and multi-resolution observational data, often plagued by noise and missing values \cite{Karpatne2019}, involving challenges such as distinguishing the effects of clouds on satellite radiance from those of temperature and moisture, while ensuring consistency with dynamic models to minimize error growth.
Leading weather centers employ sophisticated DA methods \cite{Bannister2017,geer2018all}, such as hybrid four-dimensional ensemble-variational (4DEnVar) approaches \cite{hamill2000hybrid,Buehner2005,Wang2010}, which leverage ensembles of short-range forecasts to incorporate flow-dependent background error covariances and enhance forecast accuracy \cite{Clayton2013,Lorenc2015,Buehner2015}.
These methods, though effective, are computationally expensive and typically use only 5\%–10\% \cite{bauer2015quiet} of available observational data to deliver timely analyses.
Although progress has been made in all-sky radiance assimilation for microwave sounders, challenges remain in fully leveraging satellite data across all grids, surfaces, and channels.
With the volume of observational data projected to exceed 100 terabytes per day in the coming decade \cite{Jerald2023} and higher model resolutions further exacerbating computational demands \cite{carrassi2018data}, more efficient DA systems are urgently required \cite{Andrew2022}.

The mathematical similarities between machine learning and DA, particularly in variational methods, have inspired efforts to improve DA efficiency through machine learning \cite{cheng2023machine}.
Early attempts focused on simplified dynamic systems, such as the Lorenz63 \cite{Arcucci2021,Fablet2021} and Lorenz96 \cite{Brajard2020,Wuxin2024} models, which are far less complex than NWP models.
However, extending these approaches to operational NWP models is challenging owing to the markedly higher dimensionality of such models (on the order of $10^9$) \cite{nichols2007overview}.
Recent studies have demonstrated the potential of machine learning for specific tasks within the DA workflow, such as developing linear and adjoint models for parameterizations through automatic differentiation \cite{hatfield2021building}.
The rise of machine learning forecasting models \cite{bouallegue2024rise} has reignited interest in developing fully integrated machine-learning-based DA frameworks for end-to-end weather prediction.

One such attempt is FengWu-4DVar \cite{xiao2024towards}, which uses a simplified FengWu \cite{chen2023fengwu} model to assimilate ERA5 data \cite{hersbach2020era5}.
However, its reliance on simulated observations and its lower dimensionality limit its effectiveness in real-world scenarios.
Aardvark Weather \cite{vaughan2024aardvark} processes raw observations for forecasts but falls short of the accuracy achieved by ECMWF HRES.
These cases highlight the difficulties in developing machine learning based DA systems for real-world forecasts using actual observational data.
FuXi-DA \cite{xu2024fuxida}, a machine learning based DA framework, has shown promise by assimilating raw Fengyun-4B satellite data alongside background forecasts, but its limited spatial coverage constrains its global and cyclic DA capabilities.

To address these challenges, we here present FuXi Weather, an end-to-end machine-learning weather forecasting system capable of running cyclic DA and forecasting every 6 hours using raw observations.
FuXi Weather integrates a substantially enhanced version of FuXi-DA \cite{xu2024fuxida} with fine-tuned FuXi.
Key updates to FuXi-DA include variable- and instrument-specific encoders for diverse satellite data, and a modified PointPillars \cite{lang2019pointpillars} approach for processing sparse observations.
The FuXi-Short model is fine-tuned using FuXi-DA analysis for initial conditions, while a replay-based incremental learning strategy updates FuXi-DA monthly, ensuring the stability of the system as satellite data quality and availability evolve.

FuXi Weather assimilates raw brightness temperature data from three polar-orbiting meteorological satellites ( FengYun-3E (FY-3E), Meteorological Operational Polar Satellite - C (Metop-C), and National Oceanic and Atmospheric Administration (NOAA)-20 ), along with the radio occultation (RO) data from the Global Navigation Satellite System (GNSS), across all grids, surfaces, and channels under all weather conditions.
This represents the first realization of all-grid, all-surface, all-channel, and all-sky DA capability.
FuXi Weather demonstrates comparable 10-day forecast performance to that of ECMWF HRES, extending the skillful lead time for key variables, while using considerably less observational data compared with that used by ECMWF HRES.
Furthermore, FuXi Weather consistently outperforms ECMWF HRES in regions with sparse land-based observations, such as Africa, demonstrating its potential to provide more accurate forecasts and enhance climate resilience.
This achievement challenges the prevailing view that standalone machine learning-based weather forecasting systems are not yet viable for operational use.
Furthermore, FuXi Weather's development costs are considerably lower than those of traditional NWP systems.

\section{FuXi Weather}

\begin{figure}[h]
\centering
\includegraphics[width=1.0\textwidth]{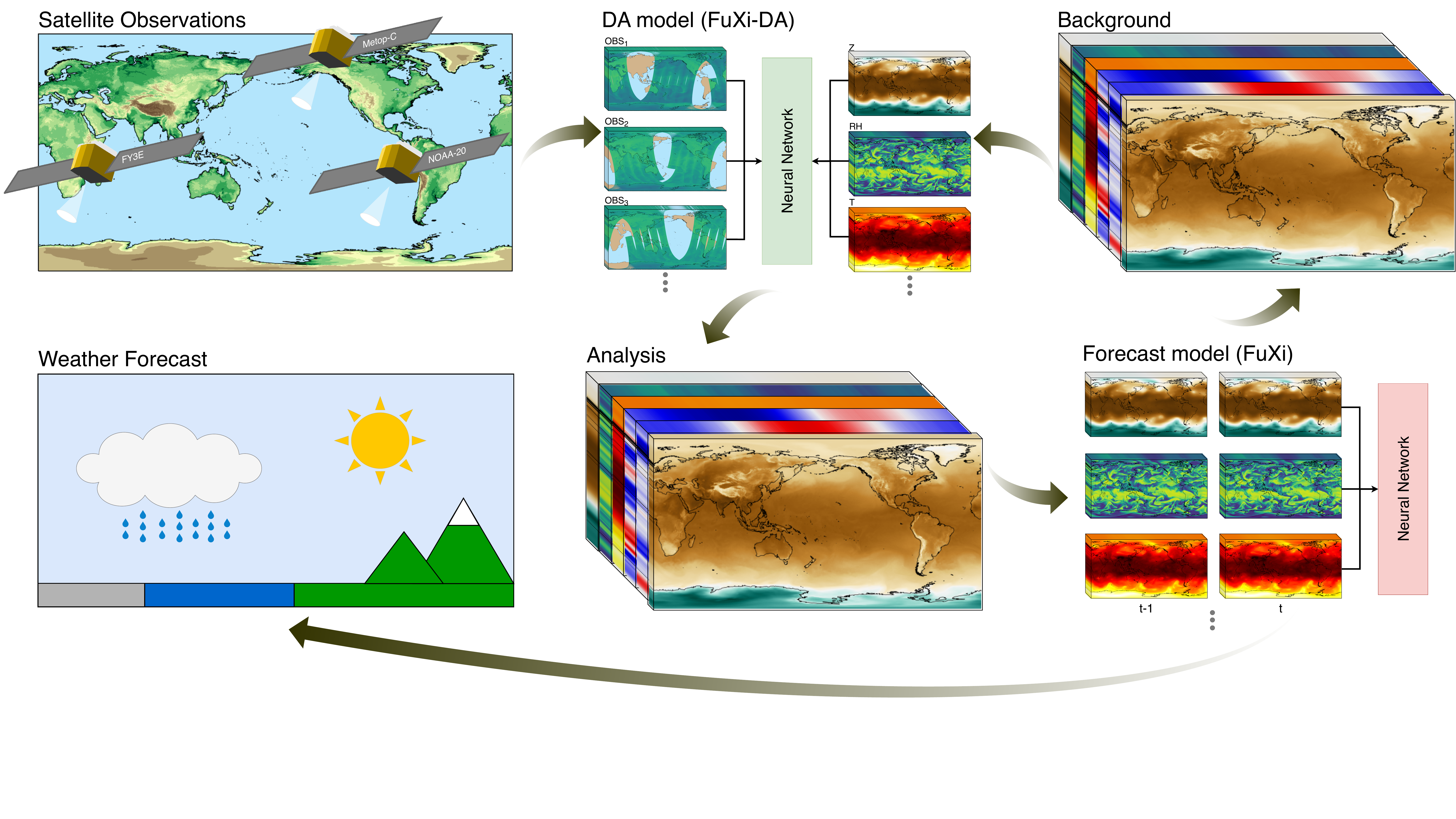}
\caption{Schematic of the FuXi Weather system. Satellite radiance observations are brought in through machine learning data assimilation (DA) coordinated with the FuXi forecast model.}
\label{model_illustration}
\end{figure}

Figure \ref{model_illustration} illustrates FuXi Weather, which generates global weather forecasts every 6 hours.
It has three main components: satellite data preprocessing, DA via FuXi-DA, and forecasting using the FuXi model.
A complete list of variables and abbreviations is provided in Extended Data Table \ref{glossary}.

The preprocessing step addresses the heterogeneity in satellite data across space and time (see Extended Data Fig\ref{satellite_coverage}). 
Nearest-neighbor interpolation maps satellite observations onto FuXi’s grid, and a scalable machine learning architecture handles various data modalities.
This study utilized brightness temperature from five microwave instruments aboard three polar-orbiting satellites (FY-3E, Metop-C, and NOAA-20) and GNSS-RO data \cite{Foelsche2008} (see Extended Data Table \ref{satellite_orbit}), processed using a modified PointPillars \cite{lang2019pointpillars} approach initially designed for three-dimensional point clouds \cite{guo2020deep}.
Missing data are handled using a masking technique, assigning a value of 1 where data are available and 0 otherwise.
Further details are provided in Supplementary Information Sections \ref{satellite_data} and \ref{satellite_data_preprocessing}.

FuXi-DA assimilates the preprocessed data with background forecasts within an 8-hour window to produce analysis fields.
Key improvements include separate processing of different upper-air and surface variables, and a refinement module for improved accuracy (see Supplementary Information Section \ref{fuxi_da_model}). 
The multi-branch architecture handles satellite data and meteorological variables in background forecasts separately, allowing for flexible integration of additional observations.
DA is performed four times per day (at 00, 06, 12, and 18 UTC), using observations from 3 hours before to 4 hours after forecast initialization, generating global analysis fields at $0.25^\circ$ resolution.
The FuXi-Short and FuXi-Medium models then produce 10-day forecasts.

End-to-end training of FuXi Weather optimizes both the analysis and forecasts using ECMWF ERA5 reanalysis data at $0.25^\circ$ resolution as the reference.
To mimic varying operational conditions, FuXi forecasts (initialized with ERA5 data) are randomly sampled across lead times of 6 hours to 5 days and used as background forecasts to train FuXi-DA. 
Owing to the limited amount of satellite data, FuXi-DA is trained on a 1-year dataset (June 1, 2022 - June 30, 2023); this contrasts with the 37-year dataset used to train FuXi models \cite{chen2023fuxi}.
A replay-based incremental learning strategy adapts the system to changes in satellite data quality and availability \cite{harris2001satellite,dee2004variational} (see Supplementary Figs.\ref{Polar_availability} and \ref{GNSS_availability}), retraining FuXi-DA monthly with data from the previous year.
Further details can be found in Supplementary section \ref{effect_incremental_learning}.

The FuXi-Short model is fine-tuned with FuXi-DA analysis fields to reconcile accuracy differences with ERA5 (Supplementary Information section \ref{finetuing}).
During testing, FuXi Weather is initialized with zero values for cyclic DA and forecasting, using a 1-year data spanning from July 1, 2023, to June 30, 2024.

\section{Results}

FuXi Weather operates in a cyclic analysis and forecasting mode, utilizing the full range of available satellite data, referred to as the control run.
Because DA is inherently an ill-posed problem \cite{blum2009data,nichols2007overview} requiring background forecasts to improve analysis accuracy, we trained a version of FuXi-DA without these forecasts to evaluate their role in DA.
Performance was assessed by comparing the accuracy of analysis fields and forecasts globally, as well as regionally in central Africa and northern South America, using ERA5 as the reference. 
The performance of FuXi Weather was compared with that of ECMWF HRES, which was evaluated using the time series of its own 0-hour lead time analysis, HRES-fc0 (see Supplementary Information section \ref{evaluation_method}). This is intentionally favorable to HRES at early lead times, since by definition it starts with low root mean square error (RMSE) and high anomaly correlation coefficient (ACC).
Single observation tests validated DA responses against theoretical expectations while data denial experiments (see Supplementary Information section \ref{data_denial}) evaluated the impact of excluding certain observations.

\subsection{Global analysis fields}
\label{comparison_analysis}

This subsection evaluates the analysis fields of FuXi Weather against 42-hour FuXi forecasts initialized with ERA5, using ERA5 as the benchmark.
Extended Data Fig\ref{analysis_RMSE} presents the globally-averaged, latitude-weighted RMSE for two FuXi Weather configurations: one incorporating background forecasts and one without.
Performance varied markedly across different variables and pressure levels.
The RMSE of analysis fields relative to forecasts was higher at 850 hPa with 300 and 500 hPa, likely owing to the lower information content from satellite observations at lower altitudes.

For relative humidity (${\textrm{R}}$), the analysis of FuXi Weather outperformed forecasts at 300 and 500 hPa, but had a slightly higher RMSE at 850 hPa.
For temperature (T), geopotential (Z), and wind components (U and V), the RMSE values were comparable to those of forecasts at higher altitudes but were consistently higher at 850 hPa.
Although satellite data primarily capture temperature and moisture information, their assimilation also improves wind fields through the dynamic relationship between wind, temperature, and moisture-.
Wind can be inferred from temperature gradients (geostrophic balance) and the movement of atmospheric constituents, such as humidity, known as the "generalized tracer effect" \cite{geer2018all}.

The inclusion of background forecasts markedly enhanced the accuracy of the analysis fields of FuXi Weather, as evidenced by lower RMSE values.
This highlights the crucial role of background forecasts in DA, which is ill-posed without prior information.
Both configurations of FuXi Weather showed similar trends over time, but the analysis without background forecasts exhibited more pronounced error peaks, especially when some satellite data were missing (see Supplementary Figs.\ref{Polar_availability} and \ref{GNSS_availability}), underscoring the stabilizing effect of background forecasts.

The shaded area in the figure represents variations across initialization times; this was more pronounced in forecasts.
Forecasts initialized at 00/12 UTC consistently outperformed those at 06/18 UTC, likely because the 12-hour observation windows of ERA5 (09-21 UTC and 21-09 UTC) \cite{hersbach2020era5} provide 9 hours of look-ahead time for 00/12 UTC but only 3 hours for 06/18 UTC \cite{lam2022graphcast}.
In contrast, the analysis fields of FuXi Weather exhibited minimal variation across initialization times, likely owing to its consistent 8-hour assimilation window.

\subsection{Global weather forecasts}

The primary criterion for evaluating an end-to-end weather forecasting system is its ability to provide reliable and accurate forecasts in a cyclic analysis and forecasting mode.
This subsection evaluates the performance of 6-hourly cycle forecasts in FuXi Weather using two types of analysis fields generated by FuXi-DA: one incorporating background forecasts and one without.
The forecasts are compared with those of ECMWF HRES.

\begin{figure}[h]
\centering
\includegraphics[width=1\textwidth]{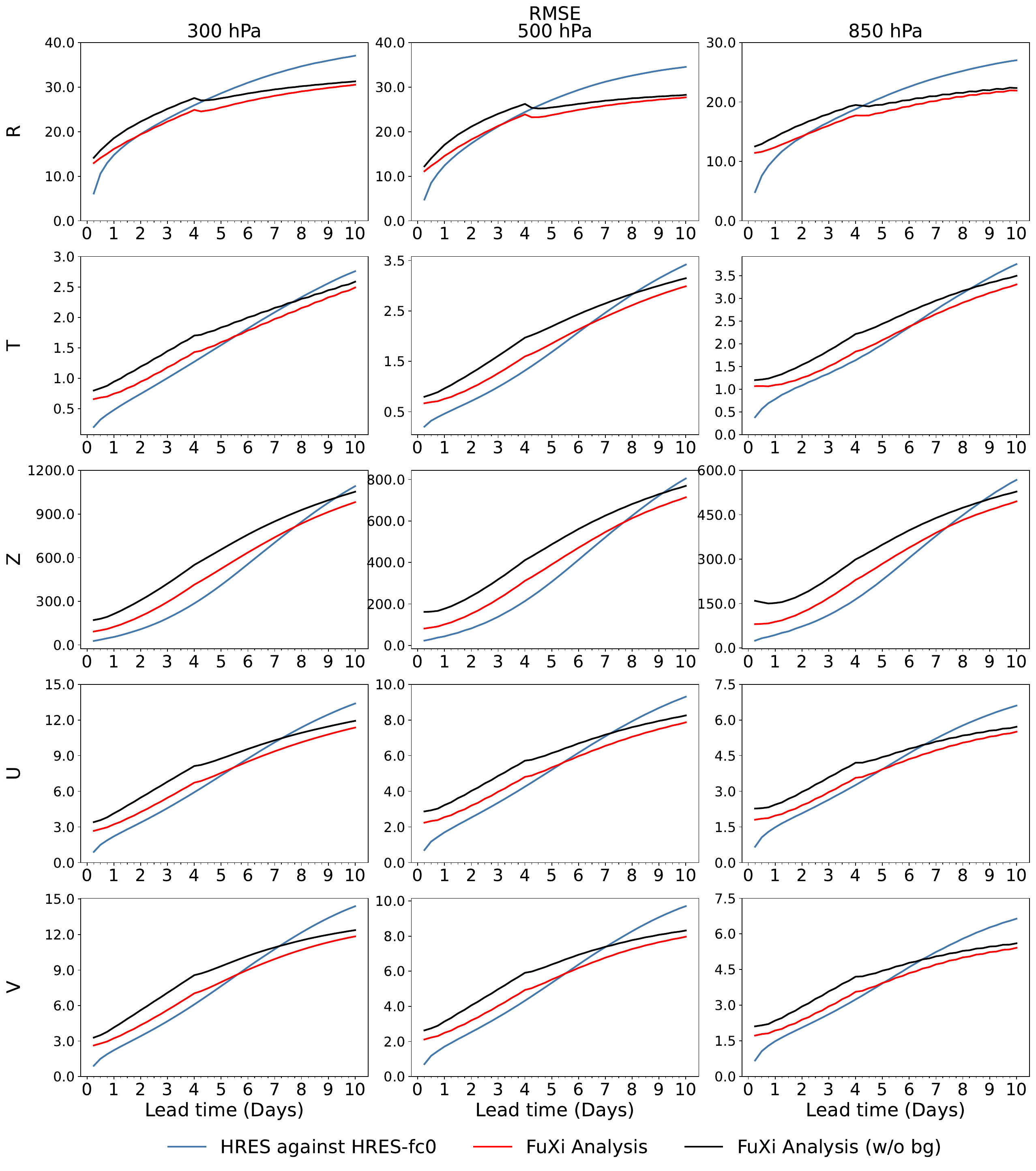}
\caption{Comparison of forecast performance among various models over a 1-year testing period, spanning from July 03, 2023, to June 30, 2024. The figure presents the globally-averaged, latitude-weighted root mean square error (RMSE) for forecasts generated by the FuXi model and ECMWF HRES (blue) in 10-day forecasts. FuXi forecasts are initialized using analysis fields produced by FuXi-DA with (red) and without (black) background forecasts. The analysis includes 5 variables: relative humidity (${\textrm{R}}$), temperature (${\textrm{T}}$), geopotential (${\textrm{Z}}$), u component of wind (${\textrm{U}}$), and v component of wind (${\textrm{V}}$), at three pressure levels (300 hPa, 500 hPa, and 850 hPa). The five rows and three columns correspond to five variables and three pressure levels, respectively. As ECMWF HRES is evaluated against its own initialization time series, it inherently exhibits lower RMSE in early lead times.}
\label{fig2}
\end{figure}

Figure \ref{fig2} shows the globally-averaged, latitude-weighted RMSE as a function of forecast lead times over a 10-day period.
FuXi Weather forecasts initialized with analysis fields that included background forecasts consistently demonstrated lower RMSE values than those without, in agreement with Extended Data Fig\ref{analysis_RMSE}.

Although FuXi Weather initially had higher RMSE values than those of ECMWF HRES, it outperformed ECMWF HRES after a variable-specific lead time, which varied according to the variable and pressure level.
For R, FuXi Weather outperformed ECMWF HRES at lead times of 2.00, 3.25, and 2.25 days for 300, 500, and 850 hPa, respectively.
For T, Z, U, and V, the critical lead times were later owing to the lower accuracy of their corresponding analysis fields.
For Z, these times were 8.00, 7.75, and 7.50 days at 300, 500, and 850 hPa, respectively.

Extended Data Fig\ref{SI_ACC} shows similar trends for the globally-averaged, latitude-weighted $\textrm{ACC}$.
FuXi Weather forecasts initialized without background forecasts performed worse, as expected.
However, Fuxi Weather forecasts initialized with analysis incorporating background forecasts, though initially less accurate than ECMWF HRES, improved over time and eventually achieved higher ACC values across all examined variables.
Using an $\textrm{ACC}$ threshold of 0.6 to define a skillful forecast, Extended Data Fig\ref{skillful_lead} compares skillful lead times.
FuXi Weather extended skillful lead times for seven out of 15 variables,  matching ECMWF HRES for six others.
For example, for ${\textrm{Z500}}$, FuXi Weather extended the skillful lead time from the ECMWF HRES value of 9.25 days to 9.50 days for forecasts initialized with background forecasts (forecasts initialized without background forecasts show a skillful lead time of only 8.25 days).
Additional forecast comparisons, including spatial RMSE distributions, are provided in Supplementary Information section \ref{comparison_forecast}.

\FloatBarrier

\subsection{Forecast performance in central Africa}

Routine evaluations of NWP systems often prioritize global or broad regional performance metrics \cite{ECMWF2021}, overlooking substantial regional disparities with socioeconomic implications.
Forecast accuracy tends to be lower in low-income countries, largely owing to limited investment in weather observation infrastructure.
This issue is especially concerning for many low-income countries, in which agriculture is a major economic sector that relies heavily on accurate weather forecasts.
Climate change further exacerbates weather-related risks, disproportionately affecting vulnerable populations with low adaptive capacities in these countries.
Therefore, improving forecast accuracy in underserved regions, especially Africa, is crucial to enhance climate resilience \cite{pachauri2014climate,Assefa2019}.

\begin{figure}[h]
\centering
\includegraphics[width=1\textwidth]{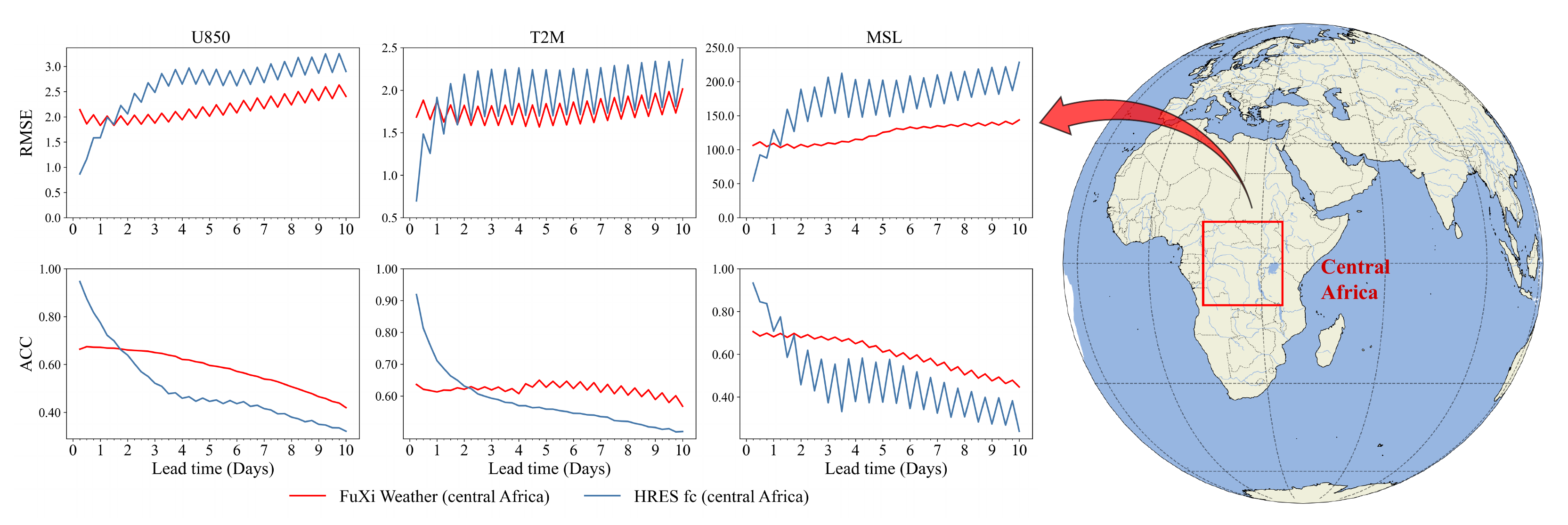}
\caption{Comparison of forecast performance over central Africa during the 1-year testing period from July 03, 2023, to June 30, 2024. Central Africa is defined as the region spanning 15$^\circ$ E to 35$^\circ$ E in longitude and 10$^\circ$ N to 10$^\circ$ S in latitude. Rows 1 and 2 show the root mean square error (RMSE), anomaly correlation coefficient (ACC) for forecasts generated by FuXi Weather (red) and ECMWF HRES (blue). FuXi Weather is initialized using analysis fields produced by FuXi-DA incorporating background forecasts. This figure includes three variables: 850 hPa u wind component ($\textrm{U850}$), 2-meter temperature ($\textrm{T2M}$), and mean sea level pressure ($\textrm{MSL}$). As ECMWF HRES is evaluated against its own initialization time series, it inherently exhibits lower RMSE and higher ACC in early lead times.}
\label{Africa_forecast}
\end{figure}

This subsection compares the performance of FuXi Weather with that of ECMWF HRES in underserved regions, focusing on central Africa.
As shown in Figure \ref{Africa_forecast}, FuXi Weather consistently outperformed ECMWF HRES in forecasting 850 hPa u wind component ($\textrm{U850}$), 2-meter temperature ($\textrm{T2M}$), and mean sea level pressure ($\textrm{MSL}$) at almost all lead times, except for a brief initial period.
FuXi Weather demonstrated smaller RMSE and higher ACC, with ACC values consistently exceeding 0.6 throughout the 10-day forecast period, indicating meaningful predictive skill.
In contrast, ECMWF HRES showed skillful lead times at only about 2 days.
As shown in Extended Data Fig\ref{africa_more_forecast}, FuXi Weather also had lower mean bias error (MBE) and smaller standard deviations of errors (STD$_\textrm{ERROR}$), suggesting both reduced systematic and random errors compared to ECMWF HRES.

Notably, FuXi Weather provided superior forecasts for surface variables without assimilating surface-based observations, demonstrating its strength in utilizing satellite data in regions with limited observational infrastructure.
Further analysis (see Supplementary Information section \ref{comparison_forecast}) revealed that FuXi Weather also performed ECMWF HRES in more data-sparse regions, such as tropical oceans and South America, although it was less competitive in areas with dense surface observations.
In central Africa, where observational networks are sparse, the efficient use of satellite data by FuXi Weather closed the performance gap with ECMWF HRES, resulting in superior forecasts.

Extended Data Fig\ref{valid_time_africa} illustrates 10-day forecast time series for two randomly selected initialization times, while Extended Data Fig\ref{3day_lead_time_africa} presents forecasts at a fixed 3-day lead time.
Both figures confirm that FuXi Weather more closely aligns with its benchmark than ECMWF RHES, reinforcing the results in Figure \ref{Africa_forecast}.
Additionally, Extended Data Fig\ref{south_america_forecast} shows that FuXi Weather outperforms ECMWF HRES, particularly for T2M and MSL over northern South America, where observational coverage is also sparse relative to Europe or North America.

These findings suggest that FuXi Weather can deliver more accurate forecasts using an substantially less data than traditional NWP systems, offering a cost-effective, satellite-based solution to improve forecasts in regions lacking observational infrastructure.

\FloatBarrier

\subsection{Physical consistency of analysis increments}

FuXi Weather, as a data-driven machine learning system, does not inherently encode prior physical knowledge of atmospheric processes.
This subsection examines the impact of assimilating a single observation on analysis increments and assesses whether these increments align with theoretical expectations.

Two FuXi-DA runs were conducted: the first using a 6-hour forecast with original observations, and the second with a perturbation introduced at a specific observation location.
The differences between these two runs reflected the analysis increments caused by the perturbation (details in Supplementary Information section \ref{single_obs_method}).
The first run, initialized at 06 UTC on July 24, 2023, assimilated all available data to generate the analysis.
In the second run, a +5 K perturbation was introduced into the NOAA-20 ATMS observation at 19.9$^\circ$ N, 125.5$^\circ$ E (marked as a purple dot in Extended Data Fig\ref{single_obs_TC_obs}), near Typhoon Doksuri over the ocean. 
The impact of this perturbation was evaluated by comparing outputs from both runs.

\begin{figure}[h]
\centering
\includegraphics[width=\textwidth]{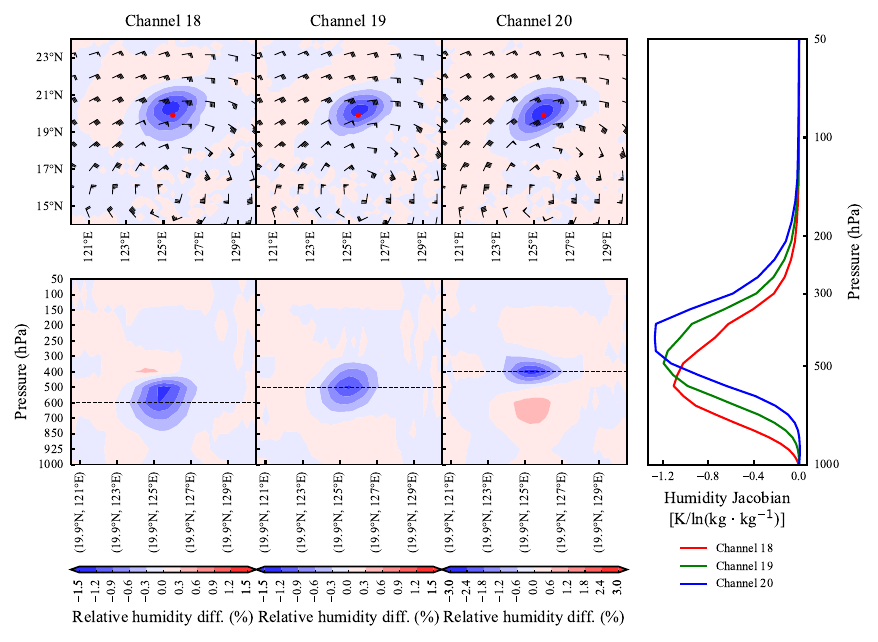}
\caption{Analysis increment resulting from the introduction of a 5 K perturbation to the NOAA-20 ATMS observation at a selected location, based on the background field for 06 UTC on July 24, 2023. The perturbation, located over the ocean near Typhoon Doksuri at 19.9$^\circ$ N, and 125.5$^\circ$ E (red dot), is introduced at 05 UTC, 1 hour before the analysis time. The two rows show, in the left panel, the horizontal spatial distribution of the analysis increment for channels 18 to 20 at 600, 500, and 400 hPa, with wind fields overlaid, as well as the corresponding vertical distribution along the same west-east cross-section. The dashed lines on figure of the second and fourth rows indicate the pressure levels for the horizontal spatial distribution. The right panel shows the Jacobian functions for 3 humidity channels derived from ATMS aboard NOAA-20. The atmospheric profile is based on the US Standard Atmosphere, and radiative transfer calculations are performed using RTTOV version 13.2.}
\label{single_obs_TC_selected}
\end{figure}

Figure \ref{single_obs_TC_selected} shows the horizontal and vertical distributions of analysis increments across three humidity channels.
The spatial patterns of these increments aligned with the radiative transfer theory: an increase in brightness temperature corresponds to a decrease in humidity, resulting in less radiation absorption \cite{sieglaff2009inferring}.
The vertical distribution showed progressive increases in the peak heights of the Jacobian functions for channels 18, 19, and 20, matched by corresponding increases in the peak heights of the humidity increments.
This pattern suggests that the DA system effectively captures the varying detection altitudes of these channels.
Additionally, flow-dependent characteristics were observed in the humidity field. 
The perturbation introduced at 05 UTC, 1 hour before the analysis, led to larger increment values downstream of the wind fields, indicating that the perturbation propagated with the wind.

In summary, FuXi Weather effectively captures the horizontal and vertical dependencies of analysis increments on satellite observations without explicitly incorporating prior knowledge.
Data denial experiments (Supplementary Information Section \ref{data_denial}) further confirm FuXi Weather's physical consistency with satellite observations, while additional tests demonstrate the robustness of its performance.

\FloatBarrier

\section{Discussion} 

In this paper, we introduce FuXi Weather, an end-to-end machine learning based weather forecasting system that performs DA and forecasting on a 6-hourly cycle using raw satellite observations.
To the best of our knowledge, FuXi Weather is the first system to achieve DA and forecasting with satellite observations across all grids, surfaces, channels, and sky conditions.
It is also the first machine learning system to match the global forecasting performance of state-of-the-art ECMWF HRES, consistently outperforming the latter in observation-sparse regions such as central Africa and northern South America.
Importantly, FuXi Weather extends the skillful forecast lead time achieved by ECMWF HRES for many regions, while using considerably fewer observations.
Single observation tests confirm that DA responses align with theory while data denial experiments demonstrate the robustness of the system, with only moderate error increases when specific observations are excluded.
Additionally, the computational efficiency and reduced complexity of FuXi Weather, when compared with traditional NWP systems, make it a cost-effective solution for improving operational forecasts in regions with limited land-based observations, thus enhancing climate resilience.

Despite these promising results, several challenges remain.
While FuXi Weather improves lead times for some variables, its short-term forecast accuracy still requires improvement.
This limitation likely arises from the reliance on data from just five microwave instruments aboard the thee polar-orbiting satellites (FY-3E, Metop-C, and NOAA-20), and a few small number of GNSS receivers; this is far fewer than the 90 satellite instruments used by the ECMWF \cite{ECMWF_observation}.
Incorporating additional satellite data, along with radiosonde soundings, and surface, marine, and radar observations, could improve the performance, particularly for surface variables.
FuXi Weather's design simplifies further development by eliminating the need for observation operators, adjoint models, or the estimation of observation and background error covariance matrices, which are computationally intensive and require specialized expertise.
Integrating ensemble-based \cite{Jeffrey2008,zhong2024fuxiens} DA systems offers opportunities to further enhance performance.
Moreover, machine learning based systems such as FuXi Weather, could foster interdisciplinary collaboration between meteorologists and machine learning scientists, overcoming traditional barriers posed by systems developed in languages such as Fortran \cite{Ott2020,zhong2023machine}.

Scaling FuXi Weather to accommodate larger models and datasets will be essential as more observations are integrated.
Optimal hybrid parallelization strategies \cite{Song2019,Rasley2020,fan2021dapple} that combine pipeline parallelism and data parallelism, could enable efficient training with increased observations.
The flexible, multi-branch architecture of FuXi Weather supports scalable implementation for additional observational data.
Data denial experiments also suggest that excluding less informative satellite data could improve efficiency without compromising accuracy.

The development of FuXi Weather builds on the foundation established by traditional NWP systems and ERA5 reanalysis, which serve as critical benchmarks. However, since ERA5 blends observations with a physics-based model, FuXi Weather inherits both its strengths and weaknesses, owing to its training with ERA5 \cite{Maddy2024}.
This dependency may limit its ability to fully surpass traditional models.
Future work should focus on using direct observations for training and evaluation, aiming to reduce the reliance on reanalysis datasets and develop a more independent and robust forecasting system \cite{Rasp2018,mcnally2024}.

\section*{Data Availability Statement}
The ERA5 reanalysis data are accessible through the Copernicus Climate Data Store at \url{https://cds.climate.copernicus.eu/}.
ECMWF HRES forecasts can be retrieved from \url{https://apps.ecmwf.int/archive-catalogue/?type=fc&class=od&stream=oper&expver=1}.
Satellite data can be obtained from the portal of the National Satellite Meteorological Center at  \url{http://satellite.nsmc.org.cn/PortalSite/Data/DataView.aspx}.

\section*{Code Availability Statement}
The FuXi model is available on Github at \url{https://github.com/tpys/FuXi}. The source code for FuXi Weather used in this study is hosted in a restricted Google Drive folder, accessible via the following link \href{https://forms.gle/UJvE9MXYyJ5L1HUq7}{FuXi-Weather}. Owing to the importance of FuXi Weather and its associated code, we used password protection for the Google Drive folder link through a Google Form. 
For inquiries and access to the Google Drive link kindly reach out to Professor Li Hao at the following email address: \url{lihao_lh@fudan.edu.cn}.

\section*{Acknowledgements}
We extend our sincere gratitude to the researchers at the ECMWF for their invaluable contributions to the collection, archiving, dissemination, and maintenance of the ERA5 reanalysis dataset and ECMWF HRES forecasts.
We also wish to acknowledge the National Satellite Meteorological Center of the China Meteorological Administration for providing the satellite data.
Finally, We would like to acknowledge the helpful comments made by Kan Dai, Yong Cao and Xianhuang Xu from the National Meteorological Information Center of the China Meteorological Administration.

\section*{Competing interests}
The authors declare no competing interests.

\noindent


\bibliography{refs}


\end{CJK*}

\clearpage



\setcounter{figure}{0}    
\setcounter{table}{0}    

\renewcommand{\figurename}{Extended Data Fig}
\renewcommand{\tablename}{Extended Data Table}

\begin{figure}[h]
    \centering
    \includegraphics[width=\linewidth]{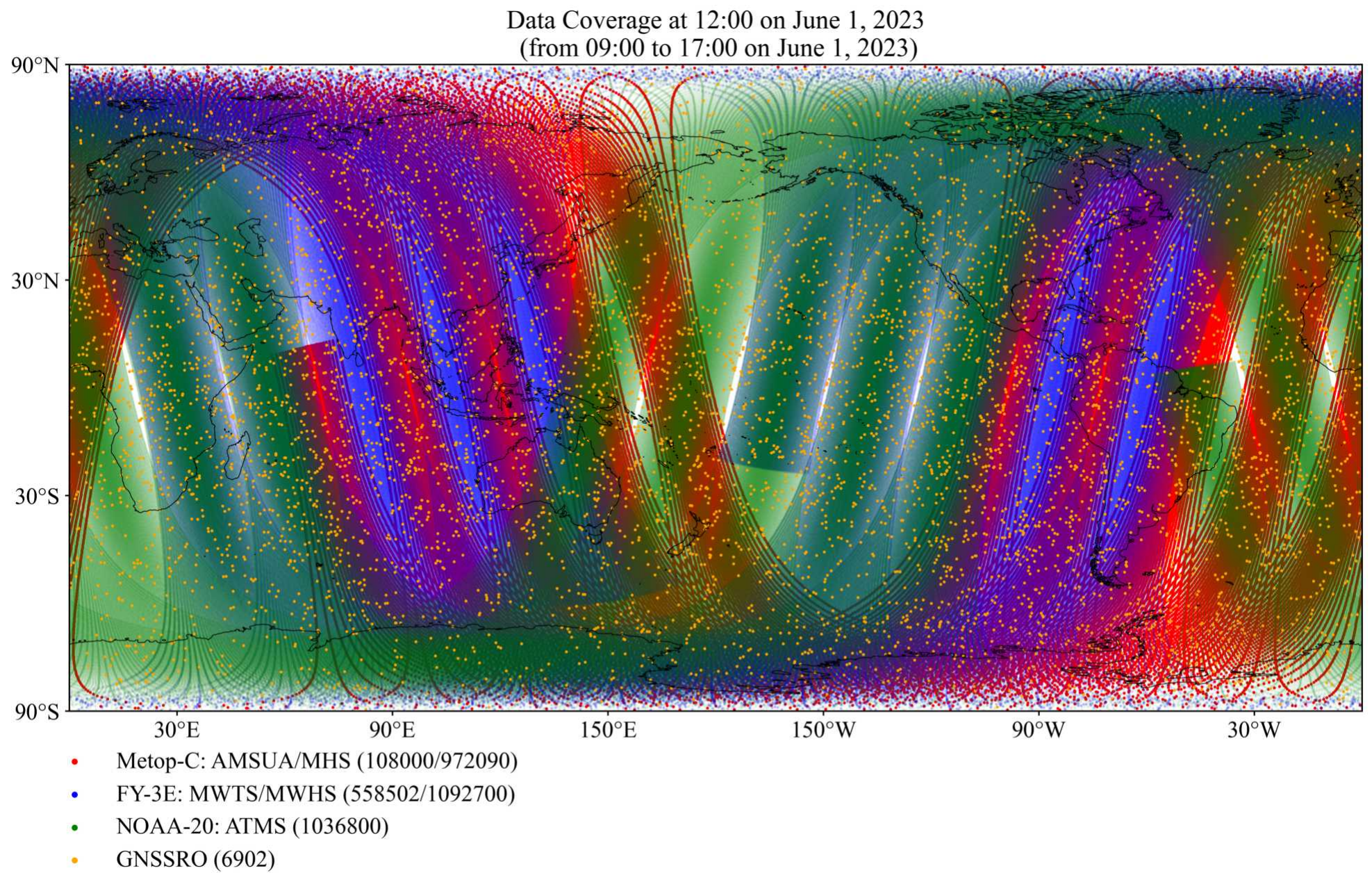}
    \caption{Typical data coverage from observations collected by FengYun-3E (blue), Meteorological Operational Polar Satellite - C (Metop-C) (red), National Oceanic and Atmospheric Administration - 20 (NOAA-20) (green), and Global Navigation Satellite System (GNSS) radio occultation (RO) (yellow). This represents data spanning the period from 3 hours before to 4 hours after 12 UTC on June 1, 2023. These data are utilized to generate analysis fields for 12 UTC on the same date.}
    \label{satellite_coverage}        
\end{figure}
\FloatBarrier

\begin{figure}[h]
\centering
\includegraphics[width=\linewidth]{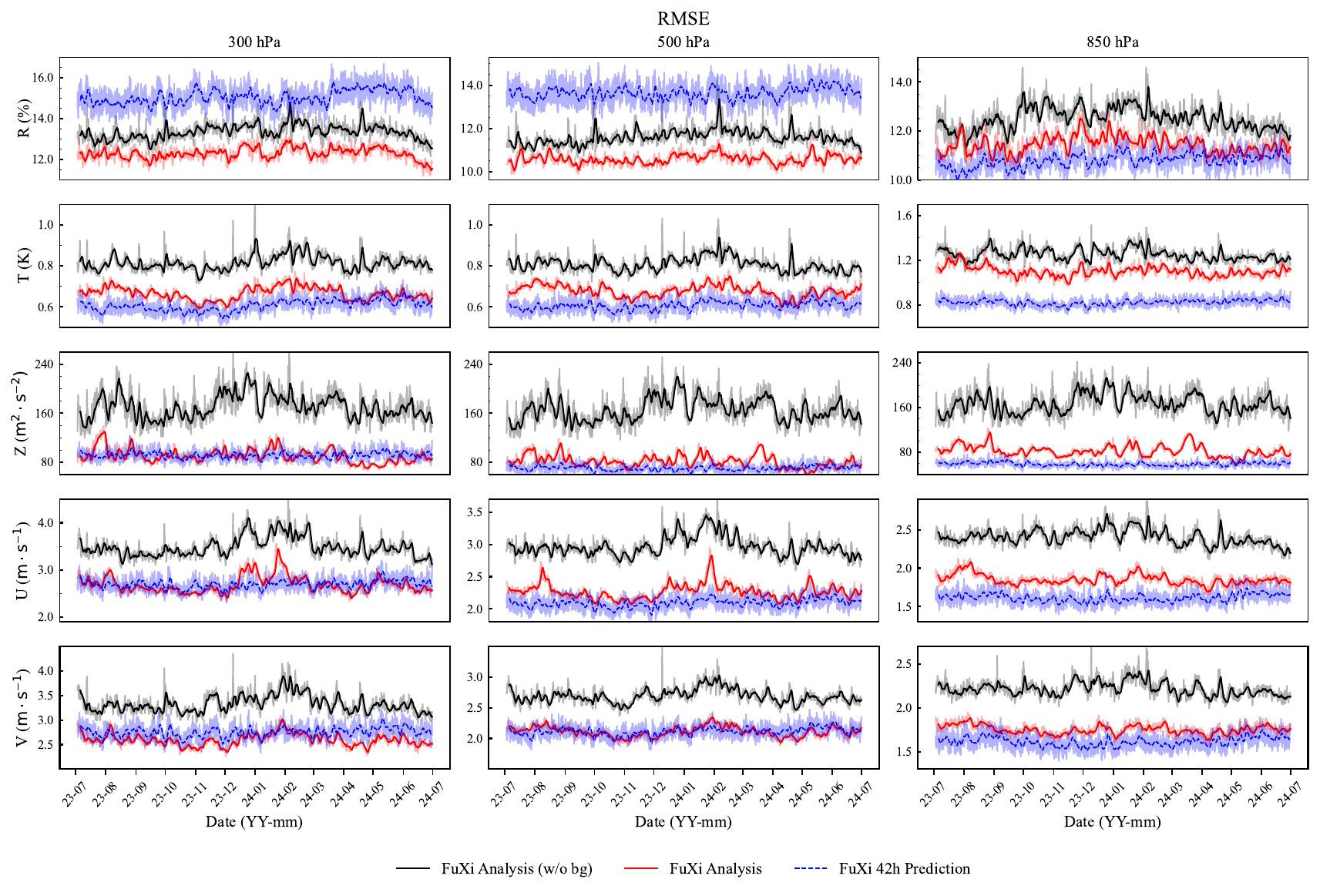}
\caption{Comparison of analysis fields produced by FuXi Weather and 42-hour FuXi forecasts over a 1-year testing period from July 03, 2023, to June 30, 2024. The time series show the globally-averaged, latitude-weighted root mean square error (RMSE) for the analysis fields of FuXi Weather with (solid red lines) and without (solid black lines) background (bg) forecasts, along with 42-hour FuXi forecasts (dashed blue lines). The comparison includes five variables: relative humidity (${\textrm{R}}$), temperature (${\textrm{T}}$), geopotential (${\textrm{Z}}$), u component of wind (${\textrm{U}}$), and v component of wind (${\textrm{V}}$), at three pressure levels (300 hPa, 500 hPa, and 850 hPa). The five rows and three columns correspond to five variables and three pressure levels, respectively. To improve clarity, the original data are shown with reduced opacity, while solid lines represent smoothed values using a 12-point moving average.}
\label{analysis_RMSE}
\end{figure}
\FloatBarrier

\begin{figure}[h]
\centering
\includegraphics[width=1\textwidth]{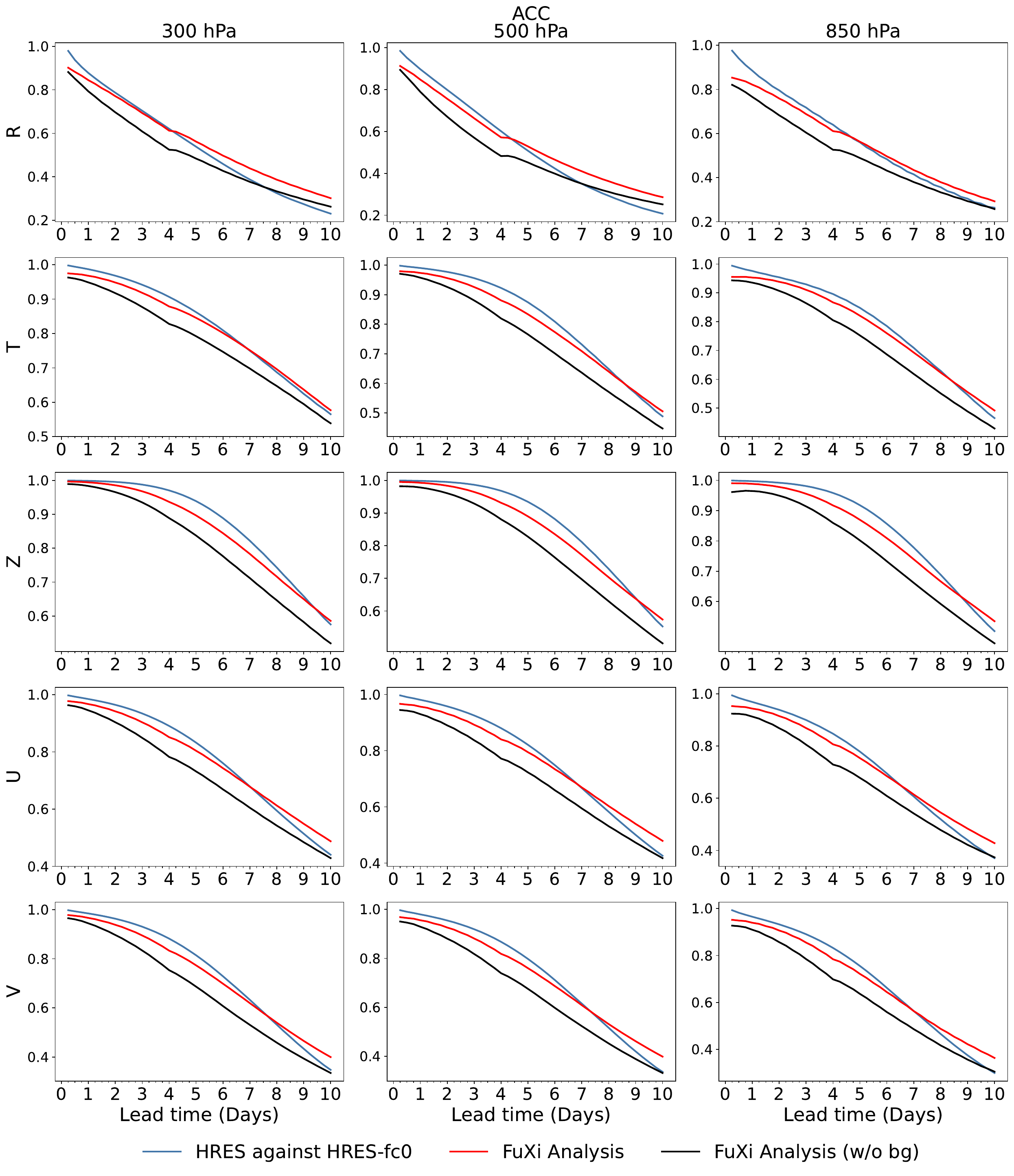}
\caption{Comparison of forecast performance among various models over a 1-year testing period, spanning July 03, 2023 - June 30, 2024. The figure presents the globally-averaged, latitude-weighted anomaly correlation coefficient ($\textrm{ACC}$) for forecasts generated by the FuXi model and ECMWF HRES in 10-day forecasts. FuXi forecasts are initialized using analysis fields produced by FuXi-DA with (red) and without (green) background (bg) forecasts. The analysis includes five variables: relative humidity (${\textrm{RH}}$), temperature (${\textrm{T}}$), geopotential (${\textrm{Z}}$), u component of wind (${\textrm{U}}$), and v component of wind (${\textrm{V}}$), at three pressure levels (300 hPa, 500 hPa, and 850 hPa). The five rows and three columns correspond to five variables and three pressure levels, respectively. As ECMWF HRES is evaluated against its own initialization time series, it inherently exhibits higher ACC in early lead times.}
\label{SI_ACC}
\end{figure}
\FloatBarrier

\begin{figure}[h]
    \centering
    \includegraphics[width=\linewidth]{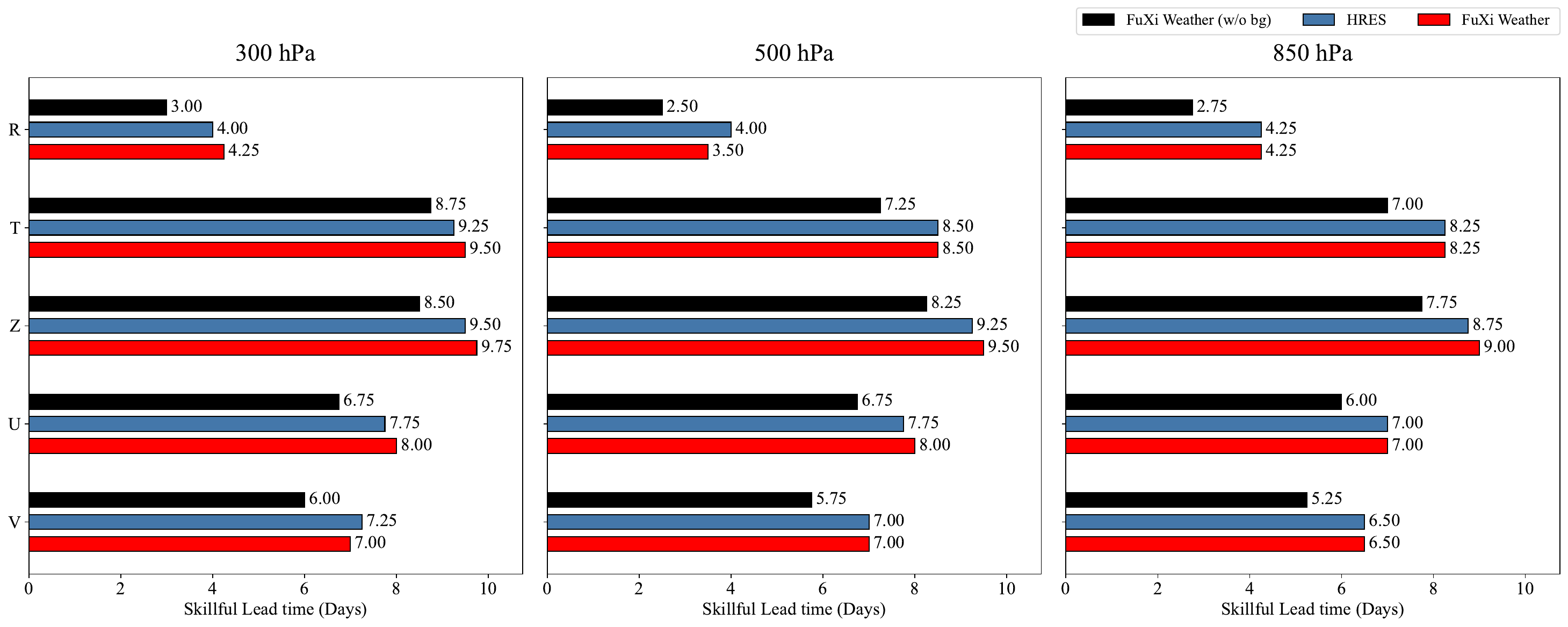}
    \caption{Skillful forecast lead time comparisons with an anomaly correlation coefficient ($\textrm{ACC}$) value of 0.6 as the threshold. Skillful forecast lead times of ECMWF HRES and FuXi Weather for five variables: relative humidity (${\textrm{R}}$), temperature (${\textrm{T}}$), geopotential (${\textrm{Z}}$), u component of wind (${\textrm{U}}$), and v component of wind (${\textrm{V}}$), at three pressure levels (300 hPa, 500 hPa, and 850 hPa), using all testing data over a 1-year testing period, spanning July 03, 2023 - June 30, 2024. The five rows and three columns correspond to five variables and three pressure levels, respectively.}
    \label{skillful_lead}    
\end{figure}

\begin{figure}[h]
\centering

\includegraphics[width=1\textwidth]{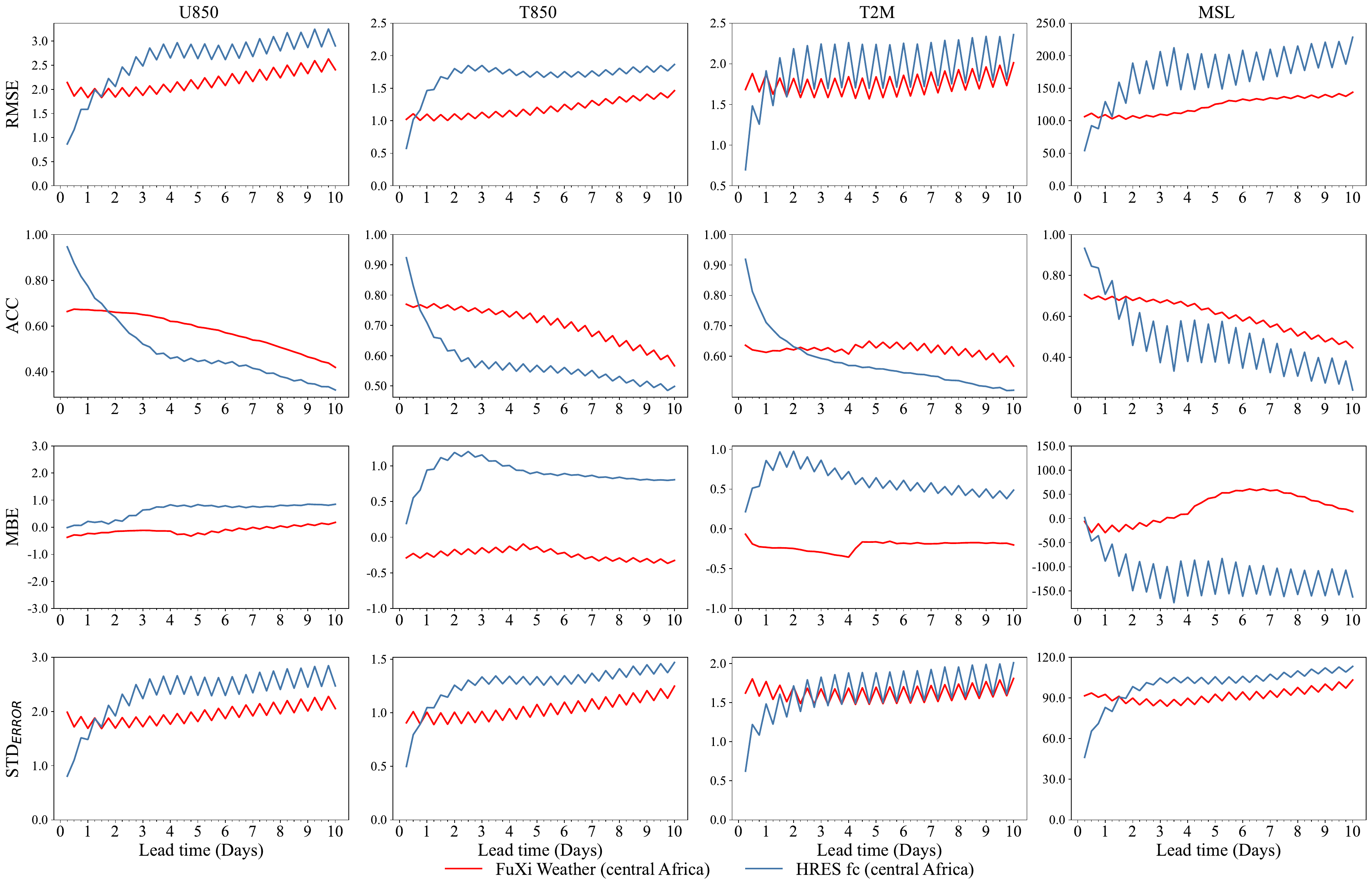}
\caption{Comparison of forecast performance over central Africa during the 1-year testing period from July 03, 2023, to June 30, 2024. Central Africa is defined as the region spanning 15$^\circ$ E to 35$^\circ$ E in longitude and 10$^\circ$ N to 10$^\circ$ S in latitude. Rows 1 to 4 show the root mean square error (RMSE), anomaly correlation coefficient (ACC), mean bias error (MBE), and standard deviation of errors (STD$_\textrm{ERROR}$) for forecasts generated by FuXi Weather (red) and ECMWF HRES (blue). FuXi Weather is initialized using analysis fields produced by FuXi-DA incorporating background forecasts. This figure includes four variables: 850 hPa u wind component ($\textrm{U850}$), 850 hPa temperature ($\textrm{T850}$), 2-meter temperature ($\textrm{T2M}$), and mean sea level pressure ($\textrm{MSL}$).}
\label{africa_more_forecast}
\end{figure}
\FloatBarrier

\begin{figure}[h]
\centering
\includegraphics[width=\linewidth]{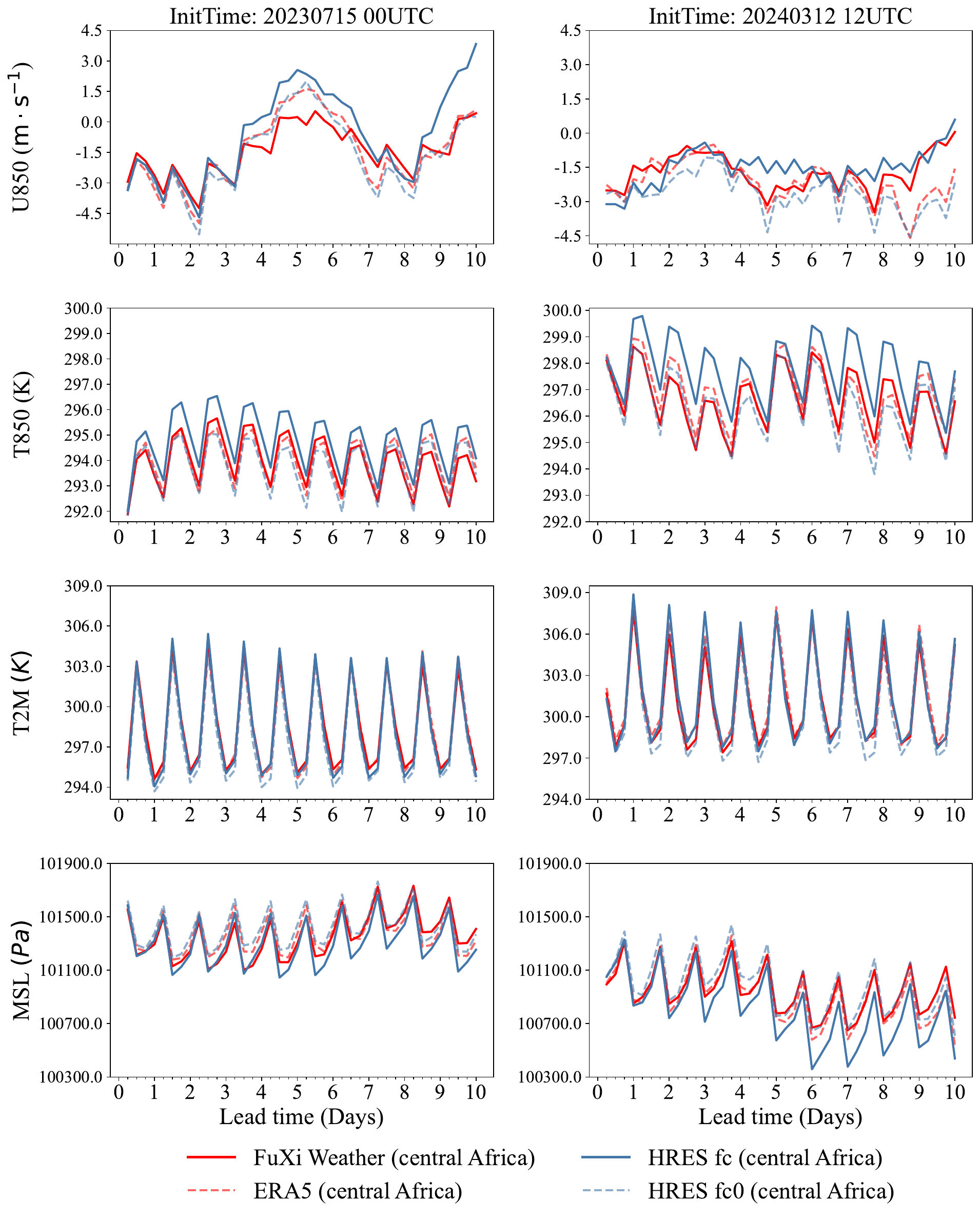}
\caption{Example time series of 10-day forecasts averaged over central Africa for two initialization times: July 15, 2023, 00 UTC (left column), and March 12, 2024, 12 UTC (right column). Central Africa is defined as the region spanning 15$^\circ$ E to 35$^\circ$ E in longitude and 10$^\circ$ N to 10$^\circ$ S in latitude. The forecasts include FuXi Weather (solid red lines) and its benchmark ERA5 (dashed red lines), as well as ECMWF HRES (solid blue lines) and its benchmark HRES-fc0 (dashed blue lines). Rows 1 to 4 show the time series for 850 hPa u wind component (U850, first row), 850 hPa temperature (T850, second row), 2-meter temperature (T2M, third row), and mean sea level pressure (MSL, fourth row).}
\label{valid_time_africa}
\end{figure}
\FloatBarrier

\begin{figure}[h]
\centering
\includegraphics[width=\linewidth]{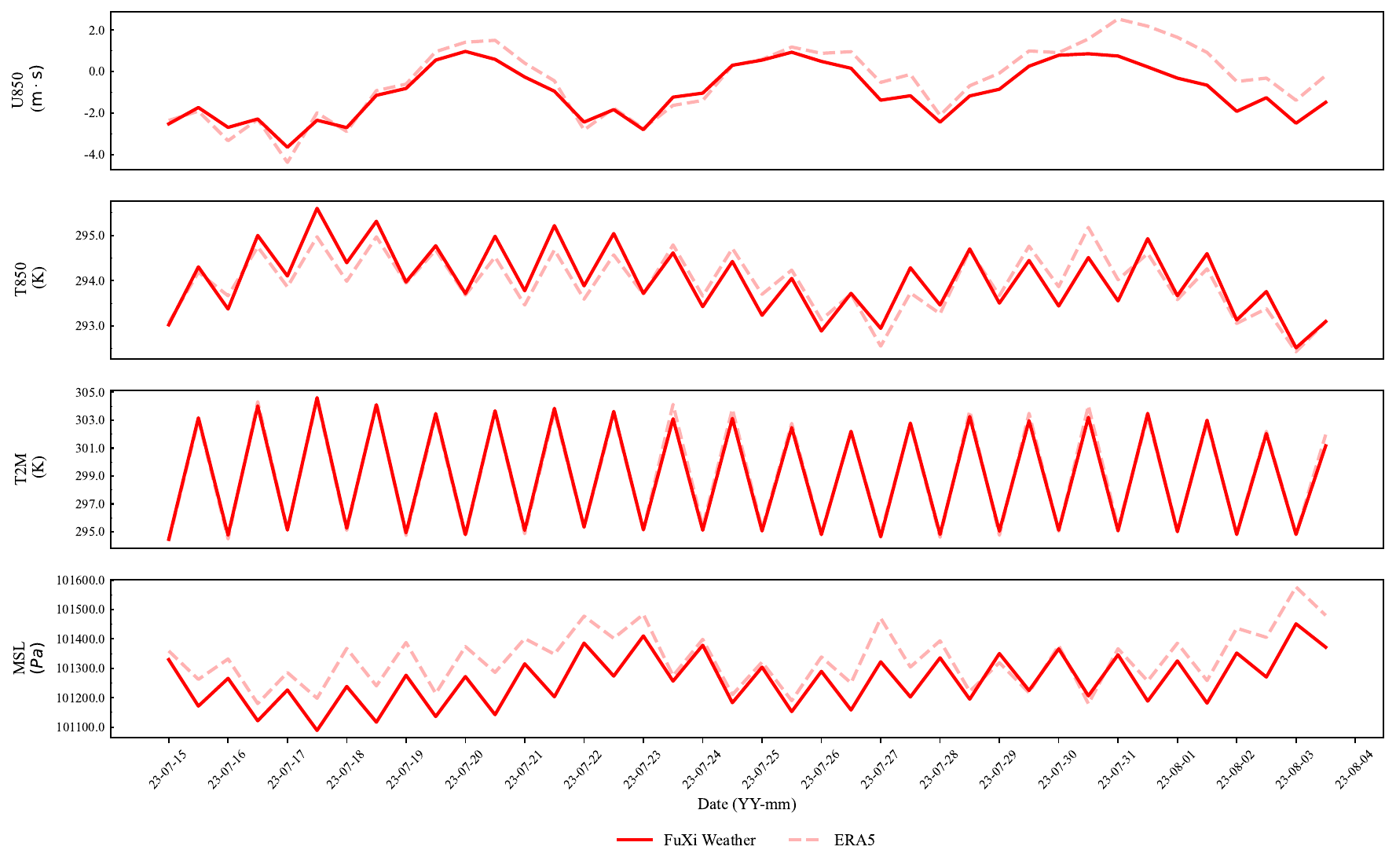}
\caption{Example time series of 3-day lead time forecasts averaged over central Africa from July 15, 2023, to August 3, 2023. Central Africa is defined as the region spanning 15$^\circ$ E to 35$^\circ$ E in longitude and 10$^\circ$ N to 10$^\circ$ S in latitude. The forecasts include FuXi Weather (solid red lines) and its benchmark ERA5 (dashed red lines). Rows 1 to 4 display the time series for 850 hPa u wind component (U850, first row), 850 hPa temperature (T850, second row), 2-meter temperature (T2M, third row), and mean sea level pressure (MSL, fourth row).}
\label{3day_lead_time_africa}
\end{figure}
\FloatBarrier

\begin{figure}[h]
\centering

\includegraphics[width=1\textwidth]{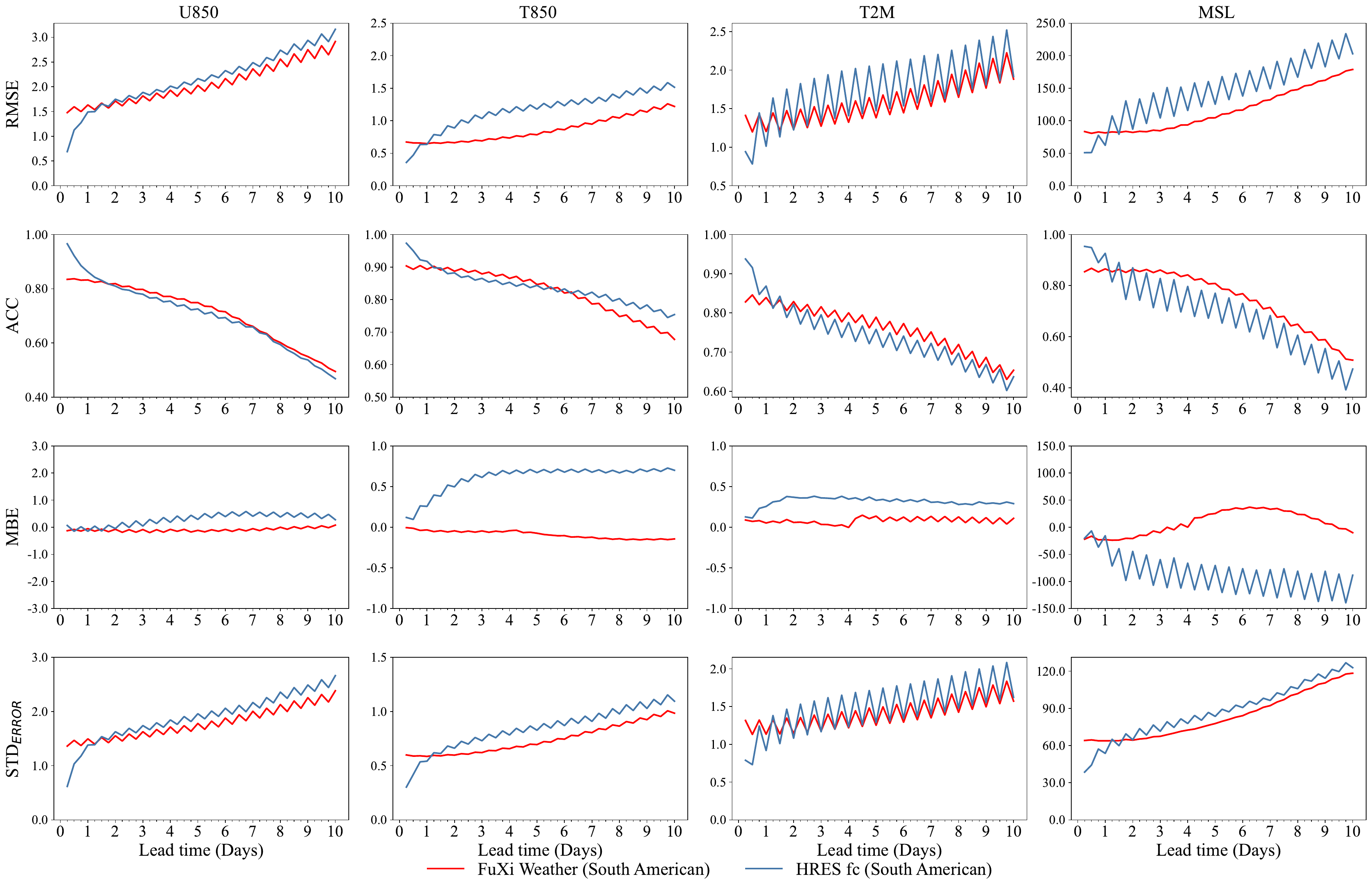}
\caption{Comparison of forecast performance over northern South America during the 1-year testing period from July 03, 2023, to June 30, 2024. Northern South America is defined as the region spanning 70$^\circ$ W to 50$^\circ$ W in longitude and 0$^\circ$ S to 20$^\circ$ S in latitude. Rows 1 to 4 show the root mean square error (RMSE), anomaly correlation coefficient (ACC), mean bias error (MBE), and standard deviation of errors (STD$_\textrm{ERROR}$) for forecasts generated by FuXi Weather (red) and ECMWF HRES (blue). FuXi Weather is initialized using analysis fields produced by FuXi-DA incorporating background forecasts. This figure includes four variables: 850 hPa u wind component ($\textrm{U850}$), 850 hPa temperature ($\textrm{T850}$), 2-meter temperature ($\textrm{T2M}$), and mean sea level pressure ($\textrm{MSL}$).}
\label{south_america_forecast}
\end{figure}
\FloatBarrier

\begin{figure}[h]
\centering
\includegraphics[scale=0.55]{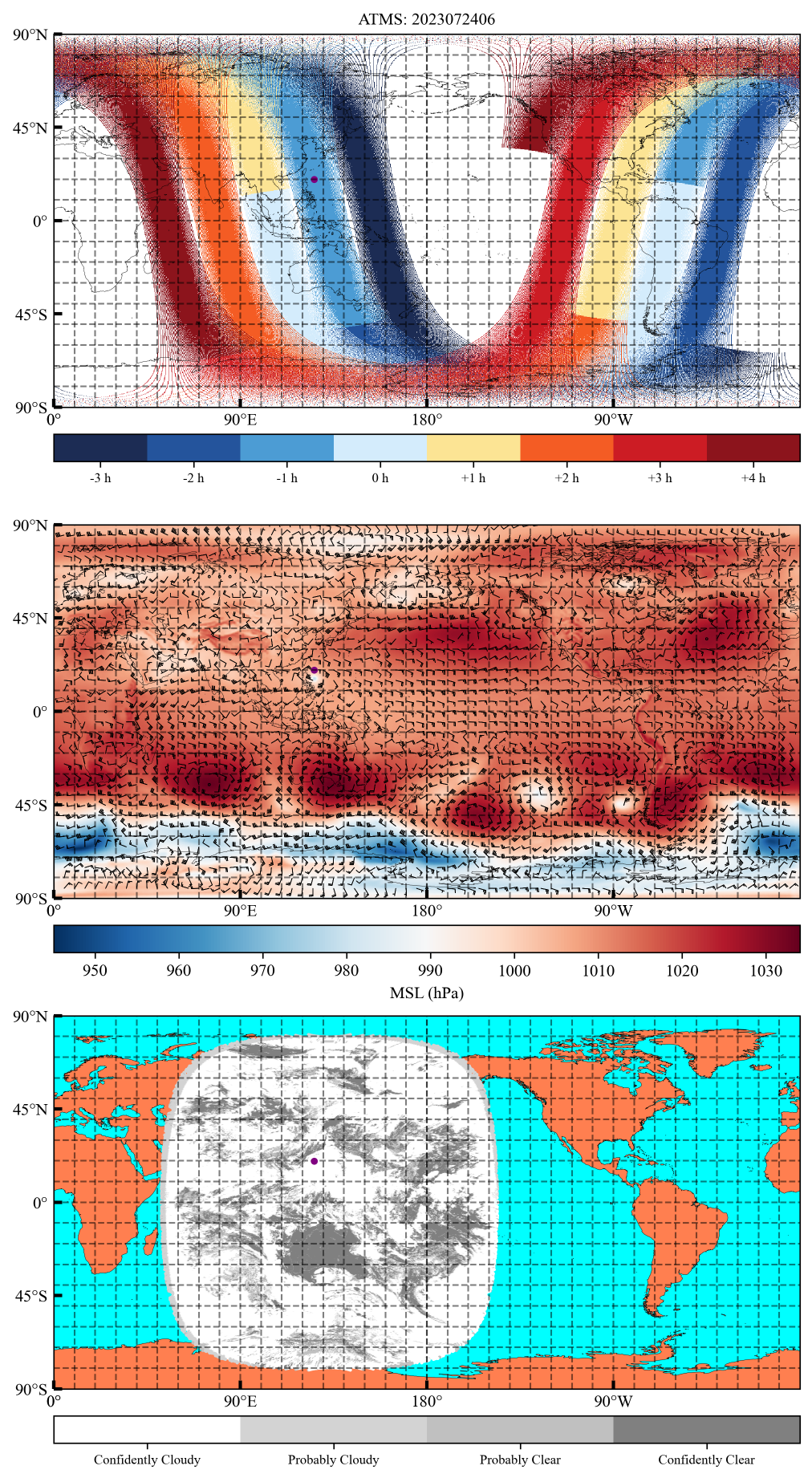}
\caption{Single observation tests with 5 K perturbation. The perturbation is located over the ocean near Typhoon Doksuri, at 19.9$^\circ$ N, 125.5$^\circ$ E (purple dot). The first panel represents the brightness temperature observed by NOAA-20 ATMS over an 8-hour period, spanning from 3 hours before to 4 hours after 06 UTC on July 24, 2023. The second panel illustrates the spatial distribution of mean sea-level pressure (${\textrm{MSL}}$) with 850 hPa wind vectors overlaid. The third panel displays the cloud mask product retrieved from the Advanced Geosynchronous Radiation Imager (AGRI) aboard Fengyun-4B. The perturbation is located in cloudy region.}
\label{single_obs_TC_obs}
\end{figure}
\FloatBarrier

\begin{table}
\centering
\caption{\label{glossary} Summary of all input and output variables of the FuXi model. The "Type" indicates whether the variable is a time-varying variable, including upper-air, and single-level, geographical variable, or temporal variable. 
The “Full name” and “Abbreviation” columns refer to the complete name of each variable and their corresponding abbreviations in this paper. The "Role" column clarifies whether each variable serves as both an input and a output, or is solely utilized as an input by our model.}
\begin{tabularx}{\textwidth}{cXcc}
\hline
\textbf{Type} & \textbf{Full name} & \textbf{Abbreviation} & \textbf{Role} \\
\hline
Upper-air variables & geopotential & ${\textrm{Z}}$ & Input and Output   \\
                    & temperature  & ${\textrm{T}}$ & Input and Output   \\
                    & u component of wind & ${\textrm{U}}$ & Input and Output  \\
                    & v component of wind & ${\textrm{V}}$ & Input and Output  \\
                    & relative humidity & ${\textrm{R}}$ & Input and Output  \\
                    \hline
Single-level variables  & 2-meter temperature & ${\textrm{T2M}}$ & Input and Output  \\
                    & mean sea-level pressure & ${\textrm{MSL}}$ & Input and Output  \\
                    & 10-meter u wind component & ${\textrm{U10}}$ & Input and Output  \\
                    & 10-meter v wind component & ${\textrm{V10}}$ & Input and Output  \\
                    & total precipitation & ${\textrm{TP}}$ & Input and Output \\                    
\hline
Geographical        & orography & ${\textrm{OR}}$ & Input \\
                    & latitude & ${\textrm{LAT}}$ & Input \\
                    & longitude & ${\textrm{LON}}$ & Input \\  
\hline                  
Temporal            & hour of day & ${\textrm{HOUR}}$ & Input \\
                    & day of year & ${\textrm{DOY}}$ & Input \\
                    & step & ${\textrm{STEP}}$ & Input \\                    
\hline
\end{tabularx}
\end{table}

\begin{table}
\centering
\caption{\label{satellite_orbit} A summary of polar-orbiting satellite information.}
\begin{tabularx}{\textwidth}{cXXcX}
\hline
Satellite & Orbit & Instruments & Channels & Preprocessed Data Dimensions \\
\hline
FY-3E & Polar-orbiting (early morning) & \begin{tabular}[X]{@{}X@{}}MWTS-III\\MWHS-II\end{tabular} & \begin{tabular}[c]{@{}c@{}}17\\15\end{tabular} & \begin{tabular}[X]{@{}X@{}}$8 \times 20 \times 720 \times 1440$\\$8 \times 18 \times 720 \times 1440$\end{tabular}\\
Metop-C & Polar-orbiting (morning) & \begin{tabular}[X]{@{}X@{}}AMSU-A\\MHS\end{tabular}  & \begin{tabular}[c]{@{}c@{}}15\\5\end{tabular} & \begin{tabular}[X]{@{}X@{}}$8 \times 18 \times 720 \times 1440$\\$8 \times 8 \times 720 \times 1440$\end{tabular}\\
NOAA-20 & Polar-orbiting (afternoon) & ATMS & 22 & $8 \times 25 \times 720 \times 1440$ \\
GNSS-RO & - & - & 512 & $8 \times 32 \times 720 \times 1440$ \\
\hline
\end{tabularx}
\end{table}

\clearpage

\maketitle

\setcounter{figure}{0}    
\setcounter{table}{0}    

\renewcommand{\figurename}{Supplementary Figure}
\renewcommand{\tablename}{Supplementary Table}

\section*{Supplementary materials}

Supplementary Text S1 to S8\\
Supplementary Figs. S1 to S23\\
Supplementary Tables. S1\\

\section*{Supplementary Text}

\setcounter{section}{0}

\section{Methods} 
\label{methods}

\subsection{ECMWF Reanalysis v5 (ERA5)}

The fifth generation of the the European Centre for Medium-Range Weather Forecasts (ECMWF) reanalysis dataset, ERA5, with a temporal resolution of 1 hour and a spatial resolution of approximately 31 km, offers data from January 1950 to the present day \cite{hersbach2020era5}. ERA5 is widely regarded as the most comprehensive and accurate global reanalysis archive. In this study, we used the 6-hourly ERA5 dataset at a spatial resolution of 0.25\textdegree 
 ($720\times1440$ latitude-longitude grid points). 

The FuXi model predicts a total of 70 variables, including five upper-air atmospheric variables across 13 pressure levels (50, 100, 150, 200, 250, 300, 400, 500, 600, 700, 850, 925, and 1000 hPa), and five surface variables. The upper-air atmospheric variables include geopotential (${\textrm{Z}}$), temperature (${\textrm{T}}$), u component of wind (${\textrm{U}}$), v component of wind (${\textrm{V}}$), and relative humidity (${\textrm{R}}$). The surface variables include 2-meter temperature (${\textrm{T2M}}$), mean sea-level pressure (${\textrm{MSL}}$), 10-meter u wind component (${\textrm{U10}}$), 10-meter v wind component (${\textrm{V10}}$), and total precipitation (${\textrm{TP}}$). A comprehensive list of these variables and their corresponding abbreviations is given in Extended Data Table 1.

The FuXi-DA model was trained using data spanning the period June 1, 2022 - June 30, 2023. To ensure robust testing, the model utilized a comprehensive dataset covering an entire year, from July 1, 2023, to June 30, 2024.
Despite the relatively limited 2-year dataset for model development, reserving one year for testing is crucial to ensure a thorough evaluation across all four seasons.
In the FuXi Weather system, cyclic forecasts were initialized using all-zero tensors as the initial conditions. To account for the spin-up phase of the model, common in physics-based Numerical Weather Prediction (NWP) and Data Assimilation (DA) systems, the first 2 days of cyclic analysis and forecasts were excluded from the evaluation to ensure the forecasting system had reached its normal operating mode.
As illustrated in the Supplementary Fig.\ref{analysis_incremental_learning}, the root-mean-squared-error (RMSE) values initially started high on July 1, 2023 then progressively declined, quickly reaching a state of statistical equilibrium.
Consequently, the evaluation period effectively covers approximately one year, from July 03, 2023 to June 30, 2024.

\subsection{Satellite data}
\label{satellite_data}

This subsection describes all the satellite data utilized in this study.
Over the past few decades, satellite observations have become increasingly important in the DA process of numerical prediction systems, primarily owing to their high spatialtemporal resolution and extensive spatial coverage. These observations have emerged as the predominant source of data, markedly enhancing the accuracy of NWP forecasts \cite{eyre2020,Eyre2022recent}. 
Joo et al. \cite{Joo2013} demonstrated that satellite observations account for $64\%$ of the error reduction in short-range forecasts, in contrast to $36\%$ for surface-based observations. Notably, polar-orbiting satellites contribute $85\%$ of the data assimilated into global NWP models \cite{Goldberg2013,Zhou2019,Jerald2023}.

Polar-orbiting satellites cross the equator at consistent local solar time (LST) every day \cite{kidder1995satellite}, providing two observations per day for a given location. These satellites are categorized by their LST when crossing the equator: early morning (EM) satellites around 06:00 (HH:MM) LST, mid-morning satellites (AM) at approximately 10:00 LST, and afternoon (PM) satellites at approximately 14:00 LST \cite{kidder1995satellite}.
Operational global DA systems require assimilating observational data every 6 hours to update the initial conditions for NWP models. A two-orbit system covers approximately $80\%$ of the global area, leaving gaps lacking observations. Therefore, an optimal polar operational constellation should be a three-orbit system (EM, AM, and PM) \cite{James2008,eyre2014wmo}. 

In this study, we selected FengYun-3E (FY-3E) as the EM satellite, Meteorological Operational Polar Satellite - C (Metop-C) as the AM satellite, and National Oceanic and Atmospheric Administration - 20 (NOAA-20) as the PM satellite.
Together, the three satellites provide full global coverage in every 8-hour DA window.
Microwave sounders are relatively insensitive to cloud and have substantially improved NWP forecasts, representing the most significant contributors among various satellite instruments \cite{bormann2019global,Eyre2022recent}. Consequently, the microwave sounders onboard these three selected satellites were used for DA in this study, as detailed in Extended Data Table 2.
Extended Data Fig.1 illustrates the spatial coverage of data collected by these polar-orbiting satellites, which was used to generate the analysis fields for 12 UTC on June 1, 2023. As shown in the figure, data from these satellites, gathered within an 8-hour observation window, provides almost global coverage.

In this study, we directly employed  the brightness temperatures from the three polar-orbiting satellites for DA.

\begin{itemize}
\setlength\itemsep{1em}
    \item FY-3E (EM orbit):
    The FY-3E, launched on 5 July, 2021 from the Jiuquan Satellite Launch Centre, is the  first civil meteorological satellite in EM orbit, filling a critical gap in global polar-orbiting satellite coverage \cite{zhang2022fy,shao2022system,Peng2024}. It crosses the equator between 05:30 and 05:50 LST and carries 11 scientific instruments. For this study, we utilized all 17 channels of the Microwave Temperature Sounder-III (MWTS-III) and 15 channels of the Microwave Humidity Sounder-II (MWHS-II).

    \item Metop-C (AM orbit):
    The Metop-C, launched on 7 November 2018, is the third satellite in a series of three polar-orbiting meteorological satellites developed collaboratively by the European Space Agency (ESA) and the European Organization for the Exploitation of Meteorological Satellites (EUMETSAT) as part of the EUMETSAT Polar System (EPS) \cite{righetti2020metop}. Crossing the equator at  approximately 09:30 LST \cite{righetti2020metop}, the Metop-C has eight instruments onboard. All five channels from the Microwave Humidity Sounder (MHS) and 15 channels from the Advanced Microwave Sounding Unit-A (AMSU-A) were used in this study.
    
    \item NOAA-20 (PM orbit):
    The NOAA-20, formerly known as JPSS-1, was launched on 18 November 2017. It is the first satellite of the U.S. next generation, polar-orbiting, environmental satellite system under the Joint Polar Satellite System (JPSS) program \cite{asbury2018jpss,WANG2021}. The NOAA-20, which crosses the equator at 13:30 LST, is equipped with five instruments \cite{goldberg2018}, including the Advanced Technology Microwave Sounder (ATMS). The ATMS has 22 channels: the first 16 channels are primarily used for temperature sounding from the surface to approximately 1 hPa (~45 km) and the remaining channels (17–22) for humidity sounding in the troposphere from the surface to approximately 200 hPa (~10 km) \cite{Goldberg2013}. This  provides high-resolution measurements of temperature and moisture, and offers detailed insights into tropical cyclone (TC) warm cores and rainfall intensity when measuring conditions within the eye of a TC. 

\end{itemize}

Radio occultation (RO) is a cost-effective remote-sensing technique that accurately measures the atmospheric gradient of atmospheric refractivity. This measurement is essential for deriving vertical profiles of temperature, pressure, and humidity \cite{Kursinski1997,kursinski2000gps,Anthes2011}.
In this technique, signals transmitted by a Global Navigation Satellite System (GNSS) satellite and received by a low-Earth orbiting satellite, are refracted by the Earth’s atmosphere. This refraction alters the path and timing of the signals. The precise positions of both the transmitting and receiving satellites enables the accurate measurement of signal delays and bending angles, which are crucial to derive weather parameters such as temperature, pressure, and humidity. Radio occultation data offers global three-dimensional (3D) coverage, high accuracy, and high vertical resolution. Unlike other remote-sensing measurements, which are often negatively affected by surface weather conditions such as clouds and precipitation, RO measurements maintain their integrity under all weather conditions, ensuring continuous data acquisition.
Global Navigation Satellite System RO data produce high-resolution profiles from the surface of the Earth up to the stratopause, an atmospheric layer typically characterized by sparse radiance data and poor NWP model performance, making them highly complementary to other sounding data (e.g., data from microwave sounders) within operational DA systems \cite{Cardinali2014,Bauer2014,Eyre2022recent}.
In this study, we used vertical profiles of refractivity retrieved from GNSS-RO. However, GNSS-RO data are inhomogeneous, non-uniform, and sparsely distributed globally, appearing as discrete data points across various atmospheric columns \cite{Shehaj2023}.
Therefore, specialized preprocessing techniques are required to transform GNSS-RO data into a gridded format, which simplifies integration into the FuXi-DA model, and aligns with the assimilation of other satellite data.
This study utilized RO data collected from multiple GNSS satellite receivers. The availability timeline, indicating periods when data were either available or missing for each receiver, is illustrated in Supplementary Fig.\ref{GNSS_availability}.

\subsection{Satellite data availability}

\begin{figure}[h]
    \centering
    \includegraphics[width=\linewidth]{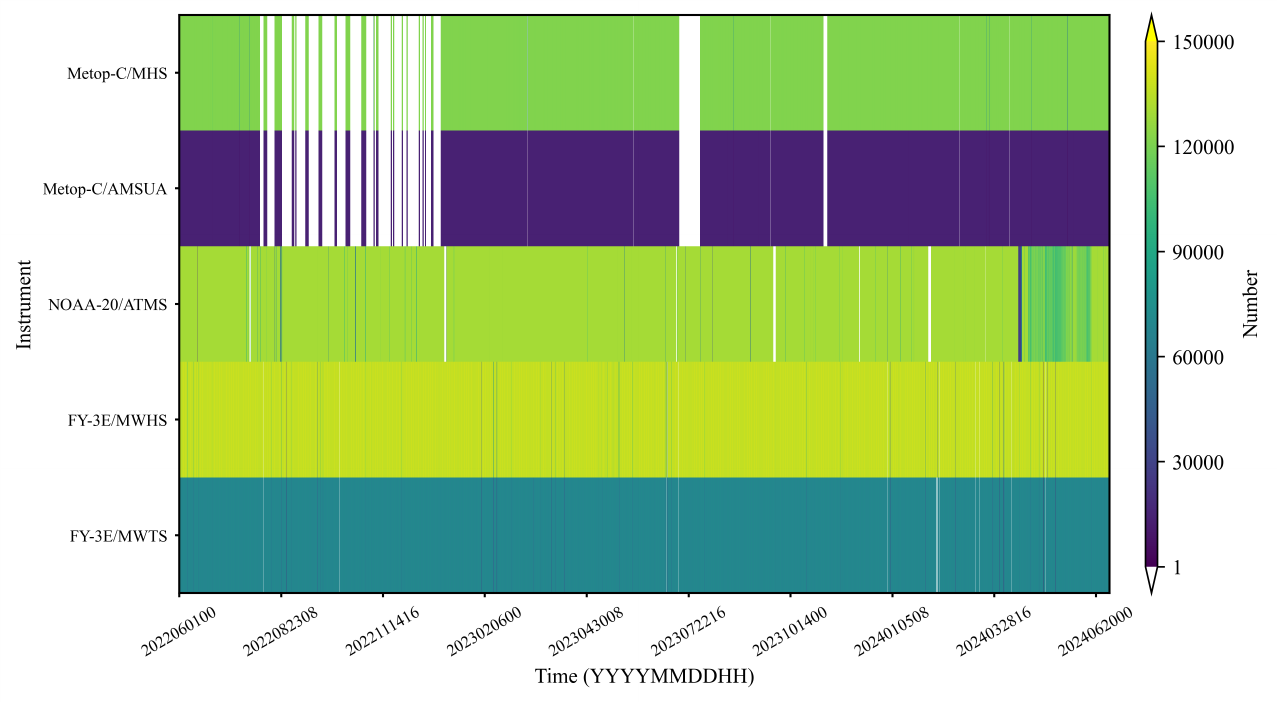}
    \caption{Satellite data availability for FengYun-3E, Meteorological Operational Polar Satellite-C (Metop-C), and National Oceanic and Atmospheric Administration-20 (NOAA-20) for the 1-year testing period. The colors denote the number of observations at the temporal resolution of 1 hour.}
    \label{Polar_availability}        
\end{figure}

\begin{figure}[h]
    \centering
    \includegraphics[width=\linewidth]{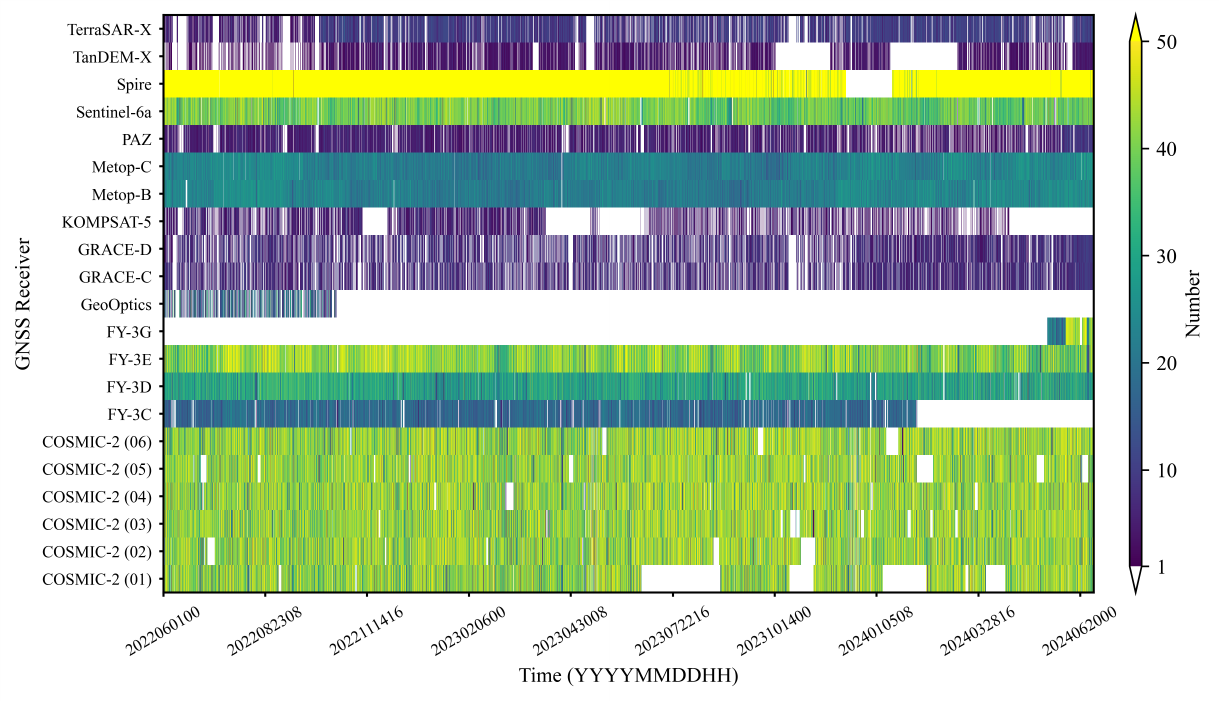}
    \caption{Satellite data availability for the 21 Global Navigation Satellite System (GNSS) receivers used herein for the 1-year testing period. The colors denote the number of observations at the temporal resolution of 1 hour.}
    \label{GNSS_availability}        
\end{figure}

Supplementary Figs.\ref{Polar_availability} and \ref{GNSS_availability} present timelines of satellite data availability of the whole dataset for the development of FuXi Weather, covering the period June 1, 2022 - June 30, 2024.
Among the three polar-orbiting satellites, the Metop-C satellite exhibited greater data unavailability compared with the other two satellites, especially during the training phase.

Supplementary Fig.\ref{GNSS_availability} illustrates the data availability from various meteorological satellites, including several commercial constellations.
The Constellation Observing System for Meteorology, Ionosphere, and Climate 2 (COSMIC-2) \cite{schreiner2020cosmic,Anthes2022}, a joint mission by Chinese Taiwan and the U.S., is one such commercial satellite constellation. 
Launched on 25 June 2019, COSMIC-2 includes six satellites designed to collect RO data using GNSS signals.
This constellation notably contributes to global weather prediction, ionospheric research, and climate monitoring efforts.

\FloatBarrier

\FloatBarrier

\section{FuXi Weather}
\label{FuXi_Weather}

\begin{figure}[h]
\centering
\includegraphics[width=1.0\textwidth]{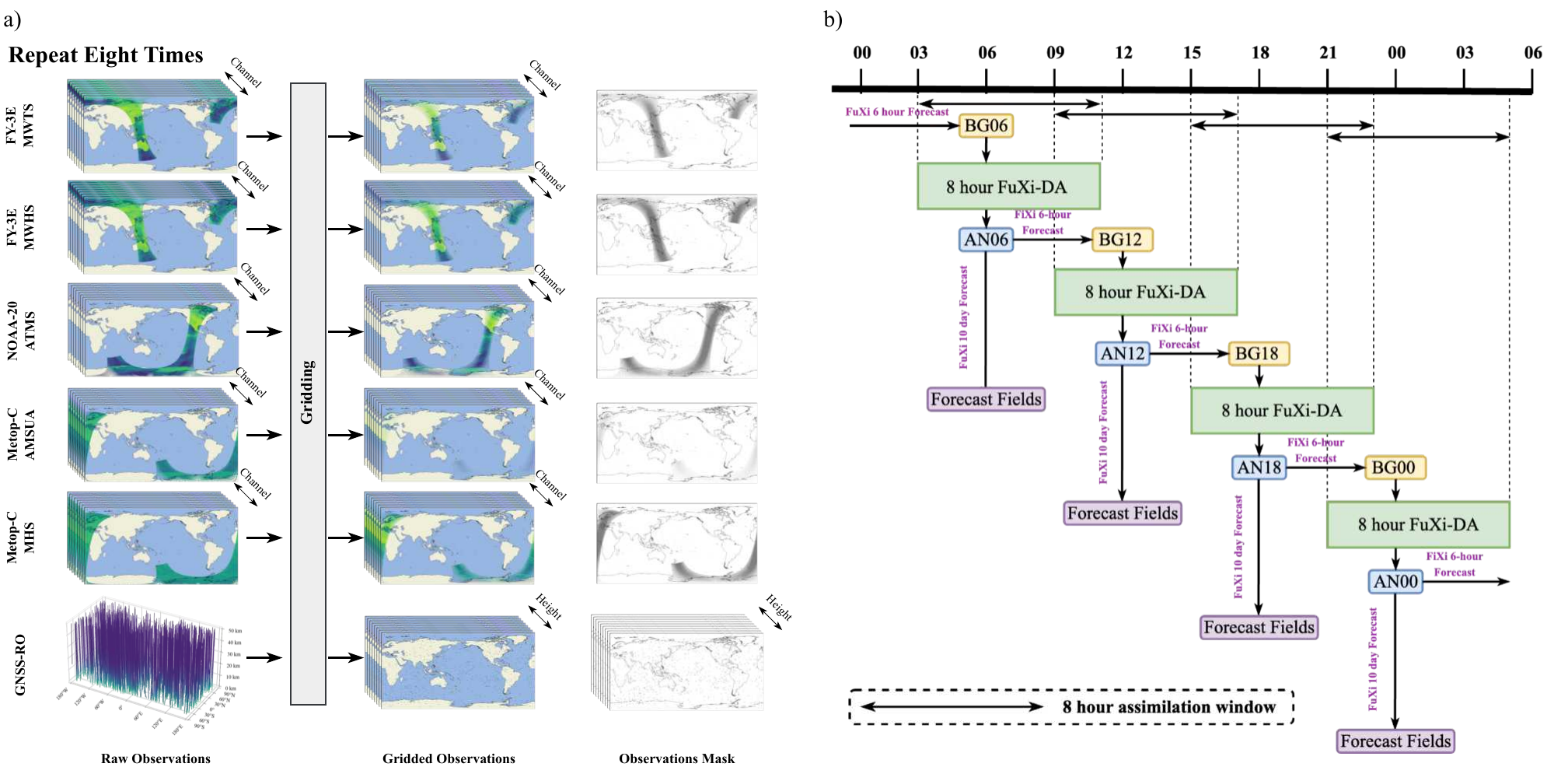}
\caption{Overall structure of the FuXi Weather system. (a) Satellite data preprocessing, incorporating data from three polar-orbiting meteorological satellites (FY-3E, Metop-C, and NOAA-20), and GNSS-RO data. (b) Data assimilation and forecasting cycles with the FuXi-DA model producing analysis ﬁelds, followed by forecast generation via the FuXi model.}
\label{model}
\end{figure}

Supplementary Fig.\ref{model} illustrates FuXi Weather, designed for generating 6-hourly updated global weather forecasts. This system includes three main components: satellite data preprocessing, DA via the FuXi-DA model, and weather forecasting via the FuXi model. During the DA phase, the FuXi-DA model combines preprocessed satellite data with prior (background) forecasts to produce analysis ﬁelds. These ﬁelds are then utilized by the FuXi model to generate 10-day weather forecasts.

\subsection{Satellite data preprocessing}
\label{satellite_data_preprocessing}

This subsection elaborates on the data preprocessing process applied to satellite data described previously. Extended Data Fig. 1 shows the spatial coverage of FY-3E, Metop-C, and NOAA-20, denoted by blue, red, and green colors, respectively, over an 8-hour period spanning from 3 hours before to 4 hours after 12 UTC on June 1, 2023. As shown in Extended Data Fig. 1, satellite data are heterogeneous in space and time; hence, nearest-neighbor interpolation from satellite data to the FuXi output grids must be performed. Data from the three polar-orbiting satellites were averaged within grid points corresponding to the FuXi forecasts ($720 \times 1440$  in latitude and longitude) to represent the satellite observations at each location. All data collected by an instrument onboard a speciﬁc satellite within this 8-hour assimilation window were concatenated before being input to FuXi-DA. 

Preprocessing GNSS-RO data poses considerable challenges owing to its inhomogeneous, non-uniform, and sparse global distribution. Global Navigation Satellite System RO data are discrete points \cite{Foelsche2008}, deﬁned by latitude, longitude, and altitude, structurally resembling 3D point clouds generated by 3D sensors such as light detection and ranging (LiDAR)-based devices, as shown in Extended Data Fig. 1. A point cloud is a collection of data points in 3D space, typically represented by Cartesian coordinates along the x, y, and z axes, and potentially includes additional features such as surface normals, red-green-blue values, and timestamps \cite{guo2020deep,bello2020deep}. The adoption of this format has surged in popularity owing to the growing availability and expanding applications of 3D sensing technologies in ﬁelds such as robotics, autonomous driving, and augmented/virtual reality \cite{zhou2018voxelnet}. Machine learning, which has been successful with two-dimensional (2D) data, faces considerable challenges when applied to 3D point clouds such as GNSS-RO data, which are characterized by an irregular global distribution and unstructured nature.

In this study, we adapted the PointPillars \cite{lang2019pointpillars} approach, originally developed for learning representations from point clouds organized into vertical columns, to process the refractivity profiles retrieved from GNSS-RO data.
First, the GNSS-RO data were discretized and aligned with the grid utilized in the FuXi forecasts, and then flattened into a set of 'pillars'.
Owing to the inherent sparsity of GNSS-RO data, most pillars were empty, with only a few containing data. 
Extended Data Fig.1 illustrates this sparsity, with yellow dots indicating the presence of GNSS-RO data at grid points over the 8-hour window from 3 hours before to 4 hours after 12 UTC on June 1, 2023.
Notably, only $0.59\%$ of ($\frac{6902}{720\times1440}$ horizontal grid points contained GNSS-RO data.
Considering the high vertical resolution of GNSS-RO data (approximately 100 m\cite{kursinski2000gps}), which far exceeds that of FuXi, the refractivity profiles were interpolated to equidistant heights up to 50 km, resulting in a 2D tensor of dimensions $512 \times N$, where 512 and N represent the number of vertical layers and the number of samples, respectively.
The subsequent encoding step involved a fully connected layer and a layer normalization \cite{ba2016layer} layer, yielding a tensor of dimensions $32 \times N$.
The encoded features are then redistributed to the FuXi forecasts' 2D grid to form a tensor sized $32 \times 720 \times 1440$, where 720 and 1440 denote the latitude and longitude grid points, respectively.
Using the PointPillars framework, these data were transformed into a refined, high-level representation, aligned on a regular grid matching the spatial resolution of background forecasts (i.e., 0.25\textdegree ;$720 \times 1440$ in latitude and longitude). This alignment facilitated the subsequent DA process via the FuXi-DA model.

To mitigate observational gaps caused by unscanned areas or data loss, we implemented a masking technique in our satellite data preprocessing. This technique assigns a value of 1 to grid points having available data and 0 where data are missing; this is crucial for the effective assimilation of satellite observations into the FuXi-DA model. These preprocessing steps are schematically illustrated in Supplementary Fig.\ref{model}a. The polar-orbiting satellite data were normalized using the z-score normalization technique, thereby ensuring uniformity in their mean and variance. The GNSS-RO refractivity data were normalized by dividing the raw data by 360.

It is important to note that, aside from excluding satellite-observed brightness temperatures above 350 K and below 50 K, no additional quality control screening is applied to the observational data input to FuXi Weather. Consequently, FuXi Weather operates as an all-pixel, all-surface, all-channel, all-sky machine-learning DA and weather forecasting system.

\subsection{FuXi-DA model architecture}
\label{fuxi_da_model}

Similar to conventional DA methods, the FuXi-DA model uses both background forecasts and observations as inputs. The background forecasts, generated by the FuXi model  \cite{chen2023fuxi},encompass both upper-air and surface variables arranged in a data cube with dimensions of $70 \times 720 \times 1440$ (variables, latitude, longitude).
Unlike the single data cube used by FuXi, FuXi-DA employs six separate data cubes: a $5 \times 720 \times 1440$ cube for five surface variables, and five $13 \times 720 \times 1440$ cubes for 5 upper-air atmospheric variables ($\textrm{Z}$, $\textrm{T}$, $\textrm{U}$), $\textrm{V}$, and $\textrm{R}$) across 13 pressure levels.
Satellite observations from six sources, including five instruments on the three-orbit system and GNSS-RO data, are structured into tensors reflecting time frames, observation channels, and geographic coordinates (latitude and longitude).
For example, the tensor dimensions for MWTS-III on FY-3E are $8 \times 20 \times 720 \times 1440$, covering 17 microwave channels and three types of encoded observational information (latitude, longitude, and satellite zenith angle).
The preprocessed data dimensions of all satellite data used in this study are summarized in Extended Data Table 2.
The encoded observational information is included because biases in satellite data often vary with scan angle and geographic location \cite{harris2001satellite, auligne2007adaptive}.
The time frames are consistently set at eight to align with the 8-hour assimilation window of FuXi-DA.
Before their integration into the FuXi-DA model, background forecast dimensions are adjusted from $721 \times 1440$ to $720 \times 1440$ using bilinear interpolation to ensure consistency with the observational data.
Before input to the FuXi-DA model, the temporal and observational information of satellite data are merged. For example, data dimensions for MWTS-III on FY-3E are restructured from $8 \times 20 \times 720 \times 1440$ to $160 \times 720 \times 1440$.

In traditional DA frameworks, the assimilation of satellite radiance requires observational operators to map background forecasts from the state vector to observational space. In contrast, FuXi-DA eliminates the need for conventional observational operators by converting observations and forecasts into separate latent spaces, and merging their information using fusion modules. The FuXi-DA model employs a multi-branch architecture, as illustrated in Supplementary Fig.\ref{FuXi-DA_model}b, incorporating dedicated branches for observations and background forecasts. Speciﬁcally, it includes six observation branches, each designed to process a set of six satellite observations: ﬁve from microwave sounders on a three-orbit system and one from GNSS-RO data.
An additional six branches handle background forecasts.
This architecture is scalable, allowing for the integration of additional satellite data not included in this study by adding more observation branches.
As illustrated in Supplementary Fig.\ref{FuXi-DA_model}, each branch processes data using two repeated blocks consisting of a downsampling layer followed by two fusion modules, then an up-sampling layer with two additional fusion modules.
The features from all branches are then concatenated and passed through a further up-sampling layer.
Finally, a refinement module, composed of three repeated U-net blocks, further improves the accuracy of the analysis fields.
Interactions among features across branches are optimized through ‘bottleneck fusion’ \cite{nagrani2021attention} across multiple layers, enhancing the integration of multi-modal data.

\begin{figure}[h]
\centering
\includegraphics[width=\linewidth]{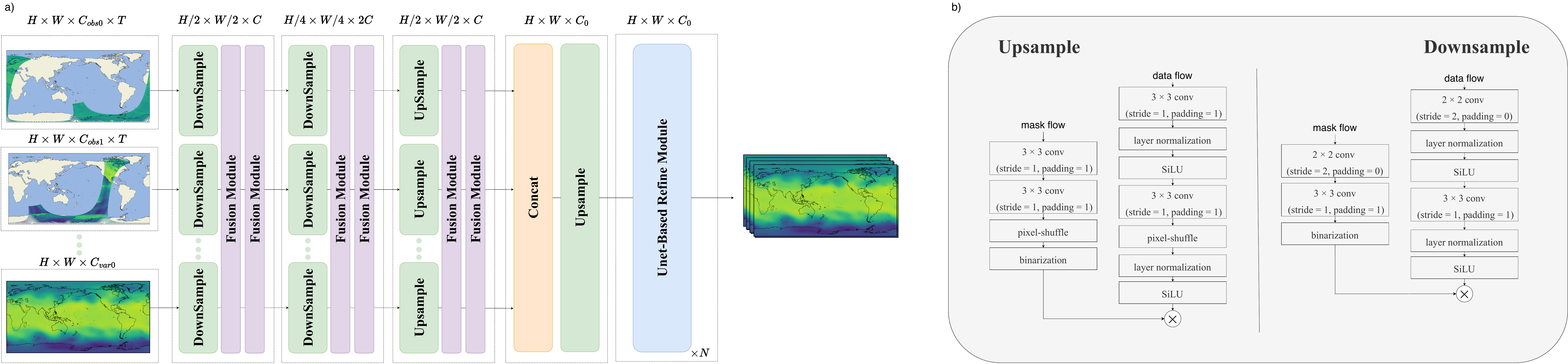}
\caption{Structure of the FuXi-DA model. (a) Overall model architecture. (b) Detailed designs of up-sampling and down-sampling layers.}
\label{FuXi-DA_model}
\end{figure}

Specifically, fusion modules adjust the weighting of observations and background data across various scales using stage-by-stage corrected condition information.
As shown in Supplementary Fig.\ref{model}b, these modules leverage the U-net architecture's down-sampling and up-sampling capabilities of the U-net architecture to enhance the interaction of information between branches and improve feature resolution through skip connections. A fusion module generates two outputs: one containing analysis increments added to the background branch and the other consisting of observed bias information added to the observation branch. This operation is based on the traditional bias correction process, in which predictions computed from the background ﬁeld are typically used for bias correction \cite{eyre1992bias,harris2001satellite,auligne2007adaptive}.

In this study, the fusion module comprised one U-Net block \cite{ronneberger2015u}, with two down-sampling and two up-sampling layers connected by skip connections to preserve local information. Each down-sampling layer includes a sequence of operations: a 2D convolution layer with a kernel size of 2 and stride of 2, followed by a layer normalization layer \cite{ba2016layer}, then a Sigmoid-weighted Linear Unit (SiLU) activation \cite{elfwing2018sigmoid}, and another 2D convolution layer with a $3 \times 3$ kernel and stride of 1.
Data dimensions are sequentially reduced from $70\times 1440 \times 720$ to $256 \times 720 \times 360$, and further to $512 \times 360 \times 180$.
Each up-sampling layer consists of a 2D convolution layer with a $3 \times 3$ kernel and stride of 1, layer normalization, SiLU activation, another 2D convolution layer (kernel size $3 \times 3$, stride 1), and a pixel-shuffle layer \cite{shi2016real} with an upscaling factor of 2.
This pixel-shuffle layer rearranges elements in a tensor of shape ($\textrm{B}$, $\textrm{C} \times r^2$, $\textrm{H}$, $\textrm{W}$) into a tensor of shape ($\textrm{B}$, $\textrm{C}$, $\textrm{H} \times r$, $\textrm{W} \times r$), where $\textrm{B}$ and $r$ denote the batch size and upscaling factor, respectively.
Subsequently, the data dimensions after two upsampling layers are restored to $256 \times 720 \times 360$ and finally to $70\times 1440 \times 720$.
In total, the FuXi-DA model consists of 0.7 billion parameters.

The FuXi-DA model performs global DA four times per day at 00, 06, 12, and 18 UTC, each within an 8-hour assimilation window that starts 3 hours before and ends 4 hours after the forecast initialization time.
For example, the initial conditions at 12 UTC incorporate all satellite observations from 09 to 16 UTC. This rigorous schedule ensures a comprehensive global analysis at 0.25° resolution for the FuXi forecast model, effectively incorporating all relevant satellite observations and adjusting for known biases related to scan angle and geographic location.

\subsection{FuXi model}

Following the generation of analysis ﬁelds by FuXi-DA, the FuXi model is employed to produce 10-day forecasts. As described by Chen et al. \cite{chen2023fuxi}, FuXi consists of three cascade models—FuXi-Short, FuXi-Medium, and FuXi-Long— optimized for speciﬁc forecast time windows of 0‒5, 5‒10, and 10‒15 days, respectively. Speciﬁcally, when ERA5 serves as the initial conditions, the 5-day forecasts from FuXi-Short are utilized as the starting point for FuXi-Medium, which then forecasts the subsequent 5‒10 days. However, the analysis ﬁelds from FuXi-DA are less accurate than ERA5, resulting in a marked degradation in forecast performance. This discrepancy is particularly evident in the reduced accuracy of the 5-day forecasts from FuXi-Short when based on FuXi-DA analysis ﬁelds (see main text Fig.2).
To mitigate this issue, the FuXi-Short model is ﬁne-tuned using the FuXi-DA analysis fields (details in Section \ref{finetuing} below). Moreover, we found that 4-day forecasts from FuXi-Short, when initialized with FuXi-DA analysis ﬁelds, achieved comparable accuracy to the 5-day forecasts initialized with ERA5. Consequently, we employed these 4-day forecasts as the initial conditions for FuXi-Medium for 4‒10-day predictions. The impact of this adjustment is shown in Supplementary Figs.\ref{fuxi_short_medium_RMSE} and \ref{fuxi_short_medium_ACC}, where it can be seen that using FuXi-Short for 4-day forecasts leads to improved accuracy in 4–10-day predictions, reflected in lower RMSE and higher anomaly correlation coefficient (ACC) values compared with those when using FuXi-Short for 5-day forecasts. Although further ﬁne-tuning of FuXi-Medium could theoretically further enhance performance, this is beyond the scope of this paper.

\begin{figure}[h]
    \centering
    \includegraphics[width=\linewidth]{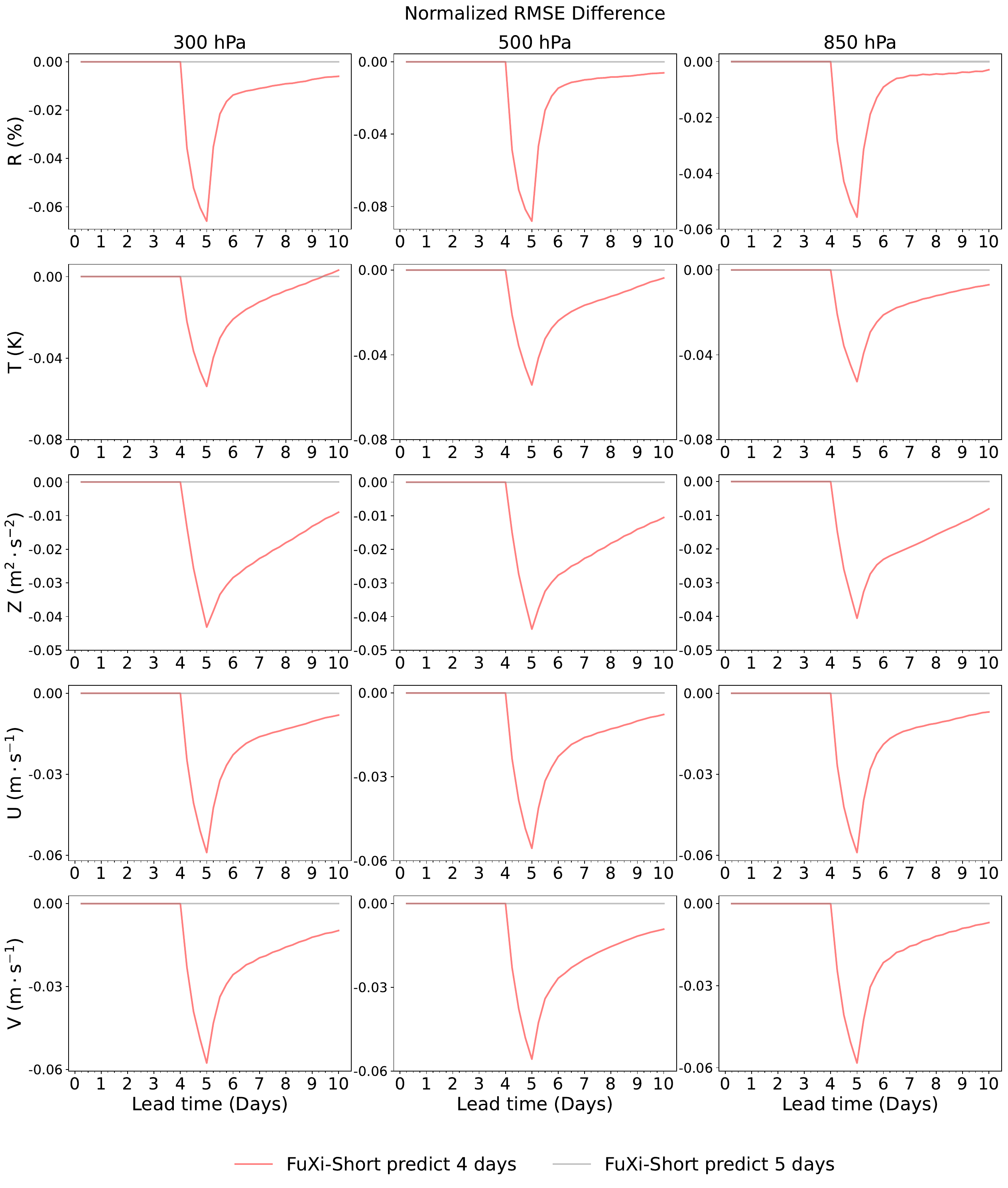}
    \caption{Comparison of forecast performance between the use of FuXi-Short for 4-day forecasts and the use of FuXi-Short for 5-day forecasts. The ﬁgure presents the normalized differences in globally averaged, latitude-weighted root-mean-square error ($\textrm{RMSE}$). The comparison spans a 1-year testing period (July 03, 2023‒June 30, 2024). The analysis includes five variables: relative humidity (${\textrm{R}}$), temperature (${\textrm{T}}$), geopotential (${\textrm{Z}}$), u component of wind (${\textrm{U}}$), and v component of wind (${\textrm{V}}$), at three pressure levels (300, 500, and 850 hPa). The five rows and three columns correspond to the five variables and three pressure levels, respectively.}
    \label{fuxi_short_medium_RMSE}        
\end{figure}
\FloatBarrier

\begin{figure}[h]
    \centering
    \includegraphics[width=\linewidth]{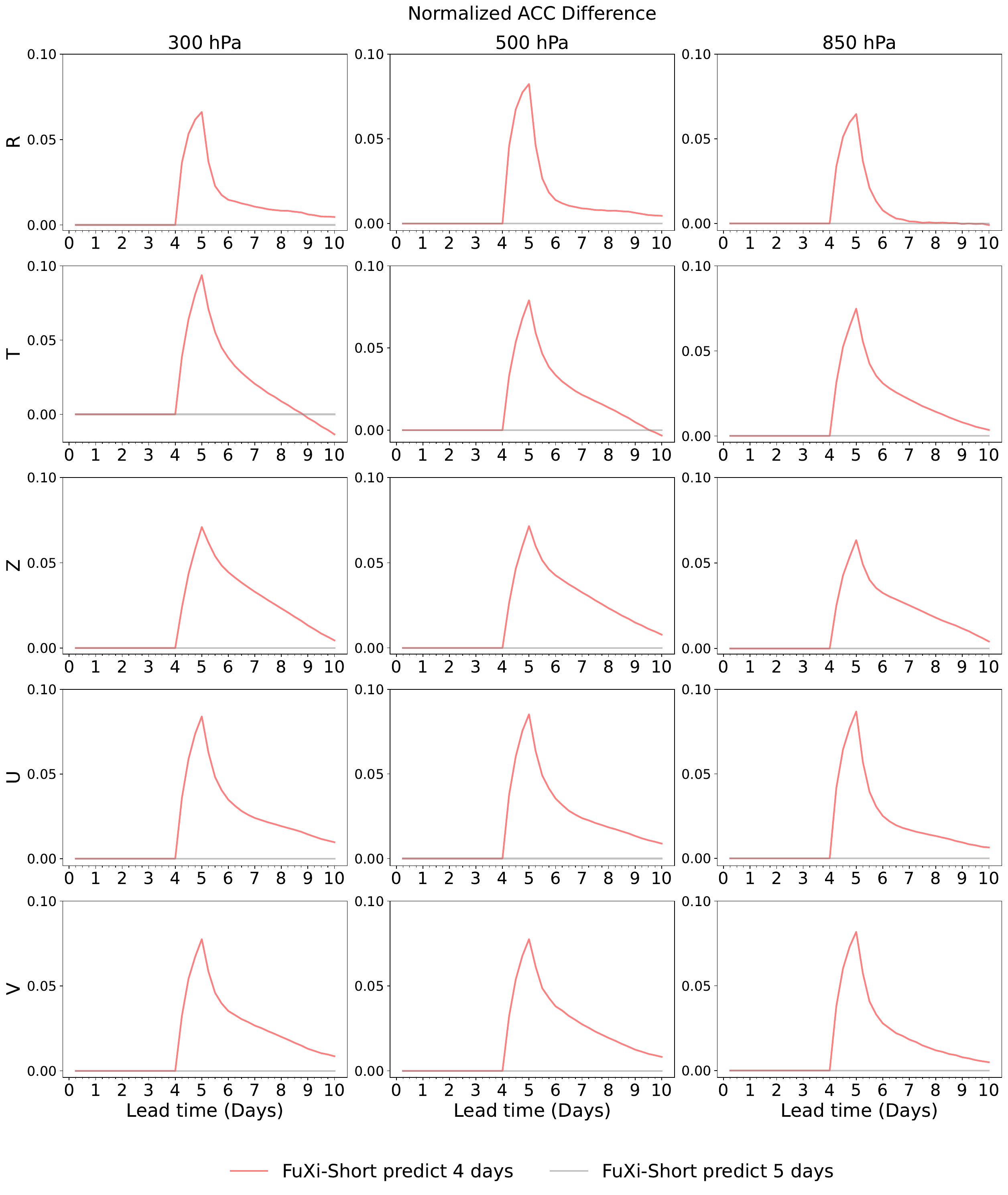}
    \caption{Comparison of forecast performance between the use of FuXi-Short for 4-day forecasts and the use of FuXi-Short for 5-day forecasts. The ﬁgure presents the normalized differences in globally averaged, latitude-weighted anomaly correlation coefficient ($\textrm{ACC}$). The comparison spans a 1-year testing period (July 03, 2023-June 30, 2024). The analysis includes five variables: relative humidity (${\textrm{RH}}$), temperature (${\textrm{T}}$), geopotential (${\textrm{Z}}$), u component of wind (${\textrm{U}}$), and v component of wind (${\textrm{V}}$), at thress pressure levels (300 hPa, 500 hPa, and 850 hPa). The five rows and three columns correspond to five variables and three pressure levels, respectively.}
    \label{fuxi_short_medium_ACC}        
\end{figure}
\FloatBarrier

\section{FuXi Weather training}
\label{FuXi_Weather_training}

\subsection{FuXi-DA model training}

As discussed previously, the FuXi-DA model operates on a 6-hour cycle, utilizing the prior 6-hour forecasts generated by FuXi Weather. The simplest approach to train the FuXi-DA model involves using these 6-hour forecasts, initialized with ERA5, as input. However, the effectiveness of this approach relies on the assumption that FuXi forecasts initialized with ERA5 are comparable in accuracy to those forecasts initialized with the analysis ﬁelds produced by FuXi-DA. It is important to note that the FuXi-DA model assimilates considerably fewer observational data compared with those used to generate ERA5, potentially resulting in discrepancies in accuracy. To better simulate operational scenarios in which FuXi-DA is used during the 6-hourly DA and forecasting cycles, we randomly sampled forecasts from the FuXi model (initialized with ERA5) over lead times ranging from 6 hours to 5 days, and used these sampled forecasts as background forecasts to train FuXi-DA. We also initialized the model training with all-zero tensors for state variables, setting all variables to zero at the start of the cyclic DA and forecasting processes for FuXi Weather.

Training of the FuXi-DA model was performed in two stages on four Nvidia A100 graphics processing units (GPUs). During the ﬁrst stage, the model underwent 24,000 gradient descent updates, with each GPU processing a batch size of 1. The model was developed using the PyTorch framework \cite{Paszke2017}, employing the AdamW optimizer with parameters $\beta_1 = 0.9$, $\beta_2 = 0.999$, and a weight decay coefficient of $10^{-5}$. The learning rate followed a warm-up and cosine annealing schedule \cite{loshchilov2017sgdr}, starting at $10^{-8}$, increasing linearly to peak at $2\times10^{-3}$ after 500 steps, and then decreasing according to a cosine annealing schedule over 24,000 iterations, totaling 48 hours. Post-training analysis revealed improved accuracy in the ﬁelds generated by the FuXi-DA model. Consequently, we adjusted our sampling strategy, reducing the maximum lead time from 5 days to 3 days for the FuXi model forecasts initialized with ERA5, and excluded all-zero tensors. In the second stage, the model underwent an additional 4000 gradient decent updates over approximately 8 hours.

To calculate the discrepancies between the outputs of the FuXi-DA model and ERA5, a latitude-weighted L1 loss function was utilized, defined as follows:
\begin{equation}
\textrm{L1}=
\frac{1}{C\times{\textrm{H}}\times{\textrm{W}}}
\displaystyle\sum_{c=1}^\textrm{C}
\displaystyle\sum_{i=1}^\textrm{H}
\displaystyle\sum_{j=1}^\textrm{W}
\alpha_i
\vert\mathbf{\widehat{X}}_{c,i,j}-\mathbf{X}_{c,i,j}\vert
\label{eq3}
\end{equation}
where $\textrm{C}$, $\textrm{H}$ and $\textrm{W}$ denote the number of channels, latitude, and longitude of the grid points, respectively. 
$\mathbf{\widehat{X}}$ represents the ground truth, and $\alpha_i= \textrm{H}\times\left.{cos\Phi_i}\middle/ \right.{\displaystyle\sum_{i=1}^Hcos\Phi_i}$ is the latitude-specific weighting factor at latitude $\Phi_i$.

To jointly optimize the analysis and forecasts, we implemented a loss function similar to the those used in traditional 4DVar approaches. Here, the forecasting model, FuXi, provides gradients only for the DA component while keeping its parameters fixed. The final loss function used for joint optimization in the FuXi-DA and FuXi model is given by: 

\begin{equation}
\textrm{L} = \textrm{L1}^0 + 
\frac{1}{\textrm{T}}
\displaystyle\sum_{t=1}^\textrm{T} \textrm{L1}^t
\label{eq4}
\end{equation}
where $\textrm{T}$ denotes the number of time steps. When $t=0$, $\mathbf{X}$ corresponds to the analysis field in $\textrm{L1}$, and when $t\gt0$, it denotes the forecast field in $\textrm{L1}$. 
In this study, we set $\textrm{T}=1$, allowing the inclusion of both the analysis field and one subsequent forecast ﬁeld at a 6-hour lead time. This setup incorporates both the current analysis ﬁeld and one subsequent forecast ﬁeld, thereby mirroring the traditional 4DVar assimilation and facilitating end-to-end training for FuXi Weather.

\subsection{Incremental learning}
\label{effect_incremental_learning}
The FuXi-DA model was developed using a 2-year dataset, with 1 year used for training and 1 year for testing. This dataset is considerably smaller than the 37-year dataset used for training cascaded FuXi forecasting models  \cite{chen2023fuxi}.
Training machine-learning models on such limited datasets poses substantial challenges in maintaining model robustness and accuracy, a scenario typical of few-shot learning \cite{ravi2017optimization,Wang2020}.
Furthermore, the quality and quantity of satellite measurements often ﬂuctuate over time owing to changes in satellite instrument characteristics and increases in the numbers of GNSS receivers (see Supplementary Fig.\ref{GNSS_availability}).

These variations mean that the FuXi-DA model must learn incrementally from new data and also retain previously acquired knowledge, a process known as incremental or continual learning. Incremental learning methods are generally categorized into three types: replay-based, regularization-based, and parameter isolation methods  \cite{van2022three}. 
For the FuXi-DA model, we adopted a replay-based incremental learning strategy, wherein the model is retrained monthly using data from the preceding year. This includes replaying data from the previous 11 months alongside new data from the current month. For example, the December 2023 update of the FuXi-DA model, intended for use in January 2024, was trained on data spanning the period December 25, 2022‒December 25, 2023. This training regime took approximately 8 hours for 4000 gradient descent updates. The effectiveness of this strategy is illustrated in Supplementary Fig.\ref{analysis_incremental_learning}, which compares the analysis ﬁelds produced by FuXi Weather with and without incremental learning. The results show a substantial improvement in performance for FuXi Weather when employing the replay-based incremental learning strategy.

\begin{figure}[h]
\centering
\includegraphics[width=\linewidth]{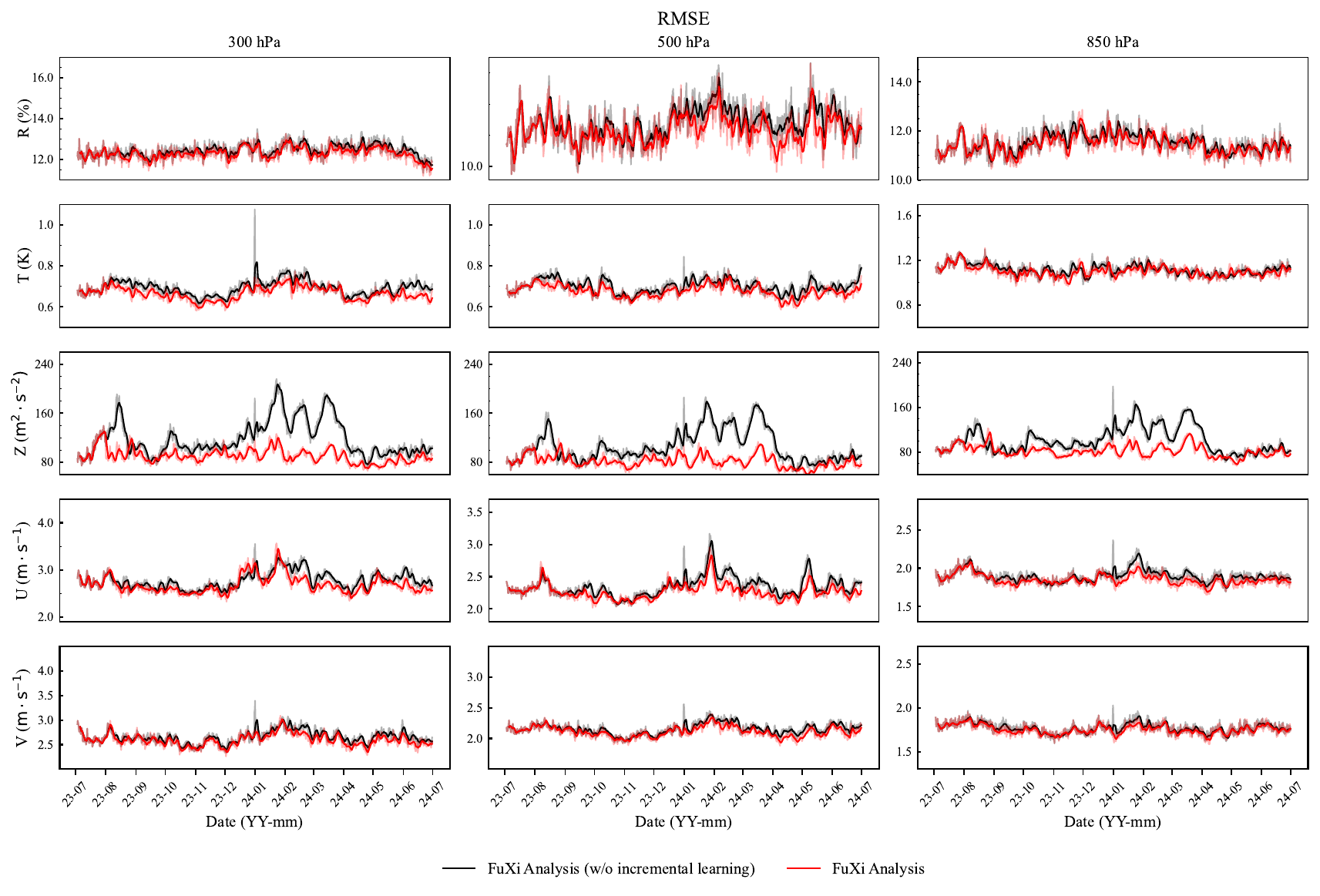}
\caption{Comparison of analysis ﬁelds produced by FuXi Weather with and without applying an incremental learning strategy over a 1-year testing period, spanning July 1, 2023‒June 30, 2024. The figure shows time series of the globally averaged, latitude-weighted root-mean-square error (RMSE) for analysis ﬁelds generated by FuXi-DA with (solid red lines) and without (solid black lines) applying the incremental learning strategy. This comparison includes five variables: relative humidity (${\textrm{R}}$), temperature (${\textrm{T}}$), geopotential (${\textrm{Z}}$), u component of wind (${\textrm{U}}$), and v component of wind (${\textrm{V}}$), at three pressure levels (300, 500, and 850 hPa). The five rows and three columns correspond to the five variables and three pressure levels, respectively. To improve clarity, the original data are shown with reduced opacity, while solid lines represent smoothed values using a 12-point moving average.}
\label{analysis_incremental_learning}
\end{figure}
\FloatBarrier

\subsection{FuXi-Short model fine-tuning}
\label{finetuing}
As previously mentioned, the quality of the analysis ﬁelds generated by FuXi-DA is inferior to that of ERA5. The degree of quality degradation varies markedly across different variables, as illustrated in main text Fig. 2, with the variable R exhibiting the highest accuracy. Consequently, we used the analysis ﬁelds produced by FuXi-DA as the initial conditions for ﬁne-tuning the FuXi model. This ﬁne-tuning process, following the methods used in GraphCast\cite{lam2022graphcast} and FuXi \cite{chen2023fuxi}, employs an autoregressive training regime and a curriculum training schedule. The training progressively increases the number of autoregressive steps from 2 to 12, with 500 gradient descent updates at each step.
Throughout this process, the model is trained at a constant learning rate of $1 \times 10^{-7}$, requiring approximately 24 hours to fine-tune on a cluster of four Nvidia A100 GPUs.
A detailed discussion of the impact of fine-tuning the FuXi-Short model is presented in Section \ref{effect_setting}.

\subsection{Experimental setup}

To enhance our understanding of the significance of incorporating background forecasts into DA, we trained a FuXi-DA model without such forecasts for comparative analysis.

Additionally, we assessed the impact of various satellite observations on global analysis and forecasting. 
In these experiments, observations from a single satellite instrument were completely excluded, and both the FuXi-DA and FuXi forecasting models were run through 1-year cycles of assimilation and forecasting to examine the effects.
Further details regarding these results can be found in the main text Section 3.4.

\section{Evaluation method}
\label{evaluation_method}

Forecasts are evaluated against a benchmark dataset at the forecast time. For FuXi model forecasts, whether initialized with ERA5 or analysis fields generated by FuXi-DA, ERA5 is used as the benchmark.
In evaluating the performance of ECMWF high-resolution (HRES) forecasts, the time series of HRES-fc0 data used to initialize these forecasts at time $t_0$ are also used as the benchmark at the evaluation time $t_0+\tau$. This choice is made to provide HRES  with the most favorable benchmark, since the aim is to demonstrate that the FuXi system can outperform it.   
 
Deterministic forecasts are evaluated using established metrics, including the root-mean-square-error (RMSE) and anomaly correlation coefficient (ACC), defined as follows:

\begin{equation}
\label{RMSE_equation}
    \textrm{RMSE}(c, \tau) =\frac{1}{\mid \textrm{D} \mid}\sum_{t_0 \in \textrm{D}} \sqrt{\frac{1}{\textrm{H} \times \textrm{W}} \sum_{i=1}^\textrm{H}\sum_{j=1}^{\textrm{W}} a_i {( \hat{\textrm{\textbf{X}}}^{t_0 +\tau}_{c,i,j} - \textrm{\textbf{X}}^{t_0 +\tau}_{c,i,j} )}^{2}}
\end{equation}

\begin{equation}
\label{ACC_equation}
    \textrm{ACC}(c, \tau) = \frac{1}{\mid \textrm{D} \mid}\sum_{t_0 \in \textrm{D}} \frac{\sum_{i, j} a_i (\hat{\textrm{\textbf{X}}}^{t_0 +\tau}_{c,i,j} - \textrm{\textbf{M}}^{t_0 +\tau}_{c,i,j}) (\textrm{\textbf{X}}^{t_0 +\tau}_{c,i,j} - \textrm{\textbf{M}}^{t_0 +\tau}_{c,i,j})} {\sqrt{ \sum_{i, j} a_i (\hat{\textrm{\textbf{X}}}^{t_0 +\tau}_{c,i,j} - \textrm{\textbf{M}}^{t_0 +\tau}_{c,i,j})^2 \sum_{i, j} a_i(\textrm{\textbf{X}}^{t_0 +\tau}_{c,i,j} - \textrm{\textbf{M}}^{t_0 +\tau}_{c,i,j})^2}}
\end{equation}
where, ${t_0}$ denotes the forecast initialization time within the testing dataset (${D}$), and ${\tau}$ is the forecast lead time. The climatological mean ($\textrm{\textbf{M}}$), calculated from ERA5 over the period 1993-2016, reflects the average conditions over these years.
To better distinguish forecast performance between models with minor differences, the normalized $\textrm{RMSE}$ difference between model A and baseline model B is calculated as \((\textrm{RMSE}_A-\textrm{RMSE}_B)/\textrm{RMSE}_B\). Similarly, the normalized $\textrm{ACC}$ difference is calculated as \((\textrm{ACC}_A-\textrm{ACC}_B)/(1-\textrm{ACC}_B)\). A negative RMSE difference and positive ACC difference indicate that model A outperforms model B.

Furthermore, RMSE can be decomposed into systematic error (or bias) and random error, which helps understanding whether the forecast errors are due to a consistent bias or random fluctuations around the observed values.
\begin{equation}
\label{MBE_equation}
    \textrm{MBE}(c, \tau) =\frac{1}{\mid \textrm{D} \mid}\sum_{t_0 \in \textrm{D}} \frac{1}{\textrm{H} \times \textrm{W}} \sum_{i=1}^\textrm{H}\sum_{j=1}^{\textrm{W}} a_i {( \hat{\textrm{\textbf{X}}}^{t_0 +\tau}_{c,i,j} - \textrm{\textbf{X}}^{t_0 +\tau}_{c,i,j} )}
\end{equation}

\begin{equation}
\label{std_equation}
    \textrm{STD}_\textrm{ERROR}(c, \tau) = \frac{1}{\mid \textrm{D} \mid}\sum_{t_0 \in \textrm{D}} \sqrt{\frac{1}{\textrm{H} \times \textrm{W}} \sum_{i=1}^\textrm{H}\sum_{j=1}^{\textrm{W}} a_i {[( \hat{\textrm{\textbf{X}}}^{t_0 +\tau}_{c,i,j} - \textrm{\textbf{X}}^{t_0 +\tau}_{c,i,j} ) - \textrm{MBE}(c, \tau)]}^{2}}
\end{equation}

\section{Single observation tests}
\label{single_obs}

\subsection{Single observation test methodology}
\label{single_obs_method}

The single observation test is a critical experiment in DA that assesses how the assimilation of one observational data point impacts the analysis field.
This involves assimilating a single observation at a specific location with a predefined innovation (observation minus background) and observation error standard deviation.
This test helps to evaluate the sensitivity of the DA system to specific observations, diagnose its performance, and improve our understanding of error propagation, ultimately refining the DA system.
The analysis increments from the test reveal valuable information, such as the ratio of background to observation error variance, variable correlations, and spatial structure of the background error covariance.
The spatial spread of the analysis increments is largely influenced by the spatial correlations within the background error.
As an initial diagnostic, the single observation test plays a crucial role in evaluating new DA systems.

For example, in the three-dimensional variational (3D-Var) DA method, the analysis increment is given by:
\begin{equation}
\mathbf{X^a-X^b}=\mathbf{BH^T(HBH^T+R)^{-1}}
(\mathbf{Y^o}-H[\mathbf{X^b}])
\label{eq7}
\end{equation}
where, $\mathbf{X^a}$ and $\mathbf{X^b}$ represent the analysis and background fields, respectively, and $\mathbf{Y^o}$ is the observation vector.
The matrices $\mathbf{B}$ and $\mathbf{R}$ denote the background and observation error covariances, respectively, while $H$ is the observation operator that maps the model state space to the observation space.
The Jacobian matrix, $\mathbf{H}=\frac{\partial H(\mathbf{x})}{\partial \mathbf{x}}$, reflects the sensitivity of the observation to the background state variables.
In the case of a single observation, the term $\mathbf{(HBH^T+R)^{-1}}(\mathbf{y^o}-H[\mathbf{x^b}])$ simplifies to a scalar, with $\mathbf B$ determining the spatial distribution of the analysis increments.
Typically, the innovation, $\mathbf{y^o}-H[\mathbf{x^b}]$ is computed directly, and $\mathbf H$ is obtained from a fast radiative transfer model (RTM) such as the radiative transfer model for television infrared observation satellite operational vertical sounder (RTTOV) \cite{hocking2021new}. 
Comparing the resulting analysis increments to theoretical expectations helps to validate the performance of the DA system.
In contrast, the FuXi-DA system does not explicitly incorporate background and observation error covariance, nor does it include an observation operator to convert the 13-level background fields to satellite observations, meaning that a precise innovation value cannot be defined for single observations.

To evaluate the impact of a single observation in FuXi-DA, followed the approach proposed by Xu et al. \cite{xu2024fuxida}, involving two runs: one with original observations and one with a small perturbation at a single grid point.
The difference between these two analysis fields represents the analysis increment caused by the perturbation. 
In this study, we introduced a 5 K perturbation to NOAA-20 ATMS over the ocean near Typhoon Doksuri at 19.9\textdegree N, 125.5\textdegree E, at 05 UTC on July 24, 2023.
The background field was a 6-hour forecast initialized at 00:00 UTC on July 24, 2023, following analysis cycles with the full observation set.
Notably, the perturbation was applied across a $5\times5$ grid point rather than a single grid point.

\subsection{Jacobians}
\label{jacobian}

In this study, Jacobians were derived using the RTTOV version 12.2, based on the US standard atmosphere profile; Jacobians can vary markedly across different atmospheric profiles.
Therefore, perfect alignment between Jacobians from the standard atmosphere and those at the perturbed location is impractical.
Ideally, Jacobians should be derived from local temperature and humidity profiles at the perturbed location, but this is not feasible with the output of FuXi Weather, which provides only 13 pressure levels and lacks the specific humidity, cloud-water, and cloud-ice data required for precise Jacobian derivation using RTTOV. Therefore, the standard US atmospheric profile was employed for illustrative purposes.

\begin{figure}[h]
\centering
\includegraphics[width=\linewidth]{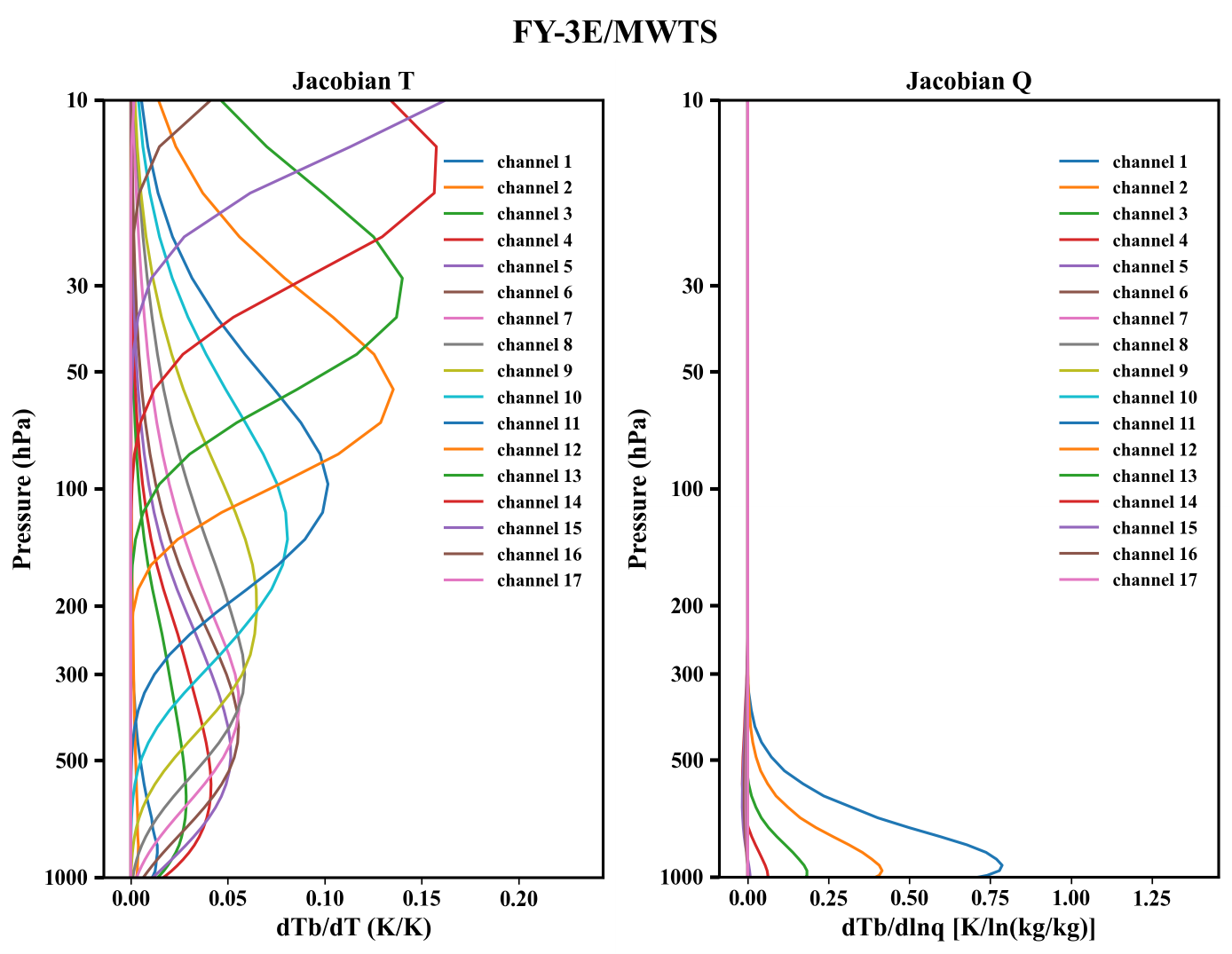}
\caption{Jacobian functions for temperature (T) and humidity (Q) derived from Microwave Temperature Sounder (MWTS) aboard FY-3E. The left panel shows T Jacobians for all channels, while the right panel displays Q Jacobians for all channels. The atmospheric profile is based on the US standard atmosphere, and radiative transfer calculations are performed using RTTOV version 12.2.}
\label{jacobian_MWTS}
\end{figure}

\begin{figure}[h]
\centering
\includegraphics[width=\linewidth]{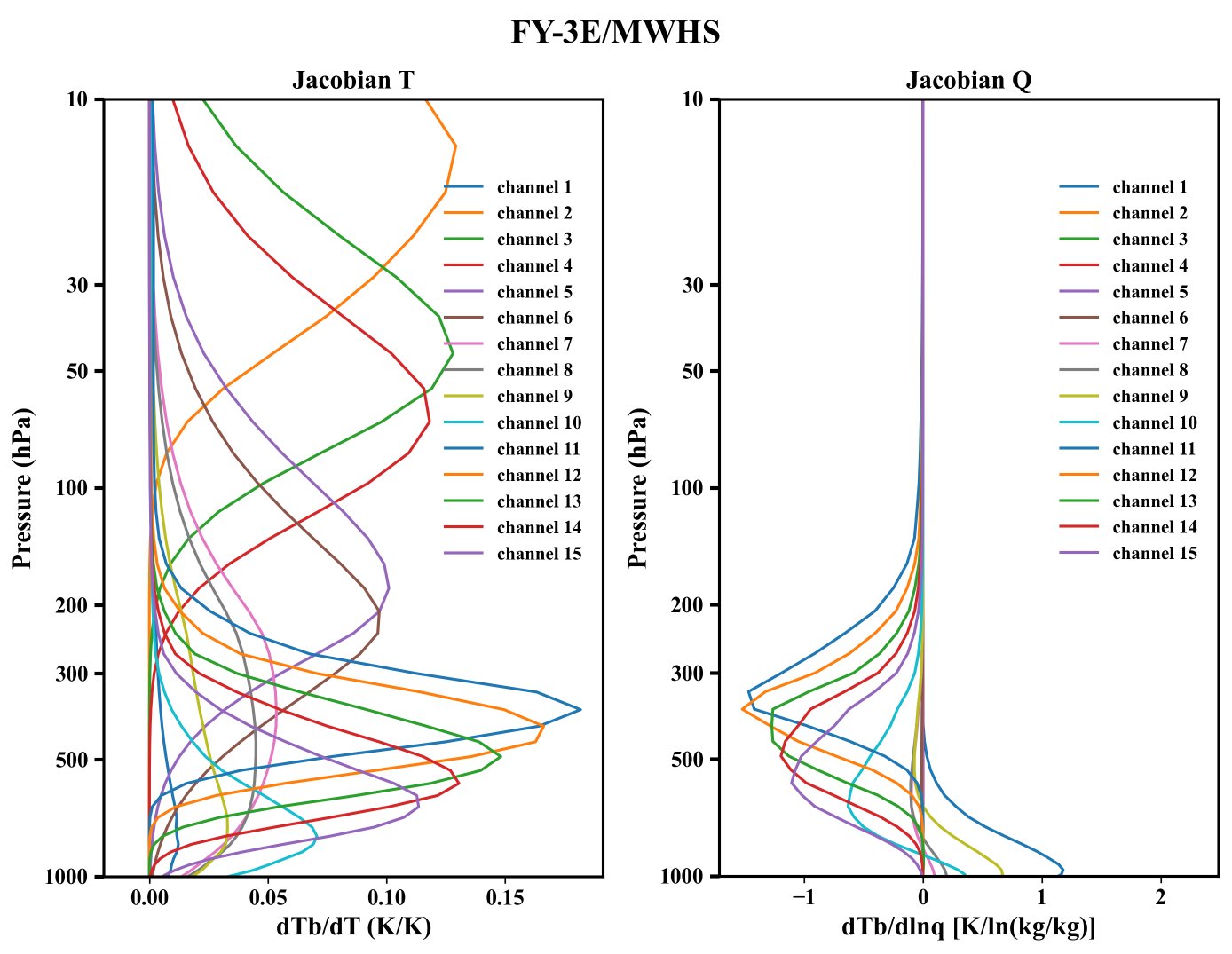}
\caption{Jacobian functions for temperature (T) and humidity (Q) derived from Microwave Humidity Sounder (MWHS) aboard FY-3E. The left panel shows T Jacobians for all channels, while the right panel displays Q Jacobians for all channels. The atmospheric profile is based on the US standard atmosphere, and radiative transfer calculations are performed using RTTOV version 12.2.}
\label{jacobian_MWHS}
\end{figure}

\begin{figure}[h]
\centering
\includegraphics[width=\linewidth]{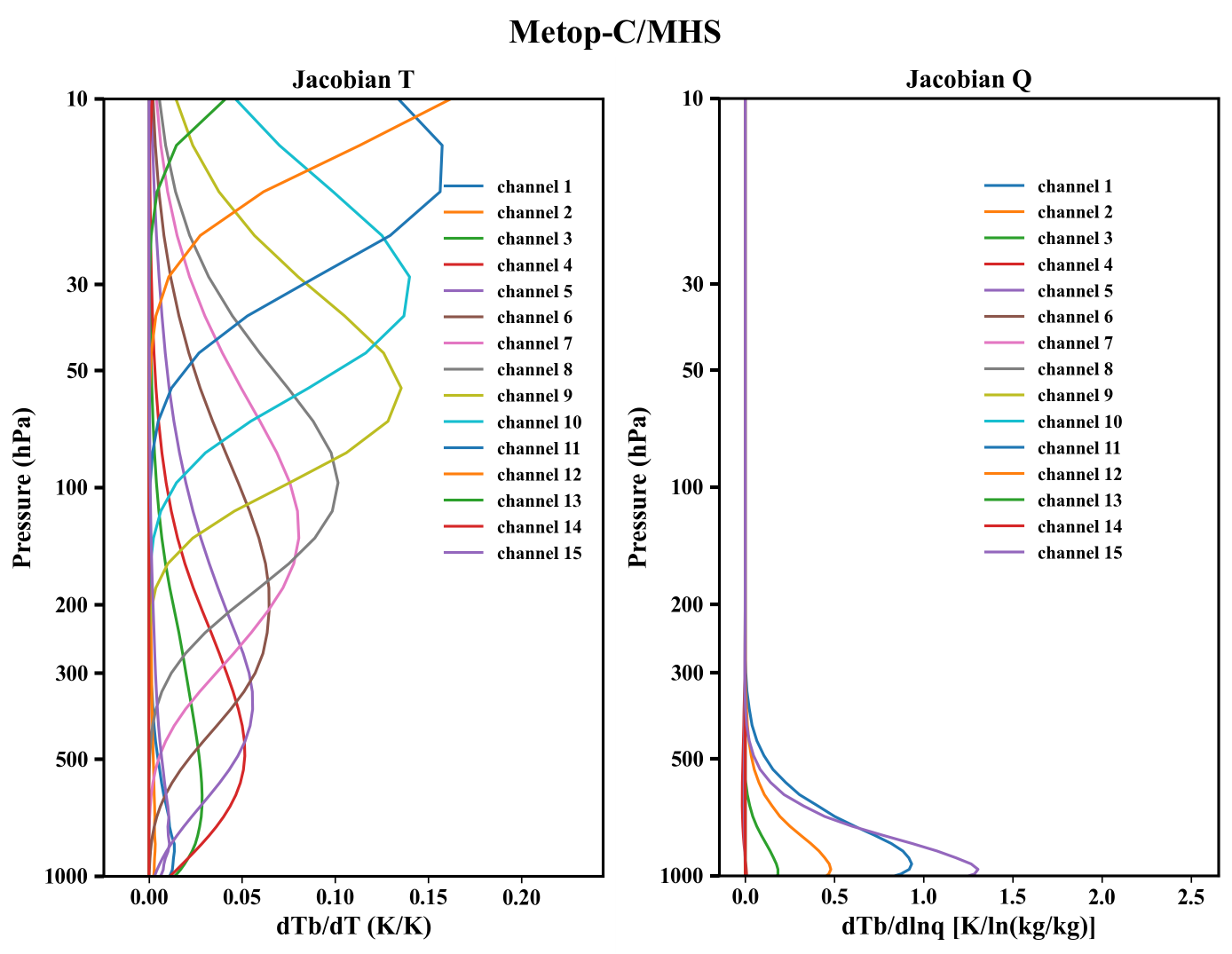}
\caption{Jacobian functions for temperature (T) and humidity (Q) derived from Advanced Microwave Sounding Unit-A (AMSU-A) aboard Metop-C. The left panel shows T Jacobians for all channels, while the right panel displays Q Jacobians for all channels. The atmospheric profile is based on the US standard atmosphere, and radiative transfer calculations are performed using RTTOV version 12.2.}
\label{jacobian_AMSUA}
\end{figure}

\begin{figure}[h]
\centering
\includegraphics[width=\linewidth]{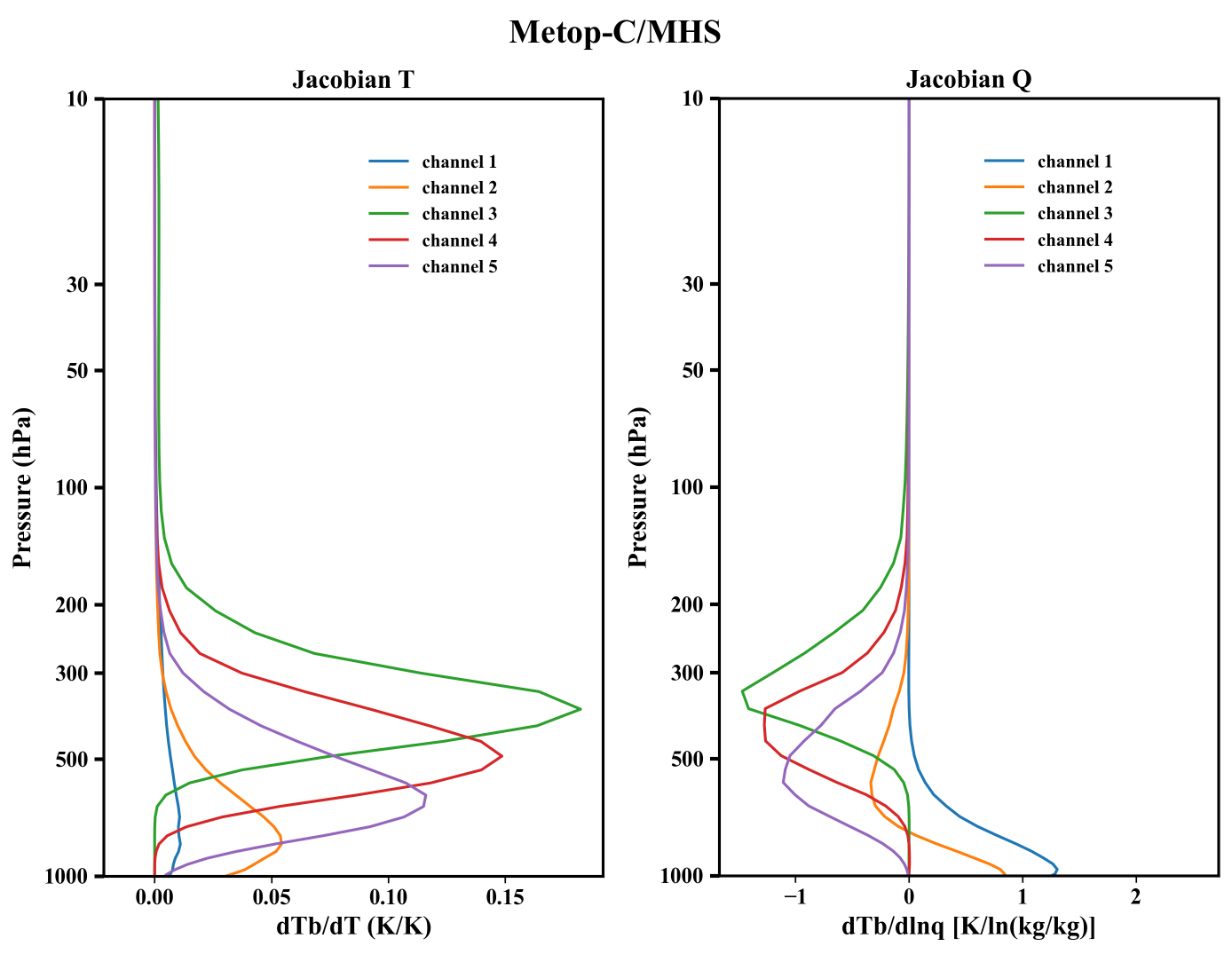}
\caption{Jacobian functions for temperature (T) and humidity (Q) derived from Microwave Humidity Sounder (MHS) aboard Metop-C. The left panel shows T Jacobians for all channels, while the right panel displays Q Jacobians for all channels. The atmospheric profile is based on the US standard atmosphere, and radiative transfer calculations are performed using RTTOV version 12.2.}
\label{jacobian_MHS}
\end{figure}

\begin{figure}[h]
\centering
\includegraphics[width=\linewidth]{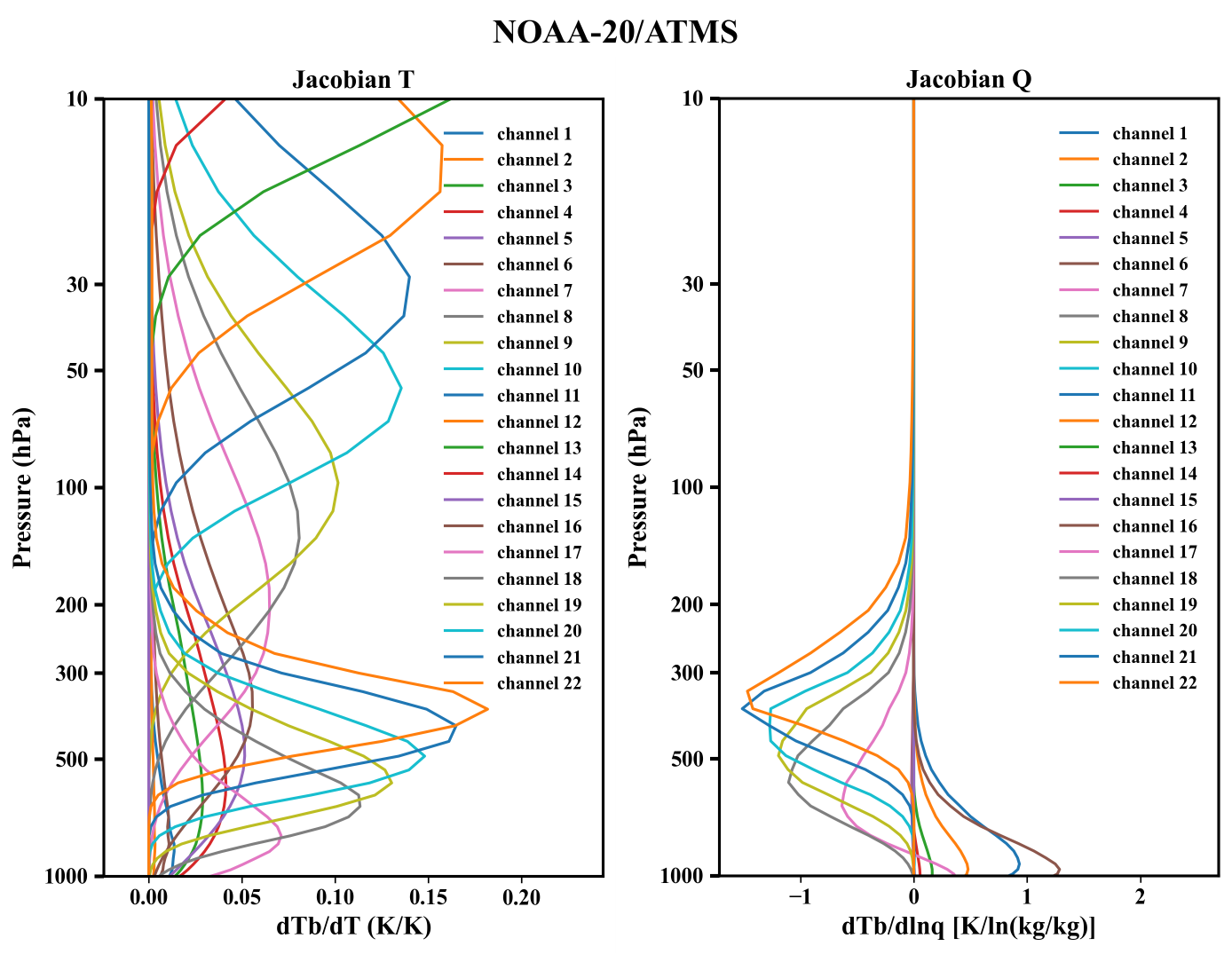}
\caption{Jacobian functions for temperature (T) and humidity (Q) derived from Advanced Technology Microwave Sounder (ATMS) aboard NOAA-20. The left panel shows T Jacobians for all channels, while the right panel displays Q Jacobians for all channels. The atmospheric profile is based on the US standard atmosphere, and radiative transfer calculations are performed using RTTOV version 12.2.}
\label{jacobian_ATMS}
\end{figure}

Supplementary Figs.\ref{jacobian_MWTS}-\ref{jacobian_ATMS} provide the temperature and humidity Jacobian functions derived from the MWHS, MWTS, AMSU-A, MHS, and ATMS instruments, respectively.
Positive temperature Jacobians indicate a positive correlation with brightness temperature, while negative humidity Jacobians signify an inverse correlation.

\FloatBarrier

\FloatBarrier

\FloatBarrier

\FloatBarrier

\FloatBarrier

\section{Data denial experiments}
\label{data_denial}

In the main text, FuXi Weather was assessed based on its control run, which assimilates a full set of satellite data.
To measure the impact of excluding specific satellite observations, two sets of data denial experiments were conducted; this is a common approach for understanding the contributions of different observations in DA systems \cite{Eyre2022recent}.

In the first set of experiments, 6-hour forecasts from the control run served as background fields at each analysis time, with specific satellite data excluded during DA.
The resulting analysis fields were compared with those from the control run.
In the second set of experiments, background forecasts from the control run were not used, and specific satellite data were withheld for DA.
The generated analysis fields were used for new 6-hour forecasts, which then became the background fields.
This setup provided insights into the robustness of FuXi Weather by examining error evolution.
Both sets of experiments quantified the impact by calculating the percentage change in RMSE, using the control run as the baseline.

\begin{figure}[h]
\centering
\includegraphics[width=1\textwidth]{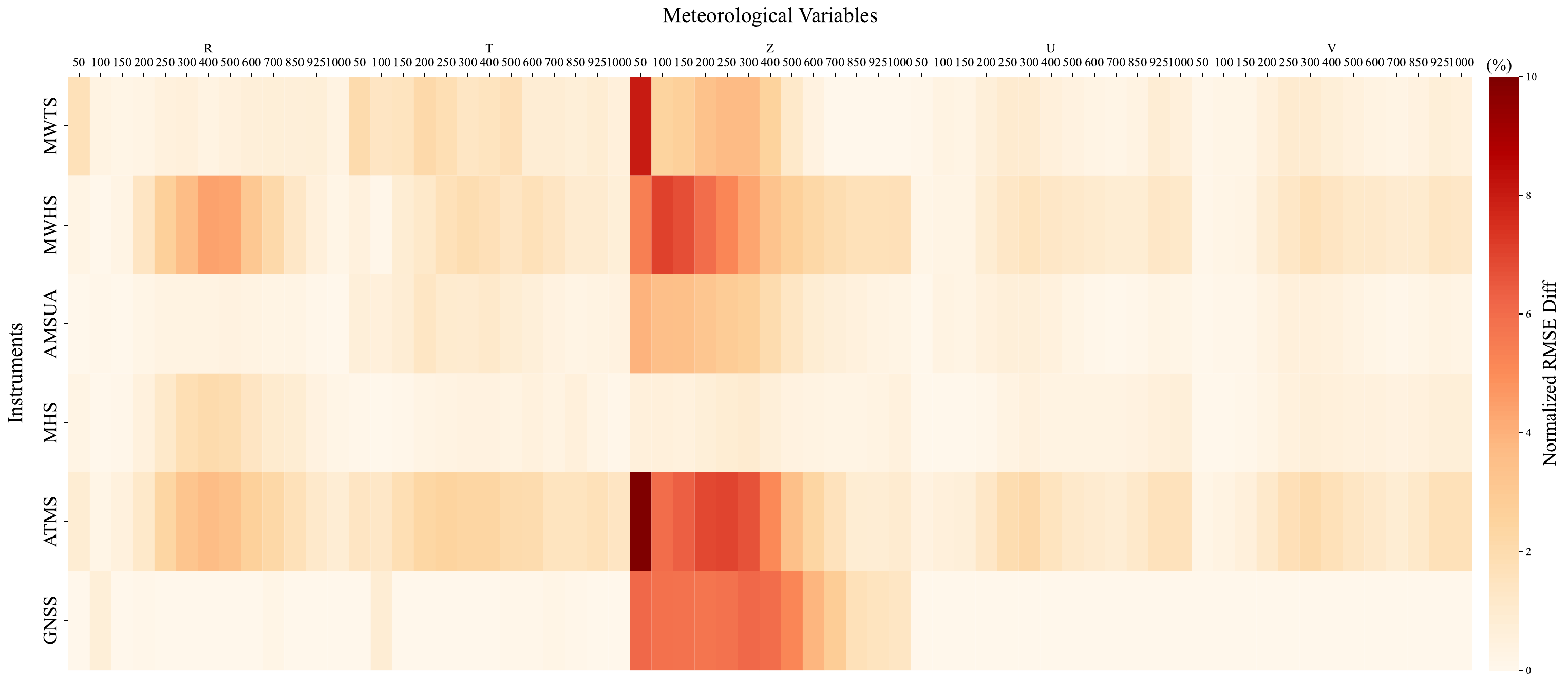}
\caption{Percentage change in the average errors of analysis fields from data denial experiments. The figure displays the normalized differences in globally-averaged, latitude-weighted root mean square error (RMSE) of analysis fields produced by the FuXi Weather system over a 1-year testing period (July 3, 2023, to June 30, 2024). The data from each satellite instrument are excluded individually, and the resulting analysis RMSE is compared with that from the control run, which assimilates all satellite data. The comparison includes a total of five upper-air atmospheric variables across 13 pressure levels (50, 100, 150, 200, 250, 300, 400, 500, 600, 700, 850, 925, and 1000 hPa). The five upper-air atmospheric variables are relative humidity (${\textrm{R}}$), temperature (${\textrm{T}}$), geopotential (${\textrm{Z}}$), u component of wind (${\textrm{U}}$), and v component of wind (${\textrm{V}}$). Red shading indicates degradation in accuracy compared with the control run, with varying impacts across variables and pressure levels.}
\label{demo_off_final}
\end{figure}

Figure \ref{demo_off_final} illustrates the normalized differences in the globally-averaged, latitude-weighted RMSE of the analysis fields.
Denying MWHS and ATMS led to the largest accuracy reductions for R, because these instruments provide global temperature and humidity soundings under all-weather conditions.
In contrast, excluding MHS had a negligible impact owing to its lower resolution and fewer channels.
Temperature was most affected by the removal of MWTS, MWHS, AMSU-A, and ATMS, which all include temperature-sounding channels.
The U and V wind components were sensitive to the denial of humidity sounders, particularly MWHS and ATMS, owing to the "generalized tracer effect".
Geopotential suffered the greatest degradation, with RMSE values increasing by 20\% at multiple pressure levels, especially when MWHS, ATMS, and GNSS-RO data were excluded, highlighting the sensitivity of geopotential to accurate representations of temperature, humidity, and wind.
It is noteworthy that the denial of ATMS primarily affected ${\textrm{R}}$ below 200 hPa, consistent with its Jacobian functions in Supplementary Fig.\ref{jacobian_ATMS}, where values are close to zero above 200 hPa.
The impact of MWTS on T was greater at higher altitudes, consistent with its greater Jacobian values at these altitudes.

\begin{figure}[h]
\centering
\includegraphics[width=1\textwidth]{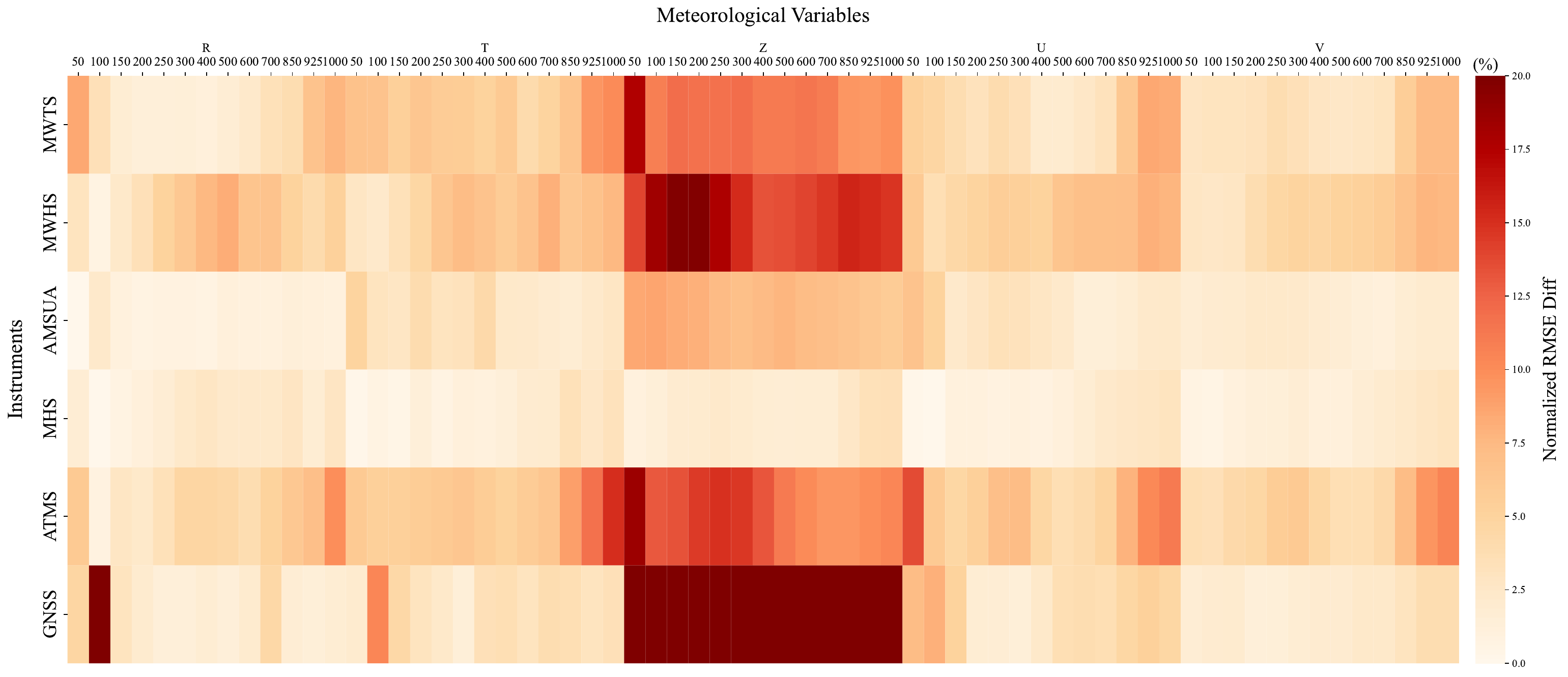}
\caption{Percentage change in the average errors of analysis fields from data denial experiments. The figure displays the normalized differences in globally-averaged, latitude-weighted root mean square error (RMSE) of analysis fields produced by the FuXi Weather system over a 1-year testing period (July 3, 2023 - June 30, 2024). The data from each satellite instrument are excluded individually, and the resulting analysis RMSE is compared with that of the control run, which assimilates all satellite data.}
\label{demo_online_final}
\end{figure}

Extended Data Fig\ref{demo_online_final} presents results from the second set of experiments, involving more complex variable correlations.
Removing specific satellite data degraded not only the analysis fields but also the subsequent forecasts, because errors propagated across variables.
For example, while GNSS-RO denial primarily affected geopotential at upper pressure levels in the first set of experiments, it impacted nearly all variables in this second set.
Despite this, FuXi Weather demonstrated robustness, with error growth remaining within acceptable limits over the 1-year testing period.

Overall, MWTS and MWHS aboard FY-3E and ATMS aboard NOAA-20 provided the most influential observations.
These data denial experiments are valuable in assessing the performance of DA systems with limited observations and help to evaluate the relative impacts of different observations as new data sources become available.
The results generally aligned with the temperature and humidity Jacobian functions, confirming the ability of FuXi Weather to evaluate the impact of different observational systems on analysis and forecast accuracy.

\FloatBarrier

\FloatBarrier

\section{Spatial distributions of global weather forecast performance}
\label{comparison_forecast}

\subsection{Forecasts for upper-air variables}
\label{comparison_upper}

\begin{figure}[h]
    \centering
    \includegraphics[width=\linewidth]{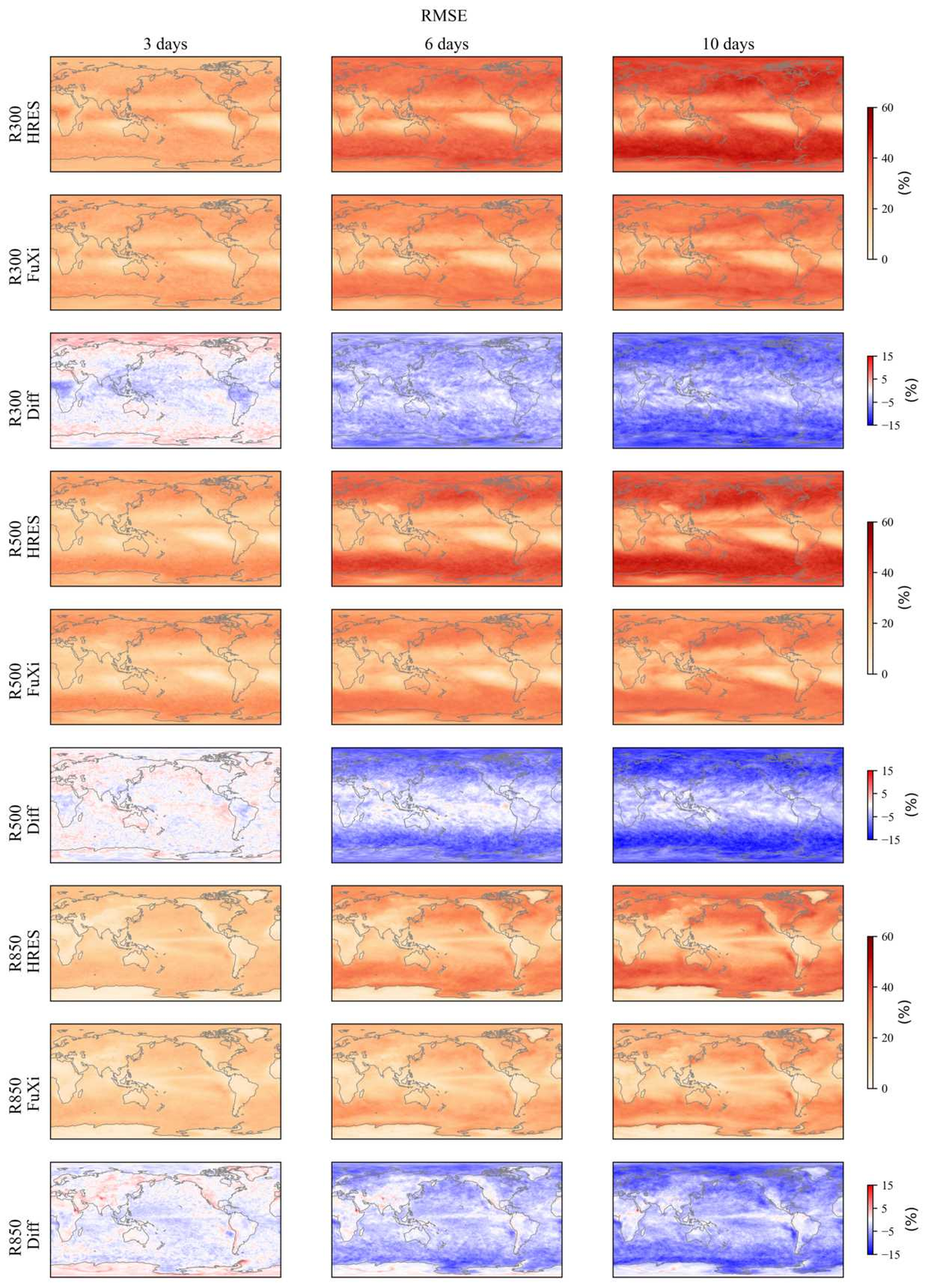}
    \caption{Spatial maps of average root-mean-square-error ($\textrm{RMSE}$) without latitude weighting of forecasts from ECMWF HRES (first, fourth, and seventh rows) and FuXi (second, fifth, and eighth rows), along with the $\textrm{RMSE}$ differences (third, sixth, and ninth rows) between FuXi and ECMWF HRES for relative humidity (${\textrm{R}}$) across three pressure levels (300, 500, and 850 hPa). The maps correspond to forecast lead times of 3 days (first column), 6 days (second column) and 10 days (third column), using all testing data over a 1-year testing period, spanning July 03, 2023-June 30, 2024.}
    \label{RMSE_spatial_R}    
\end{figure}

\begin{figure}[h]
    \centering
    \includegraphics[width=\linewidth]{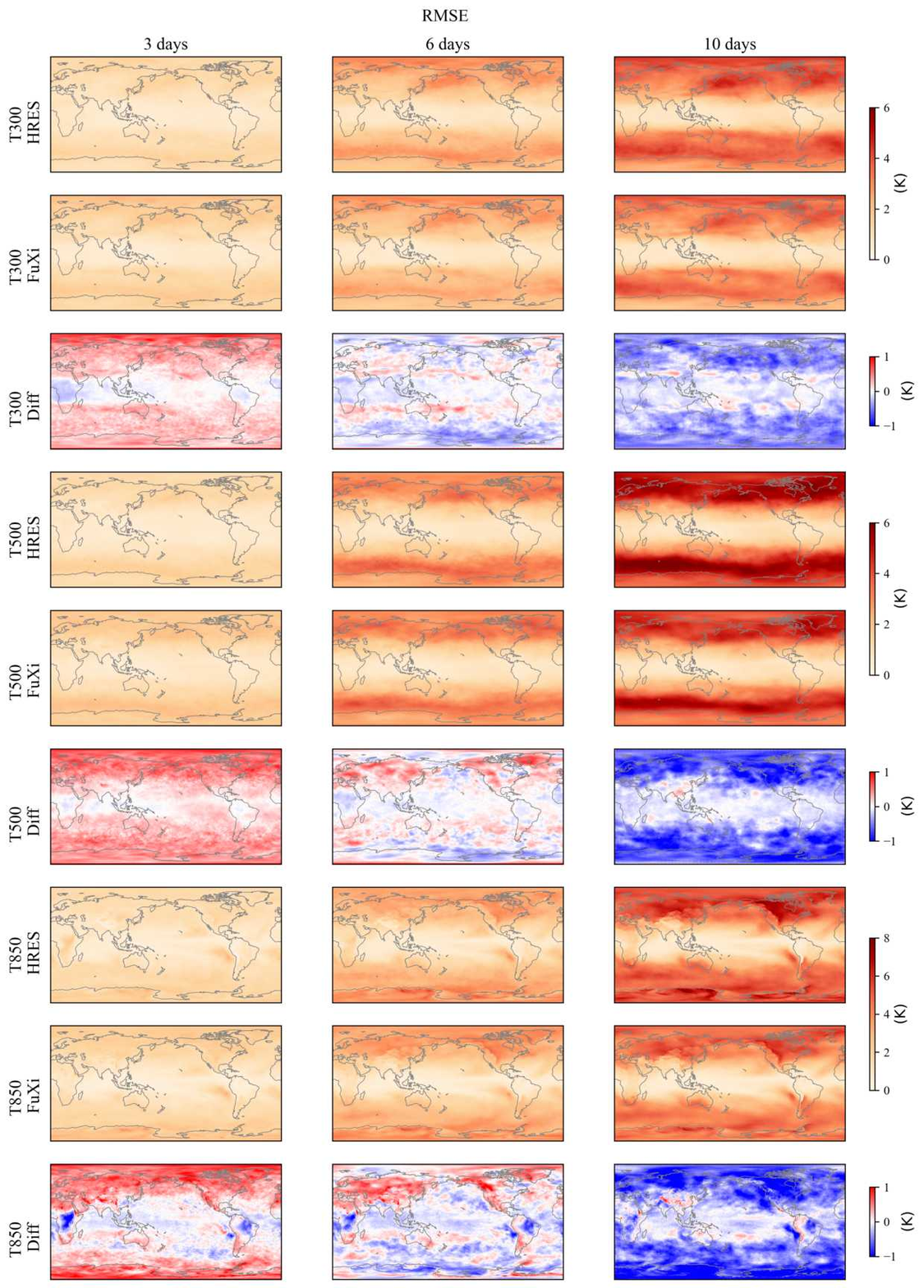}
    \caption{Spatial maps of average root-mean-square-error ($\textrm{RMSE}$) without latitude weighting of forecasts from ECMWF HRES (first, fourth, and seventh rows) and FuXi (second, fifth, and eighth rows), along with the $\textrm{RMSE}$ differences (third, sixth, and ninth rows) between FuXi and ECMWF HRES for temperature (${\textrm{T}}$) across three pressure levels (300, 500, and 850 hPa). The maps correspond to forecast lead times of 3 days (first column), 6 days (second column) and 10 days (third column), using all testing data over a 1-year testing period, spanning July 03, 2023-June 30, 2024.}
    \label{RMSE_spatial_T}    
\end{figure}

\begin{figure}[h]
    \centering
    \includegraphics[width=\linewidth]{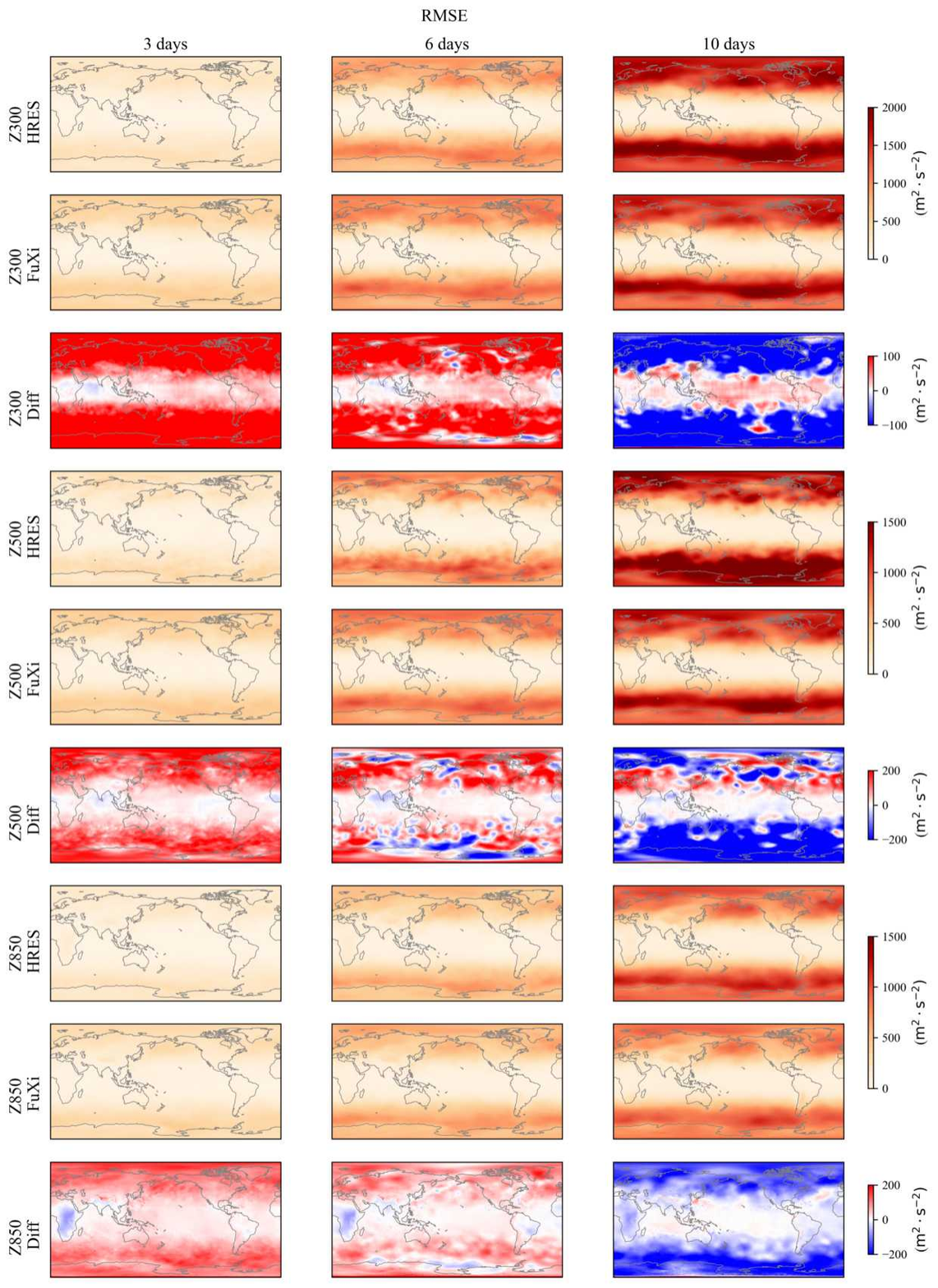}
    \caption{Spatial maps of average root-mean-square-error ($\textrm{RMSE}$) without latitude weighting of forecasts from ECMWF HRES (first, fourth, and seventh rows) and FuXi (second, fifth, and eighth rows), along with the $\textrm{RMSE}$ differences (third, sixth, and ninth rows) between FuXi and ECMWF HRES for geopotential (${\textrm{Z}}$) across three pressure levels (300, 500, and 850 hPa). The maps correspond to forecast lead times of 3 days (first column), 6 days (second column) and 10 days (third column), using all testing data over a 1-year testing period, spanning July 03, 2023-June 30, 2024.}
    \label{RMSE_spatial_Z}    
\end{figure}

\begin{figure}[h]
    \centering
    \includegraphics[width=\linewidth]{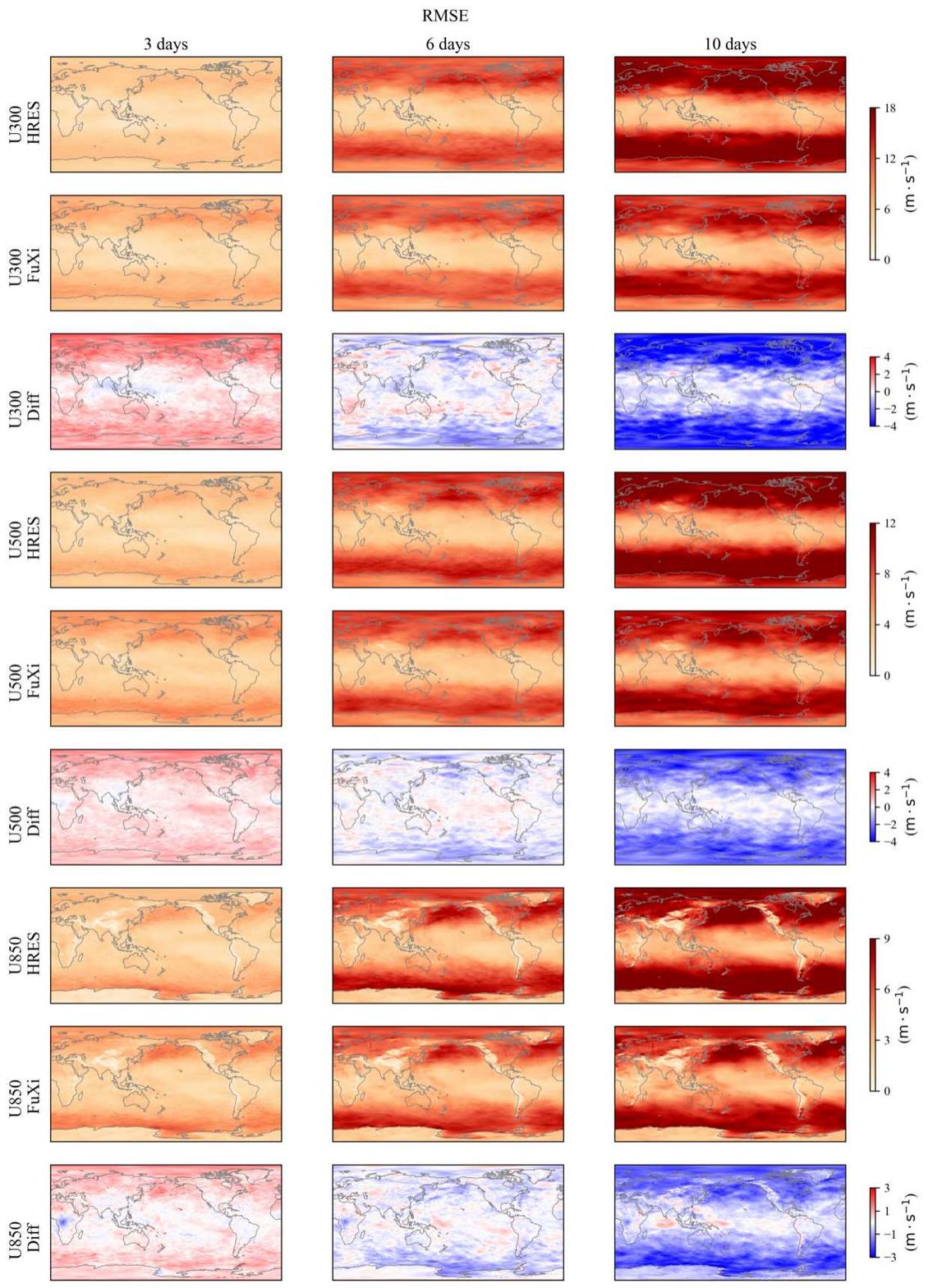}
    \caption{Spatial maps of average root-mean-square-error ($\textrm{RMSE}$) without latitude weighting of forecasts from ECMWF HRES (first, fourth, and seventh rows) and FuXi (second, fifth, and eighth rows), along with the $\textrm{RMSE}$ differences (third, sixth, and ninth rows) between FuXi and ECMWF HRES for the u component of wind (${\textrm{U}}$) across three pressure levels (300, 500, and 850 hPa). The maps correspond to forecast lead times of 3 days (first column), 6 days (second column) and 10 days (third column), using all testing data over a 1-year testing period, spanning July 03, 2023-June 30, 2024.}
    \label{RMSE_spatial_U}    
\end{figure}

\begin{figure}[h]
    \centering
    \includegraphics[width=\linewidth]{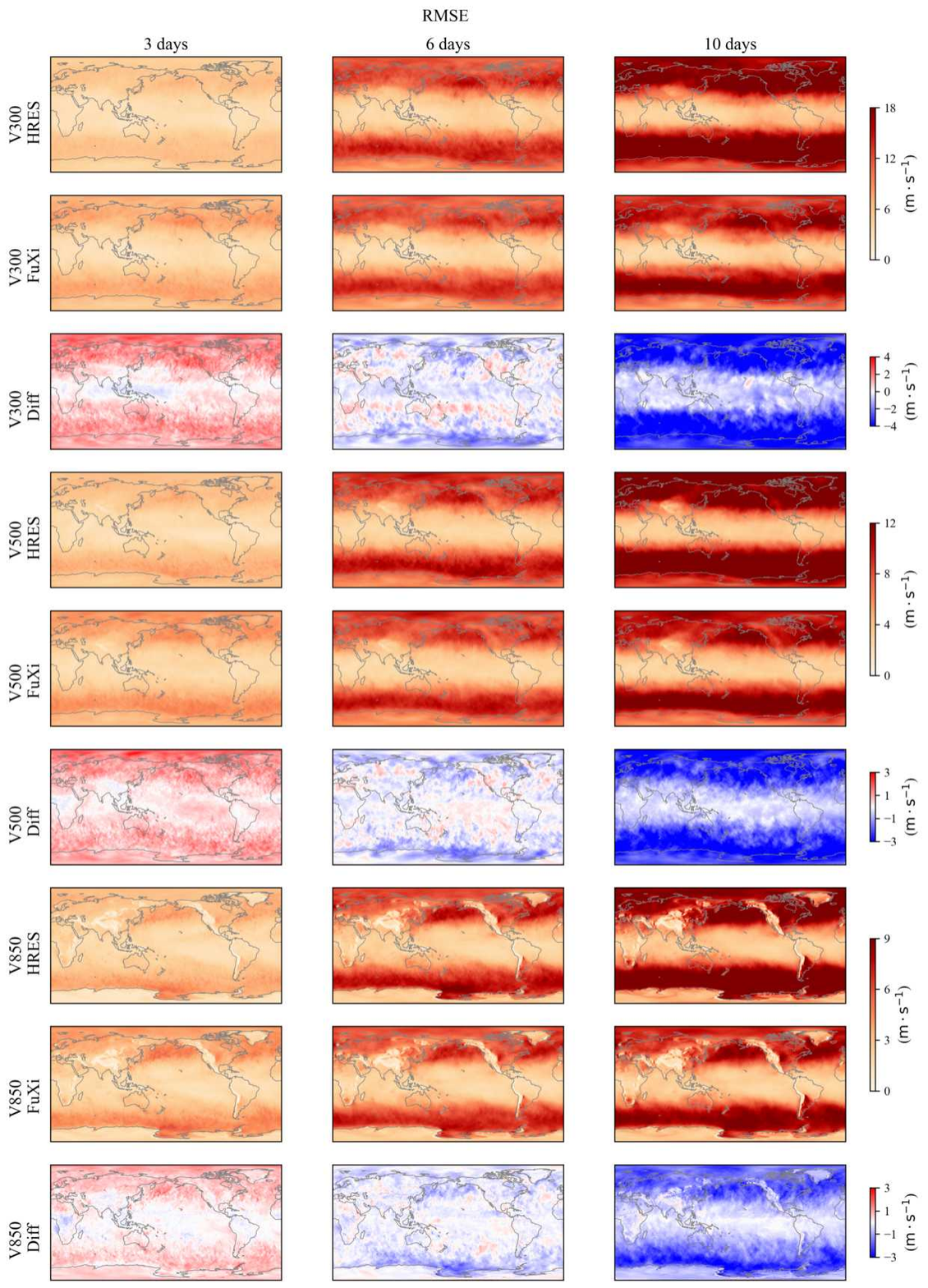}
    \caption{Spatial maps of average root-mean-square-error ($\textrm{RMSE}$) without latitude weighting of forecasts from ECMWF HRES (first, fourth, and seventh rows) and FuXi (second, fifth, and eighth rows), along with the $\textrm{RMSE}$ differences (third, sixth, and ninth rows) between FuXi and ECMWF HRES for the v component of wind (${\textrm{V}}$) across three pressure levels (300, 500, and 850 hPa). The maps correspond to forecast lead times of 3 days (first column), 6 days (second column) and 10 days (third column), using all testing data over a 1-year testing period, spanning July 03, 2023-June 30, 2024.}
    \label{RMSE_spatial_V}    
\end{figure}

Supplementary Figs. \ref{RMSE_spatial_R}-\ref{RMSE_spatial_V} present spatial comparisons of the average $\textrm{RMSE}$ without latitude weighting for forecasts from ECMWF HRES and FuXi.
These figures illustrate $\textrm{RMSE}$ and values for three pressure levels (300, 500, and 850 hPa) and five upper-air variables: ${\textrm{R}}$, ${\textrm{T}}$, ${\textrm{Z}}$, ${\textrm{U}}$, and ${\textrm{V}}$.
Forecast lead times of 3, 6, and 10 days are represented in the first, second, and third columns of each figure, respectively.
Darker red shades indicate higher RMSE values, showing a general increase in error with longer lead times, particularly in extra-tropical regions.

Spatial maps of RMSE differences reveal areas where FuXi outperforms ECMWF (blue), where ECMWF is superior (red), and where the two performances are comparable (white).
At a 3-day lead time, red dominates the maps for the variables T, Z, U, and V, suggesting that ECMWF is superior, especially in extra-tropical regions. However, for R, FuXi shows comparable or better performance at 300 and 850 hPa, aligning with the latitude-weighted RMSE analysis in main text Fig.2.
At a 10-day lead time, most regions exhibit predominantly blue or white colors, suggesting an overall better performance by FuXi, particularly in extratropical areas.
The intermediate 6-day forecast shows a more prominent shift to blue for variable R than for the four other variables, indicating the enhanced performance of FuXi for this variable among all evaluated upper-air parameters.

Notably, FuXi consistently demonstrates lower RMSE values over central Africa for R, T, and Z at all lead times, suggesting a consistently better performance in this region compared with other parts of the world.
This is likely result of the scarcity of conventional observations in Africa, which limits the data available for ECMWF HRES assimilation.
Consequently, satellite data plays a more prominent role in Africa.
Additionally, owing to the predominantly westerly winds and high topography of eastern Africa, including prominent mountain ranges, these improvements tend to propagate westwards toward South America rather than eastwards.
As a result, the forecast accuracy over South America tends to increase with longer forecast lead times.
The superior performance of FuXi Weather also propagates to the North Pole and South Pole; this is particularly pronounced at a lead time of 10 days.

\FloatBarrier

\subsection{Forecasts for surface variables}
\label{comparison_surface}

\begin{figure}[h]
    \centering
    \includegraphics[width=\linewidth]{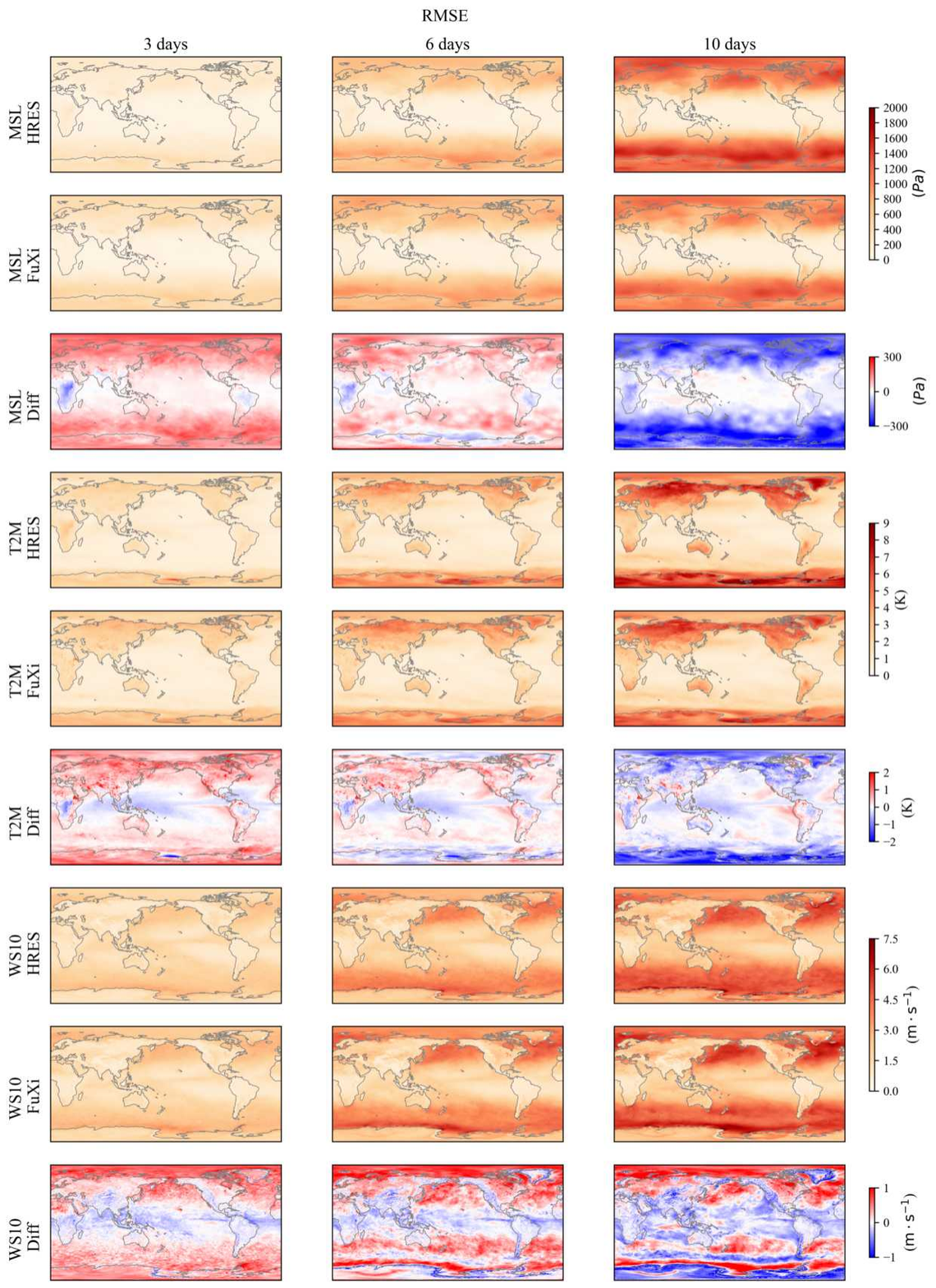}
    \caption{Spatial maps of average root-mean-square-error ($\textrm{RMSE}$) without latitude weighting of forecasts from ECMWF HRES (first, fourth, and seventh rows) and FuXi (second, fifth, and eighth rows), along with the $\textrm{RMSE}$ differences (third, sixth, and ninth rows) between FuXi and ECMWF HRES for mean sea-level pressure ($\textrm{MSL}$), 2-meter temperature ($\textrm{T2M}$), and 10-meter wind speed ($\textrm{WS10M}$). The maps correspond to forecast lead times of 3 days (first column), 6 days (second column) and 10 days (third column), using all testing data over a 1-year testing period, spanning July 03, 2023-June 30, 2024.}
    \label{RMSE_spatial_surface}    
\end{figure}

Supplementary Fig. \ref{RMSE_spatial_surface} presents spatial comparisons of the average RMSE, without latitude weighting, for ECMWF HRES and FuXi forecasts of surface variables, including $\textrm{MSL}$, $\textrm{T2M}$, and 10-meter wind speed ($\textrm{WS10M}$).
Consistent with the results for upper-air variables, the figure shows that RMSE differences are predominantly negative over central Africa at forecast lead times of 3, 6, and 10 days, indicating superior performance of FuXi Weather.

Additionally, FuXi Weather outperforms ECMWF HRES in other regions where there are sparse land-based observations, such as India and areas near the North Pole and South Pole.
This outperformance becomes more pronounced as the forecast lead time extends from 6 to 10 days, with an increasing area showing negative RMSE differences; this indicates the growing advantage in forecast accuracy of FuXi Weather over longer lead times.

\section{Effect of different training settings}
\label{effect_setting}
This section evaluates the performance of FuXi Weather under four training configurations, detailed in Supplementary Table \ref{setting}.
Corresponding to the FuXi Weather configuration discussed in the main text, setting 1 includes incremental learning in FuXi-DA and fine-tuning of FuXi-Short. Setting 2 excludes incremental learning, setting 3 omits fine-tuning of FuXi-Short, and setting 4 excludes both.

\begin{figure}[h]
    \centering
    \includegraphics[width=\linewidth]{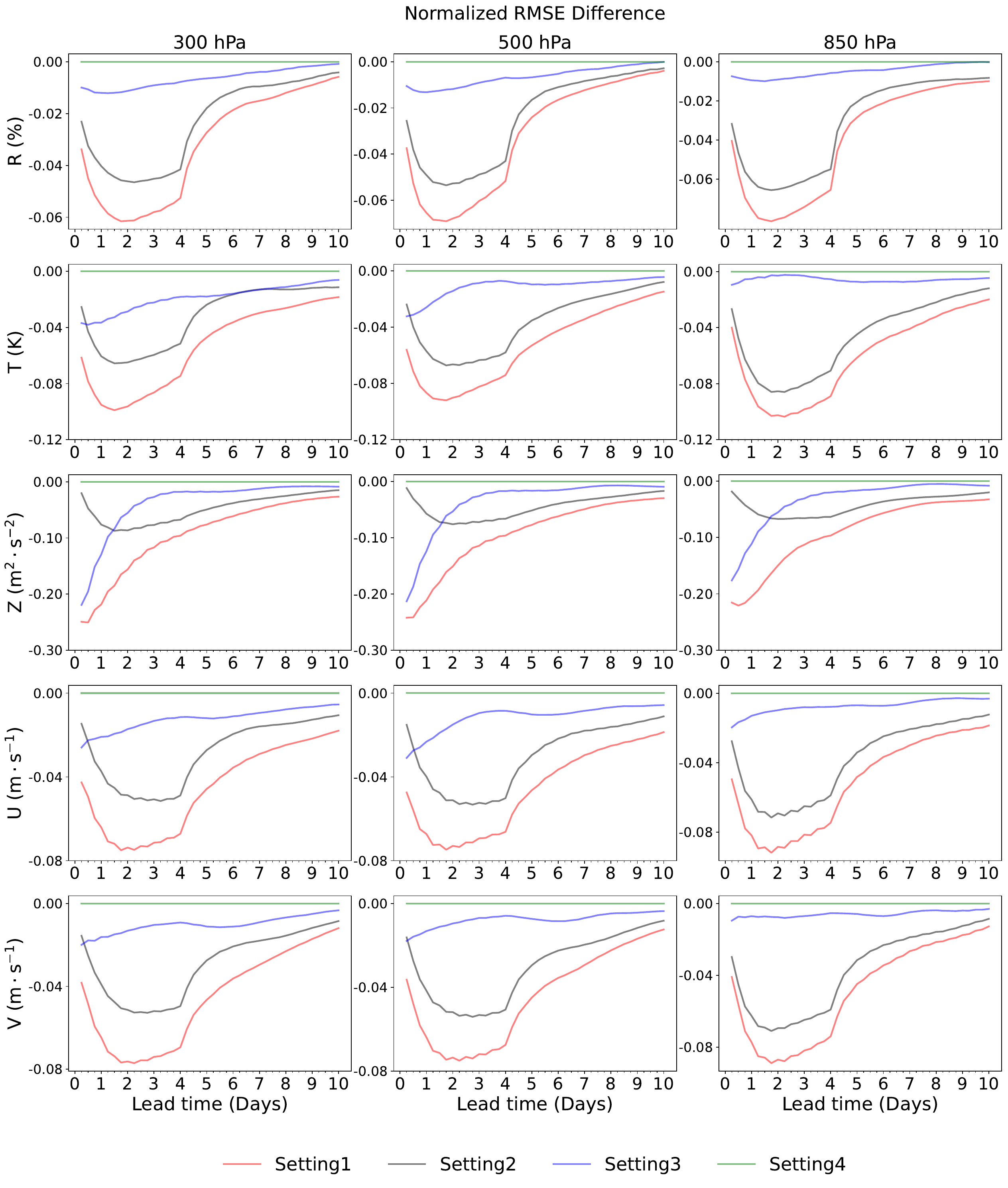}
    \caption{Comparison of normalized differences in globally-averaged, latitude-weighted root-mean-square-error ($\textrm{RMSE}$) across four FuXi Weather training settings. The settings differ based on the implementation of incremental learning in FuXi-DA and whether the FuXi-Short forecasting model is fine-tuned (see Supplementary Table \ref{setting}). The comparison spans a 1-year testing period, spanning July 03, 2023-June 30, 2024. The analysis includes five variables: relative humidity (${\textrm{R}}$), temperature (${\textrm{T}}$), geopotential (${\textrm{Z}}$), u component of wind (${\textrm{U}}$), and v component of wind (${\textrm{V}}$), at three pressure levels (300, 500, and 850 hPa). The five rows and three columns correspond to the five variables and three pressure levels, respectively.}
    \label{paper_Norm_RMSE_5obs_abstudy}        
\end{figure}

\begin{figure}[h]
    \centering
    \includegraphics[width=\linewidth]{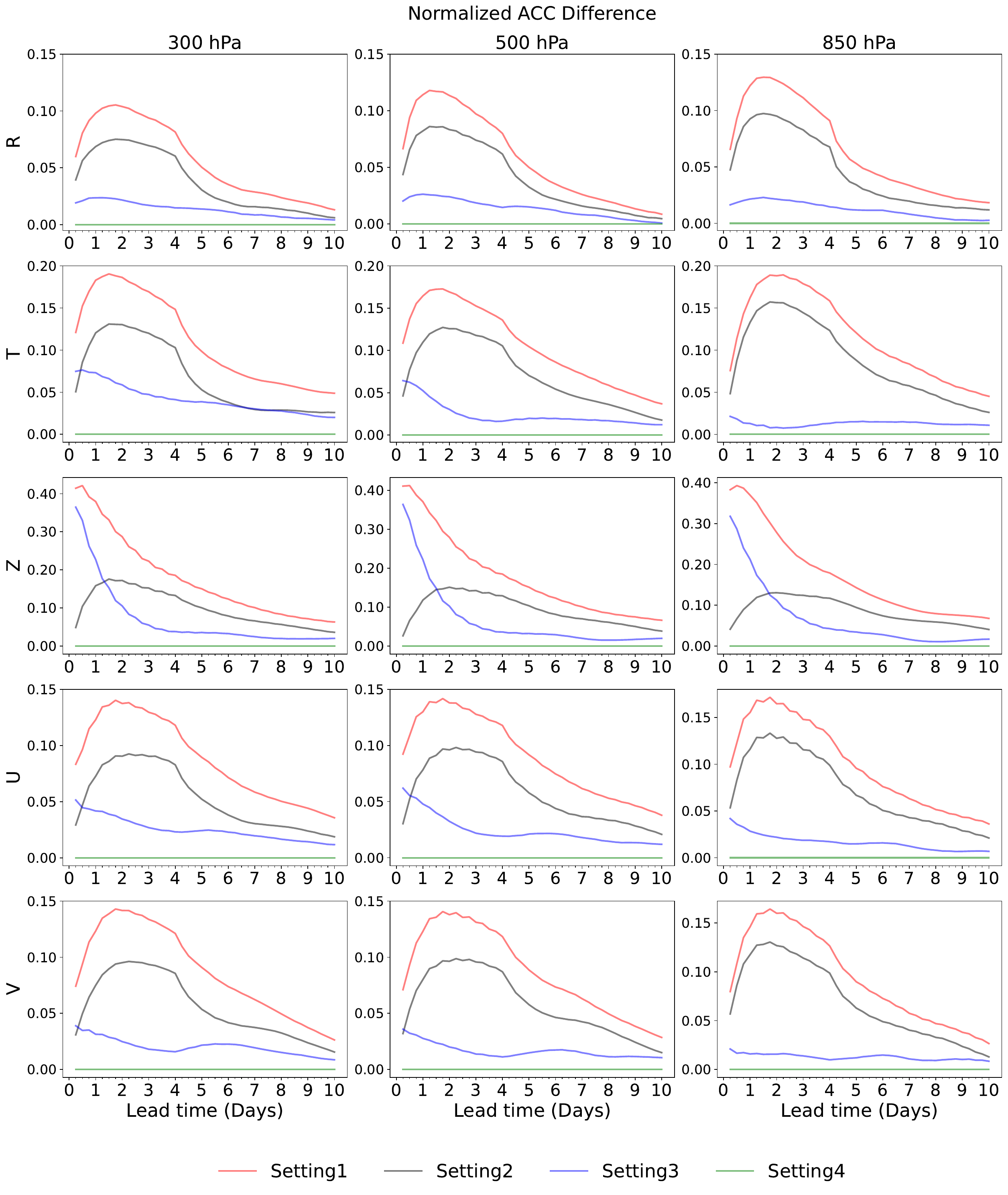}
    \caption{Comparison of normalized differences in globally-averaged, latitude-weighted anomaly correlation coefficient ($\textrm{ACC}$) across four FuXi Weather training settings. The settings differ based on the implementation of incremental learning in FuXi-DA and whether the FuXi-Short forecasting model is fine-tuned (see Supplementary Table \ref{setting}). The comparison spans a 1-year testing period, spanning July 03, 2023-June 30, 2024. The analysis includes five variables: relative humidity (${\textrm{R}}$), temperature (${\textrm{T}}$), geopotential (${\textrm{Z}}$), u component of wind (${\textrm{U}}$), and v component of wind (${\textrm{V}}$), at three pressure levels (300 hPa, 500 hPa, and 850 hPa). The five rows and three columns correspond to the five variables and three pressure levels, respectively.}
    \label{paper_Norm_ACC_5obs_abstudy}        
\end{figure}

Supplementary Figs.\ref{paper_Norm_RMSE_5obs_abstudy} and \ref{paper_Norm_ACC_5obs_abstudy} present the normalized differences in globally averaged, latitude-weighted RMSE and ACC for the four models, using setting 4 as the baseline. The results indicate that setting 1, which incorporates both incremental learning and fine-tuning, yields the best performance. A comparison between settings 2 and 3 reveals that incremental learning has a more substantial positive impact than does fine-tuning alone.

\begin{figure}[h]
    \centering
    \includegraphics[width=\linewidth]{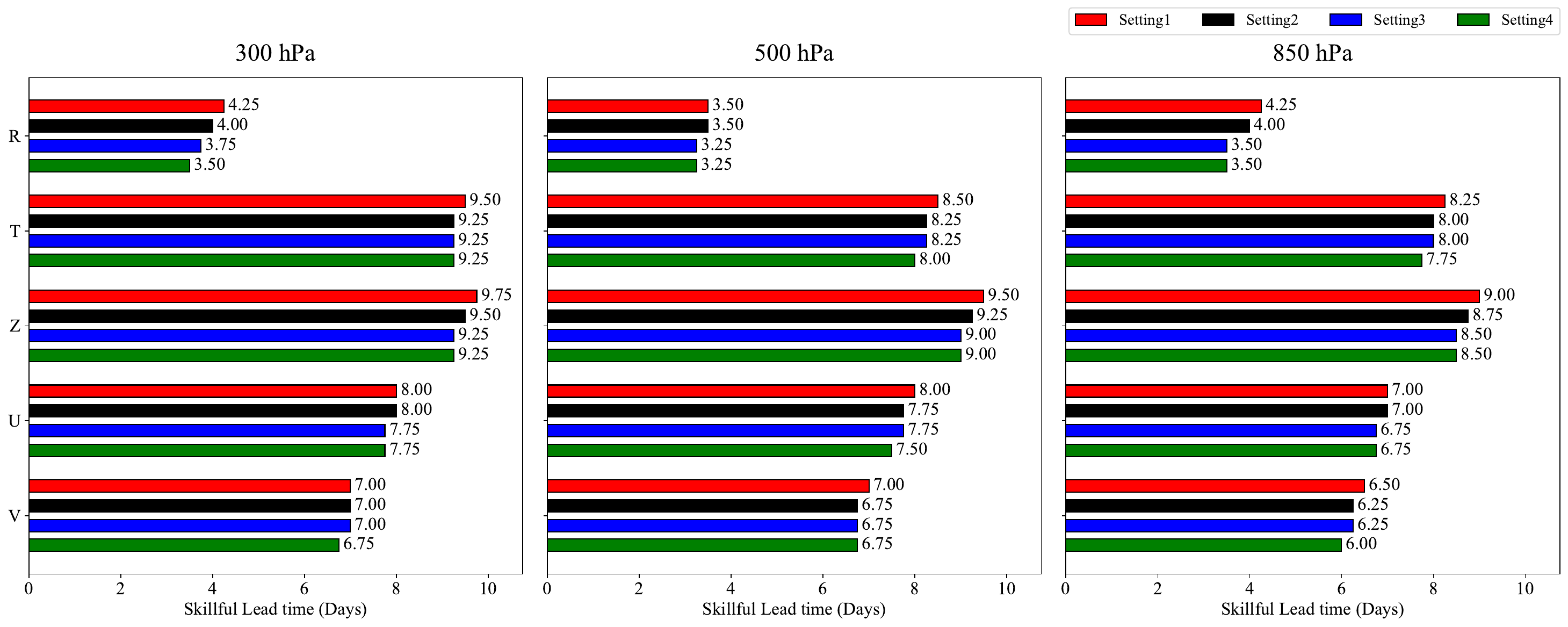}
    \caption{Comparisons of skillful forecast lead times across four FuXi Weather training settings. The settings differ based on the implementation of incremental learning in FuXi-DA and whether the FuXi-Short forecasting model is fine-tuned (see Supplementary Table \ref{setting}). Skillful forecast lead time is defined as the anomaly correlation coefficient ($\textrm{ACC}$) value above the threshold of 0.6. The comparison spans a 1-year testing period, spanning July 03, 2023-June 30, 2024. The analysis includes five variables: relative humidity (${\textrm{R}}$), temperature (${\textrm{T}}$), geopotential (${\textrm{Z}}$), u component of wind (${\textrm{U}}$), and v component of wind (${\textrm{V}}$), at three pressure levels (300, 500, and 850 hPa). The five rows and three columns correspond to the five variables and three pressure levels, respectively.}
    \label{skillful_lead_setting}    
\end{figure}

Supplementary Fig. \ref{skillful_lead_setting} further compares skillful forecast lead times across the four models, supporting the findings described above. The model trained with setting 1 achieves the longest skillful lead times across all 15 variable and pressure-level combinations, confirming the benefits of combining incremental learning with fine-tuning.

\begin{table}[h]

\centering
\caption{\label{setting} Overview of training settings for FuXi Weather. The settings vary based on the implementation of incremental learning in FuXi-DA and whether the FuXi-Short forecasting model is fine-tuned. Setting 1, which includes incremental learning in FuXi-DA and fine-tuning of FuXi-Short, corresponds to the FuXi Weather configuration discussed in the main text.}
\begin{tabularx}{\textwidth}{cXX}
\hline
\textbf{Setting} & \textbf{Incremental learning} & \textbf{Fine-tuning of FuXi-Short} \\
\hline
Setting 1 & \Checkmark &  \Checkmark  \\
Setting 2 & \XSolid &  \Checkmark  \\
Setting 3 & \Checkmark &  \XSolid  \\
Setting 4 & \XSolid &  \XSolid  \\
\hline
\end{tabularx}
\end{table}
\FloatBarrier

\FloatBarrier

\FloatBarrier

\noindent

\end{document}